\documentclass{article}

\PassOptionsToPackage{square,numbers,sort&compress}{natbib}
\usepackage[final]{neurips_2021}

\usepackage[utf8]{inputenc} %
\usepackage[T1]{fontenc}    %
\usepackage{hyperref}       %
\usepackage{url}            %
\usepackage{booktabs}       %
\usepackage{amsfonts}       %
\usepackage{nicefrac}       %
\usepackage{microtype}      %
\usepackage{xcolor}         %

\usepackage{pgffor} %
\usepackage{xstring} %
\usepackage{todonotes}
\usepackage{wrapfig}
\usepackage{tabularx} %
\usepackage{standalone}
\usepackage{caption}
\usepackage{subcaption}
\usepackage{wasysym} %
\usepackage{amssymb} %
\usepackage{tabularx} %
\usepackage{listings} %

\usepackage{xcolor}
\usepackage{color}
\definecolor{red}{RGB}{165, 30, 55}
\definecolor{gold}{RGB}{180, 160, 105}
\definecolor{metallic}{RGB}{50, 65, 75}
\definecolor{blue1}{RGB}{65, 90, 140}
\definecolor{blue2}{RGB}{0, 105, 170}
\definecolor{blue3 }{RGB}{80, 170, 200}
\definecolor{green1}{RGB}{50, 110, 30}
\definecolor{green2}{RGB}{125, 165, 75}
\definecolor{green3}{RGB}{130, 185, 160}
\definecolor{grey1}{RGB}{180, 160, 150}
\definecolor{purple1}{RGB}{175, 110, 150}
\definecolor{red1}{RGB}{200, 80, 60}
\definecolor{brown1}{RGB}{145, 105, 70}
\definecolor{orange1}{RGB}{210, 150, 0}
\definecolor{orange2}{RGB}{215, 180, 105}
\definecolor{black}{RGB}{0, 0, 0}
\definecolor{supervised.grey}{RGB}{150, 150, 150}

\definecolor{bitm.purple2}{RGB}{153, 142, 195}
\definecolor{red.arrow}{RGB}{255, 0, 0}
\definecolor{clip.hard.labels}{RGB}{235, 217, 181}
\definecolor{clip.soft.labels}{RGB}{215, 180, 105}

\title{Partial success in closing the gap between\\human and machine vision}

\author{
  Robert Geirhos\textsuperscript{1-2$\S$}
  \And
  Kantharaju Narayanappa\textsuperscript{1}
  \And
  Benjamin Mitzkus\textsuperscript{1}
  \AND
  Tizian Thieringer\textsuperscript{1}
  \And
  Matthias Bethge\textsuperscript{1$\ast$}
  \And
  Felix A.\ Wichmann\textsuperscript{1$\ast$}
  \And
  Wieland Brendel\textsuperscript{1$\ast$}
  \AND
  \vspace{-0.5cm}
  \\ \multicolumn{1}{c}{\textsuperscript{1}{University of T\"{u}bingen}}
  \\ \multicolumn{1}{c}{\textsuperscript{2}{International Max Planck Research School for Intelligent Systems}}
  \\ \multicolumn{1}{c}{\textsuperscript{$\ast$}{Joint senior authors}}
  \\ \multicolumn{1}{c}{\textsuperscript{$\S$}{To whom correspondence should be addressed: \texttt{robert.geirhos@uni-tuebingen.de}}}
}

\graphicspath{{./figures/results/}}

\begin{document}

\newcommand{\figwidth}{0.245\textwidth}
\newcommand{\captionspace}{-1.5\baselineskip}
\newcommand{\captionspaceII}{0.2\baselineskip}
\newcommand{\captionspaceBenchmark}{-1.3\baselineskip}

\maketitle

\begin{abstract}
A few years ago, the first CNN surpassed human performance on ImageNet. However, it soon became clear that machines lack robustness on more challenging test cases, a major obstacle towards deploying machines ``in the wild'' and towards obtaining better computational models of human visual perception. Here we ask: Are we making progress in closing the gap between human and machine vision? To answer this question, we tested human observers on a broad range of out-of-distribution (OOD) datasets, recording 85,120 psychophysical trials across 90 participants. We then investigated a range of promising machine learning developments that crucially deviate from standard supervised CNNs along three axes: objective function (self-supervised, adversarially trained, CLIP language-image training), architecture (e.g.\ vision transformers), and dataset size (ranging from 1M to 1B).\\ 
Our findings are threefold. (1.) The longstanding \emph{distortion robustness gap} between humans and CNNs is closing, with the best models now exceeding human feedforward performance on most of the investigated OOD datasets. (2.) There is still a substantial image-level \emph{consistency gap}, meaning that humans make different errors than models. In contrast, most models systematically agree in their categorisation errors, even substantially different ones like contrastive self-supervised vs.\ standard supervised models. (3.) In many cases, human-to-model consistency improves when training dataset size is increased by one to three orders of magnitude. Our results give reason for cautious optimism: While there is still much room for improvement, the behavioural difference between human and machine vision is narrowing. In order to measure future progress, 17 OOD datasets with image-level human behavioural data and evaluation code are provided as a toolbox and benchmark at \url{https://github.com/bethgelab/model-vs-human/}.
\end{abstract}

\section{Introduction}
Looking back at the last decade, deep learning has made tremendous leaps of progress by any standard. What started in 2012 with AlexNet \citep{krizhevsky2012imagenet} as the surprise winner of the ImageNet Large-Scale Visual Recognition Challenge quickly became the birth of a new AI ``summer'', a summer lasting much longer than just a season. With it, just like with any summer, came great expectations: the hope that the deep learning revolution will see widespread applications in industry, that it will propel breakthroughs in the sciences, and that it will ultimately close the gap between human and machine perception. We have now reached the point where deep learning has indeed become a significant driver of progress in industry \citep[e.g.][]{wang2018smart, villalba2019deep}, and where many disciplines are employing deep learning for scientific discoveries \citep{angermueller2016deep, marblestone2016toward, goh2017deep, ching2018opportunities, guest2018deep, senior2020improved}---\emph{but are we making progress in closing the gap between human and machine vision?} 

\paragraph{IID vs.\ OOD benchmarking.} For a long time, the gap between human and machine vision was mainly approximated by comparing benchmark accuracies on IID (independent and identically distributed) test data: as long as models are far from reaching human-level performance on challenging datasets like ImageNet, this approach is adequate \citep{Russakovsky2015}. Currently, models are routinely matching and in many cases even outperforming humans on IID data. At the same time, it is becoming increasingly clear that models systematically exploit shortcuts shared between training and test data \citep{jo2017measuring, beery2018recognition, niven2019probing, geirhos2020shortcut}. Therefore we are witnessing a major shift towards measuring model performance on out-of-distribution (OOD) data rather than IID data alone, which aims at testing models on more challenging test cases where there is still a ground truth category, but certain image statistics differ from the training distribution. Many OOD generalisation tests have been proposed: ImageNet-C \citep{hendrycks2019benchmarking} for corrupted imaes, ImageNet-Sketch \citep{wang2019learning} for sketches, Stylized-ImageNet \citep{geirhos2019imagenettrained} for image style changes, \citep{alcorn2019strike} for unfamiliar object poses, and many more \citep{lake2015human, barbu2019objectnet, michaelis2019benchmarking, taori2020measuring, djolonga2021robustness, dunn2021exposing, he2021towards, hendrycks2021many, hendrycks2021natural, madan2021small, shankar2021image}. While it is great to have many viable and valuable options to measure generalisation, most of these datasets unfortunately lack human comparison data. This is less than ideal, since we can no longer assume that humans reach near-ceiling accuracies on these challenging test cases as they do on standard noise-free IID object recognition datasets. In order to address this issue, we carefully tested human observers in the Wichmannlab's vision laboratory on a broad range of OOD datasets, providing some 85K psychophysical trials across 90 participants. Crucially, we showed exactly the same images to multiple observers, which means that we are able to compare human and machine vision on the fine-grained level of individual images \citep{green1964consistency,meding2019perceiving, geirhos2020beyond}). The focus of our datasets is measuring \emph{distortion robustness}: we tested 17 variations that include changes to image style, texture, and various forms of synthetic additive noise.

\paragraph{Contributions \& outlook.} The resulting 17 OOD datasets with large-scale human comparion data enable us to investigate recent exciting machine learning developments that crucially deviate from ``vanilla'' CNNs along three axes: objective function (supervised vs.\ self-supervised, adversarially trained, and CLIP's joint language-image training), architecture (convolutional vs.\ vision transformer) and training dataset size (ranging from 1M to 1B images). Taken together, these are some of the most promising directions our field has developed to date---but this field would not be machine learning if new breakthroughs weren't within reach in the next few weeks, months and years. Therefore, we open-sourced \href{https://github.com/bethgelab/model-vs-human}{\texttt{modelvshuman}}, a Python toolbox that enables testing both PyTorch and TensorFlow models on our comprehensive benchmark suite of OOD generalisation data in order to measure future progress. Even today, our results give cause for (cautious) optimism. After a method overview (Section~\ref{sec:methods}), we are able to report that the human-machine \emph{distortion robustness gap} is closing: the best models now match or in many cases even exceed human feedforward performance on most of the investigated OOD datasets (Section~\ref{sec:robustness_across_models}). While there is still a substantial image-level \emph{consistency gap} between humans and machines, this gap is narrowing on some---but not all---datasets when the size of the training dataset is increased~(Section~\ref{sec:consistency_between_models}).

\section{Methods: datasets, psychophysical experiments, models, metrics, toolbox}
\label{sec:methods}

\paragraph{OOD datasets with consistency-grade human data.} We collected human data for 17 generalisation datasets (visualized in Figures~\ref{fig:methods_nonparametric_stimuli} and \ref{fig:methods_parametric_stimuli} in the Appendix, which also state the number of subjects and trials per experiment) on a carefully calibrated screen in a dedicated psychophysical laboratory (a total of 85,120 trials across 90 observers). Five datasets each correspond to a single manipulation (sketches, edge-filtered images, silhouettes, images with a texture-shape cue conflict, and stylized images where the original image texture is replaced by the style of a painting); the remaining twelve datasets correspond to parametric image degradations (e.g.\ different levels of noise or blur). Those OOD datasets have in common that they are designed to test ImageNet-trained models. OOD images were obtained from different sources: sketches from ImageNet-Sketch \citep{wang2019learning}, stylized images from Stylized-ImageNet \citep{geirhos2019imagenettrained}, edge-filtered images, silhouettes and cue conflict images from \citep{geirhos2019imagenettrained}\footnote{For those three datasets consisting of 160, 160 and 1280 images respectively, consistency-grade psychophysical data was already collected by the authors and included in our benchmark with permission from the authors.}, and the remaining twelve parametric datasets were adapted from \citep{geirhos2018generalisation}. For these parametric datasets, \citep{geirhos2018generalisation} collected human accuracies but unfortunately, they showed different images to different observers implying that we cannot use their human data to assess image-level consistency between humans and machines. Thus we collected psychophysical data for those images ourselves by showing exactly the same images to multiple observers for each of those twelve datasets. Additionally, we cropped the images from \citep{geirhos2018generalisation} to $224\times224$ pixels to allow for a fair comparison to ImageNet models (all models included in our comparison receive $224\times224$ input images; \citep{geirhos2018generalisation} showed $256\times256$ images to human observers in many cases).

\paragraph{Psychophysical experiments.} 90 observers were tested in a darkened chamber. Stimuli were presented at the center of a 22'' monitor with $1920\times 1200$ pixels resolution (refresh rate: 120 Hz). Viewing distance was 107 cm and target images subtended $3\times3$ degrees of visual angle. Human observers were presented with an image and asked to select the correct category out of 16 basic categories (such as chair, dog, airplane, etc.). Stimuli were balanced w.r.t.\ classes and presented in random order. For ImageNet-trained models, in order to obtain a choice from the same 16 categories, the 1,000 class decision vector was mapped to those 16 classes using the WordNet hierarchy \citep{miller1995wordnet}. In Appendix~\ref{app:mapping_decisions}, we explain why this mapping is optimal. We closely followed the experimental protocol defined by \citep{geirhos2018generalisation}, who presented images for 200 ms followed by a 1/$f$ backward mask to limit the influence of recurrent processing (otherwise comparing to feedforward models would be difficult). Further experimental details are provided in Appendix~\ref{app:experimental_details}.

\paragraph{Why not use crowdsourcing instead?} Our approach of investigating few observers in a high-quality laboratory setting performing many trials is known as the so-called ``small-N design'', the bread-and-butter approach in high-quality psychophysics---see, e.g., the review ``Small is beautiful: In defense of the small-N design'' \citep{smith2018small}. This is in contrast to the ``crowdsourcing approach'' (many observers in a noisy setting performing fewer trials each). The highly controlled conditions of the Wichmannlab's psychophysical laboratory come with many advantages over crowdsourced data collection: precise timing control (down to the millisecond), carefully calibrated monitors (especially important for e.g. low-contrast stimuli), controlled viewing distance (important for foveal presentation), full visual acuity (we performed an acuity test with every observer prior to the experiment), observer attention (e.g. no multitasking or children running around during an experiment, which may happen in a crowdsourcing study), just to name a few \citep{haghiri2019comparison}. Jointly, these factors contribute to high data quality.

\paragraph{Models.} In order to disentangle the influence of objective function, architecture and training dataset size, we tested a total of 52 models: 24 standard ImageNet-trained CNNs \citep{marcel2010torchvision}, 8 self-supervised models \citep{wu2018unsupervised, he2020momentum, chen2020improved, misra2020self, tian2020makes, chen2020simple},\footnote{We presented a preliminary and much less comprehensive version of this work at the NeurIPS 2020 workshop SVRHM \citep{geirhos2020surprising}.} 6 Big Transfer models \citep{kolesnikov2019big}, 5 adversarially trained models \citep{salman2020adversarially}, 5 vision transformers \citep{dosovitskiy2020image, rw2019timm}, two semi-weakly supervised models \citep{yalniz2019billion} as well as Noisy Student \citep{xie2020self} and CLIP \citep{radford2021learning}. Technical details for all models are provided in the Appendix.

\paragraph{Metrics.}
In addition to \emph{OOD accuracy} (averaged across conditions and datasets), the following three metrics quantify how closely machines are aligned with the decision behaviour of humans.

\emph{Accuracy difference} $A(m)$ is a simple aggregate measure that compares the accuracy of a machine $m$ to the accuracy of human observers in different out-of-distribution tests,
\begin{equation}
A(m): \mathbb{R}\rightarrow [0,1], m \mapsto \frac{1}{|D|}\sum_{d \in D }\frac{1}{|H_d|}\sum_{h \in H_d }\frac{1}{|C_d|}\sum_{c \in C_d }(\mathrm{acc}_{d,c}(h) - \mathrm{acc}_{d,c}(m))^2
\end{equation}
where $\mathrm{acc}_{d,c}(\cdot)$ is the accuracy of the model or the human on dataset $d\in D$ and condition $c\in C_d$ (e.g.\ a particular noise level), and $h\in H_D$ denotes a human observer tested on dataset $d$. Analogously, one can compute the average accuracy difference between a human observer $h_1$ and all other human observers by substituting $h_1$ for $m$ and $h \in H_D\setminus\{h_1\}$ for $h \in H_D $ (which can also be applied for the two metrics defined below).

Aggregated metrics like $A(m)$ ignore individual image-level decisions. Two models with vastly different image-level decision behaviour might still end up with the same accuracies on each dataset and condition. Hence, we include two additional metrics in our benchmark that are sensitive to decisions on individual images.

\emph{Observed consistency} $O(m)$ \citep{geirhos2020beyond} measures the fraction of samples for which humans and a model $m$ get the same sample either both right \emph{or} both wrong. More precisely, let $b_{h,m}(s)$ be one if both a human observer $h$ and $m$ decide either correctly or incorrectly on a given sample $s$, and zero otherwise. We calculate the average observed consistency as

\begin{equation}
O(m): \mathbb{R}\rightarrow [0,1], m \mapsto \frac{1}{|D|}\sum_{d \in D }\frac{1}{|H_d|}\sum_{h \in H_d }\frac{1}{|C_d|}\sum_{c \in C_d }\frac{1}{|S_{d,c}|}\sum_{s \in S_{d,c}}b_{h,m}(s)
 \end{equation}

where $s\in S_{d,c}$ denotes a sample $s$ (in our case, an image) of condition $c$ from dataset $d$. Note that this measure can only be zero if the accuracy of $h$ and $m$ are exactly the same in each dataset and condition.

\emph{Error consistency} $E(m)$ \citep{geirhos2020beyond} tracks whether there is above-chance consistency. This is an important distinction, since e.g. two decision makers with 95\% accuracy each will have at least 90\% observed consistency, even if their 5\% errors occur on non-overlapping subsets of the test data (intuitively, they both get most images correct and thus observed overlap is high). To this end, error consistency (a.k.a.\ Cohen's kappa, cf.~\citep{cohen1960coefficient}) indicates whether the observed consistency is larger than what could have been expected given two independent binomial decision makers with matched accuracy, which we denote as $\hat o_{h,m}$. This can easily be computed analytically \citep[e.g.][equation~1]{geirhos2020beyond}. Then, the average error consistency is given by
\begin{equation}
E(m): \mathbb{R}\rightarrow [-1,1], m \mapsto \frac{1}{|D|}\sum_{d \in D }\frac{1}{|H_d|}\sum_{h \in H_d }\frac{1}{|C_d|}\sum_{c \in C_d }\frac{(\frac{1}{|S_{d,c}|}\sum_{s \in S_{d,c}} b_{h,m}(s))-\hat o_{h,m}(S_{d,c})}{1-\hat o_{h,m}(S_{d,c})}
\end{equation}

\paragraph{Benchmark \& toolbox.}
$A(m)$, $O(m)$ and $E(m)$ each quantify a certain aspect of the human-machine gap. We use the mean rank order across these metrics to determine an overall model ranking (Table~\ref{tab:benchmark_table_humanlike} in the Appendix). However, we would like to emphasise that the primary purpose of this benchmark is to generate insights, not winners. Since insights are best gained from detailed plots and analyses, we open-source \texttt{modelvshuman}, a Python project to benchmark models against human data.\footnote{Of course, comparing human and machine vision is not limited to object recognition behaviour: other comparisons may be just as valid and interesting.} The current model zoo already includes 50+ models, and an option to add new ones (both PyTorch and TensorFlow). Evaluating a model produces a 15+ page report on model behaviour. All plots in this paper can be generated for future models---to track whether they narrow the gap towards human vision, or to determine whether an algorithmic modification to a baseline model (e.g., an architectural improvement) changes model behaviour.

\begin{figure}[t!]
	\begin{subfigure}{0.49\linewidth}
			\centering
			\includegraphics[width=\linewidth]{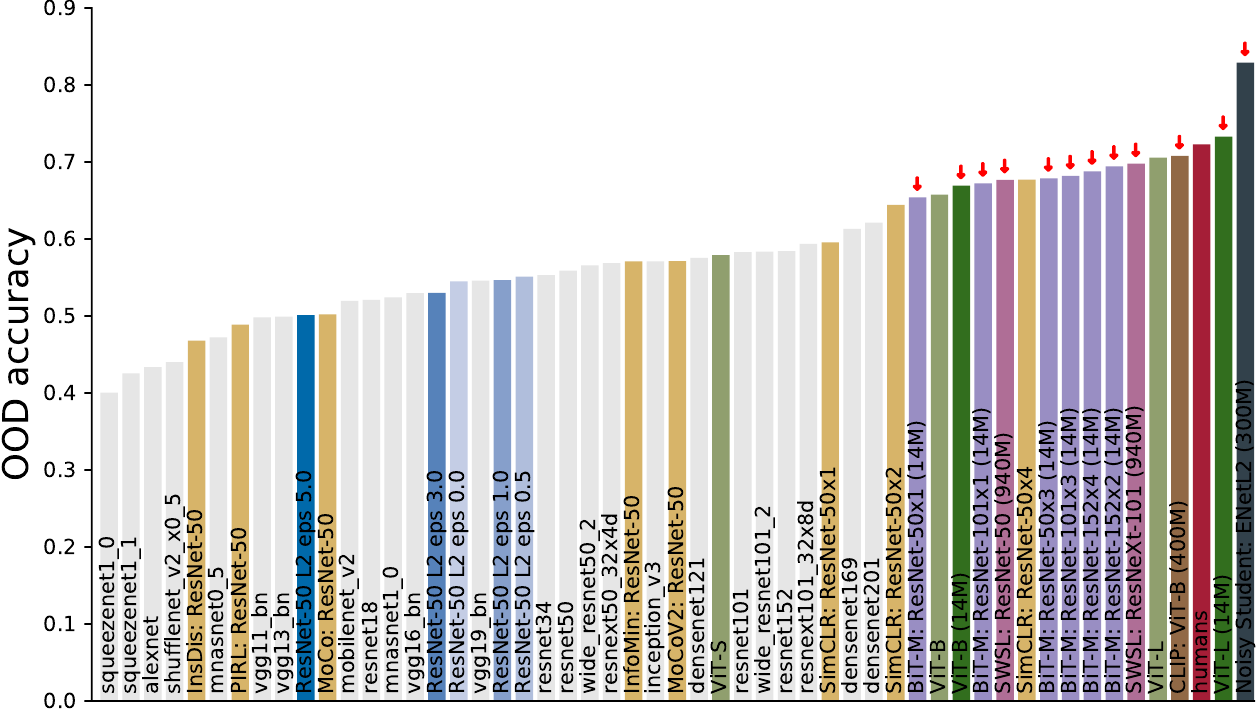}
			\caption{OOD accuracy (higher = better).}
			\label{subfig:benchmark_a}
			\vspace{\captionspaceII}
	\end{subfigure}\hfill
	\begin{subfigure}{0.49\linewidth}
			\centering
			\includegraphics[width=\linewidth]{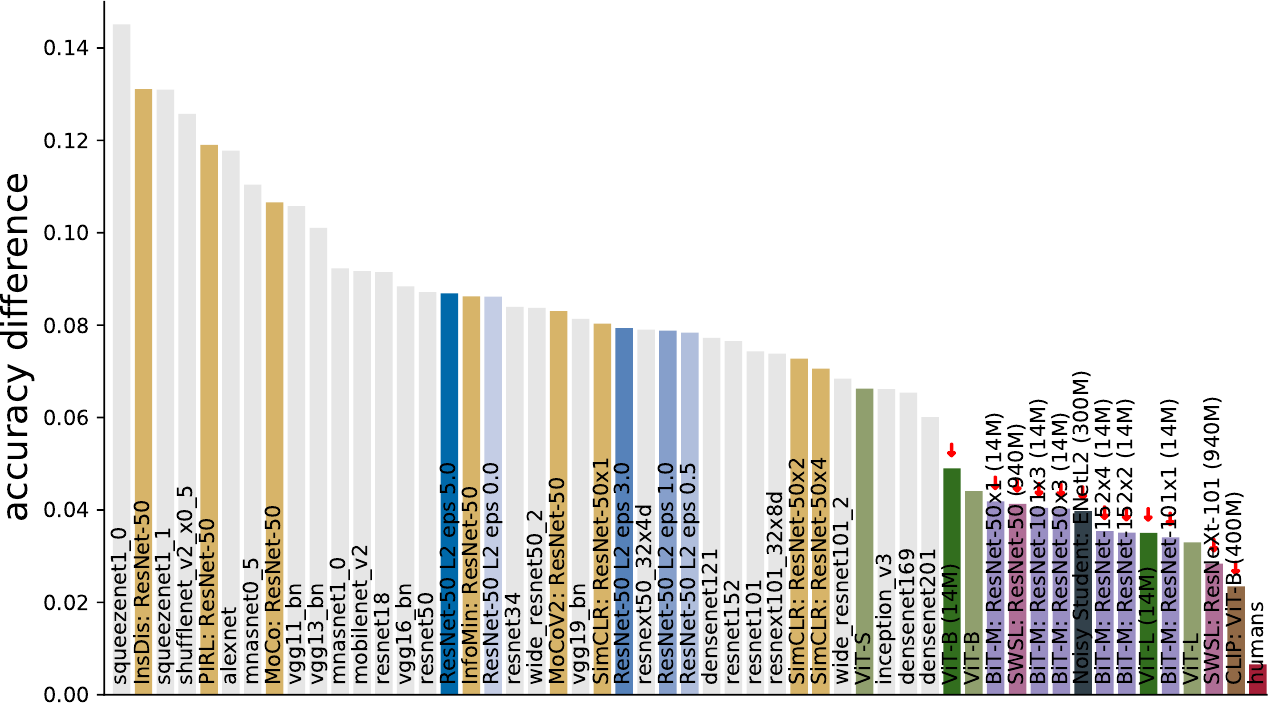}
			\caption{Accuracy difference (lower = better).}
			\label{subfig:benchmark_b}
			\vspace{\captionspaceII}
	\end{subfigure}\hfill
	\begin{subfigure}{0.49\linewidth}
			\centering
			\includegraphics[width=\linewidth]{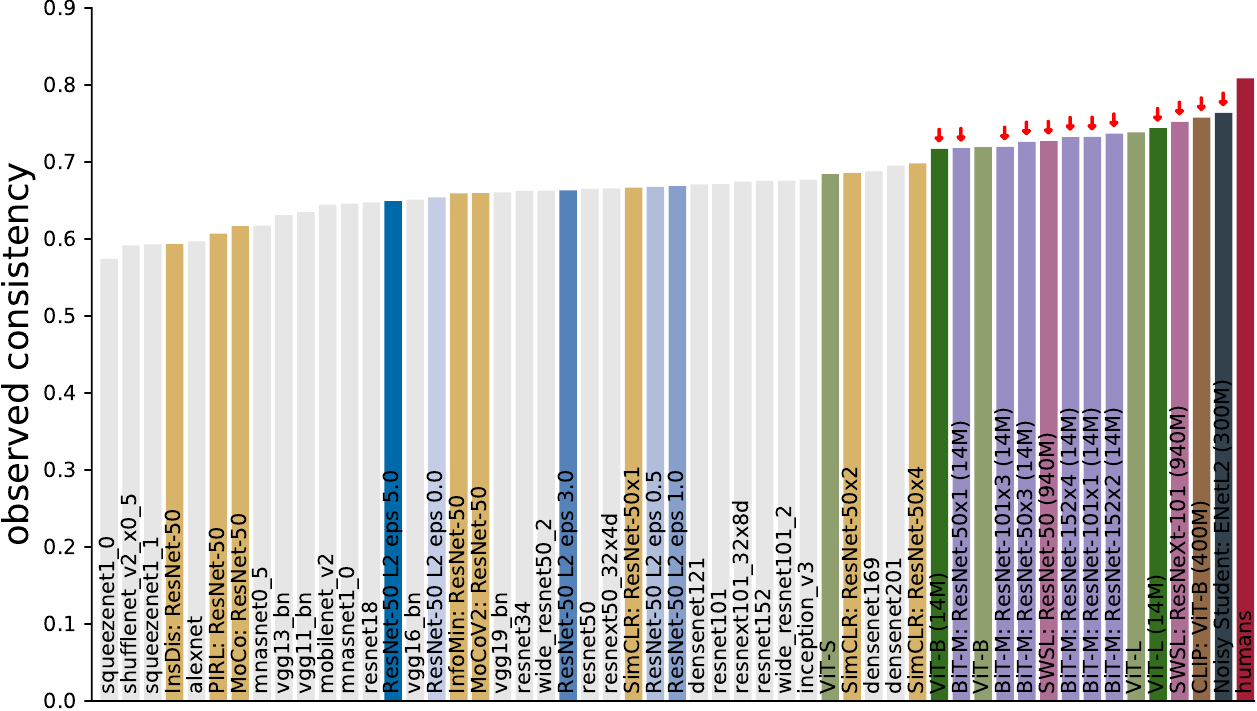}
			\caption{Observed consistency (higher = better).}			\label{subfig:benchmark_c}
			\vspace{\captionspaceII}
	\end{subfigure}\hfill
	\begin{subfigure}{0.49\linewidth}
			\centering
			\includegraphics[width=\linewidth]{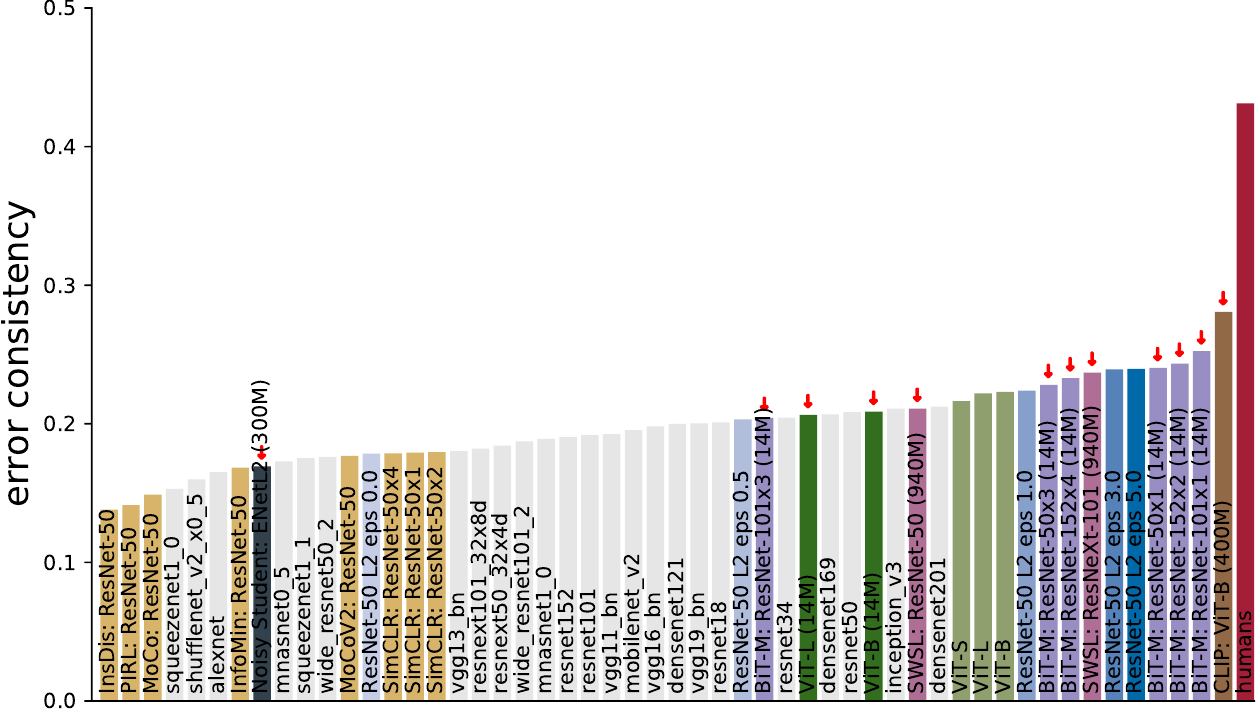}
			\caption{Error consistency (higher = better).}			\label{subfig:benchmark_d}
			\vspace{\captionspaceII}
	\end{subfigure}\hfill
	\caption{Core results, aggregated over 17 out-of-distribution (OOD) datasets: The OOD robustness gap between human and machine vision is closing (top), but an image-level consistency gap remains (bottom). Results compare \textcolor{red}{humans}, \textcolor{supervised.grey}{standard supervised CNNs}, \textcolor{orange1}{self-supervised models}, \textcolor{blue1}{adversarially trained models}, \textcolor{green1}{vision transformers}, \textcolor{black}{noisy student}, \textcolor{bitm.purple2}{BiT}, \textcolor{purple1}{SWSL} and \textcolor{brown1}{CLIP}. For convenience, \textcolor{red.arrow}{$\downarrow$} marks models that are trained on large-scale datasets. Metrics defined in Section~\ref{sec:methods}. Best viewed on screen.}
	\label{fig:benchmark_barplots}
\end{figure}

\section{Robustness across models: the OOD distortion robustness gap between human and machine vision is closing}
\label{sec:robustness_across_models}
We are interested in measuring whether we are making progress in closing the gap between human and machine vision. For a long time, CNNs were unable to match human robustness in terms of generalisation beyond the training distribution---a large OOD \emph{distortion robustness gap} \citep{dodge2017study, wichmann2017methods, geirhos2018generalisation, serre2019deep, geirhos2020shortcut}. Having tested human observers on 17 OOD datasets, we are now able to compare the latest developments in machine vision to human perception.
Our core results are shown in Figure~\ref{fig:benchmark_barplots}: the OOD distortion robustness gap between human and machine vision is closing (\ref{subfig:benchmark_a}, \ref{subfig:benchmark_b}), especially for models trained on large-scale datasets. On the individual image level, a human-machine consistency gap remains (especially \ref{subfig:benchmark_d}), which will be discussed later.

\begin{figure*}[t]
	\begin{subfigure}{\figwidth}
			\centering
			\includegraphics[width=\linewidth]{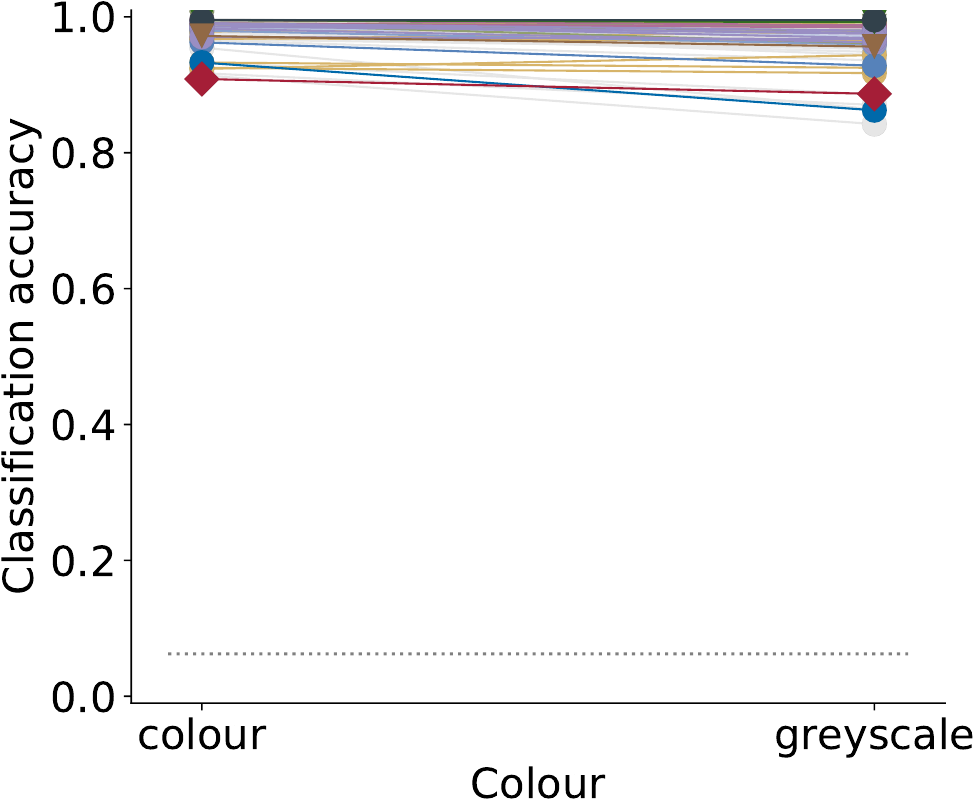}
			\vspace{\captionspace}
			\caption{Colour vs. greyscale}
			\vspace{\captionspaceII}
		\end{subfigure}\hfill
		\begin{subfigure}{\figwidth}
			\centering
			\includegraphics[width=\linewidth]{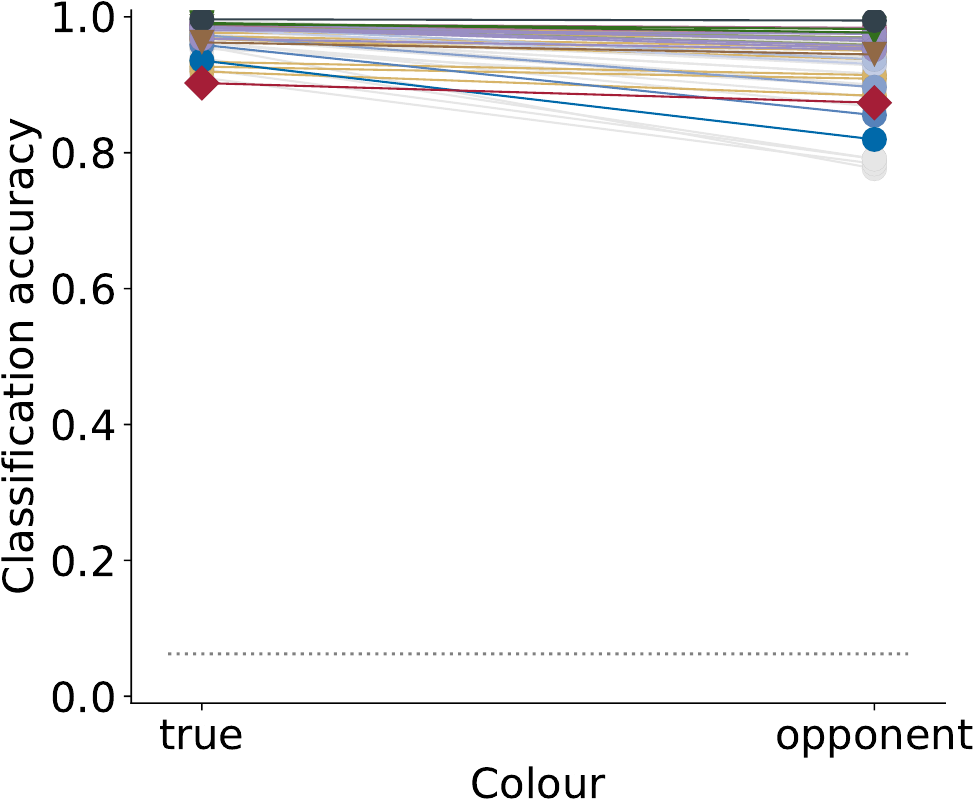}
			\vspace{\captionspace}
			\caption{True vs. false colour}
			\vspace{\captionspaceII}
		\end{subfigure}\hfill
		\begin{subfigure}{\figwidth}
	    	\centering
		    \includegraphics[width=\linewidth]{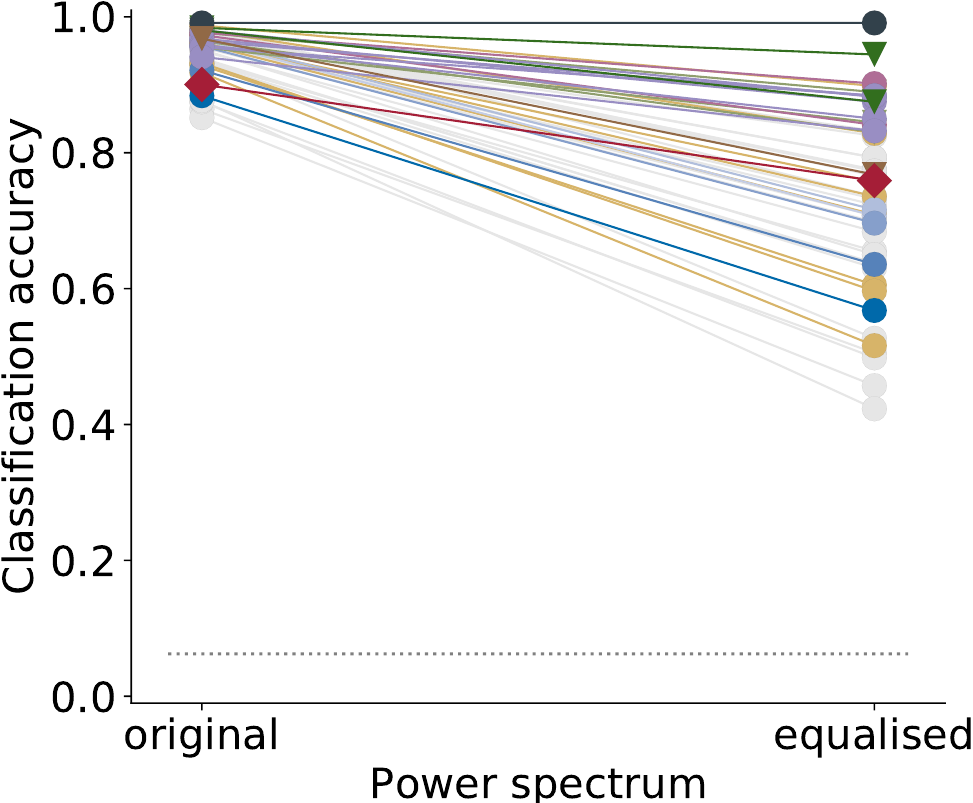}
		    \vspace{\captionspace}
		    \caption{Power equalisation}
		    \vspace{\captionspaceII}
	    \end{subfigure}\hfill
		\begin{subfigure}{\figwidth}
    		\centering
	    	\includegraphics[width=\linewidth]{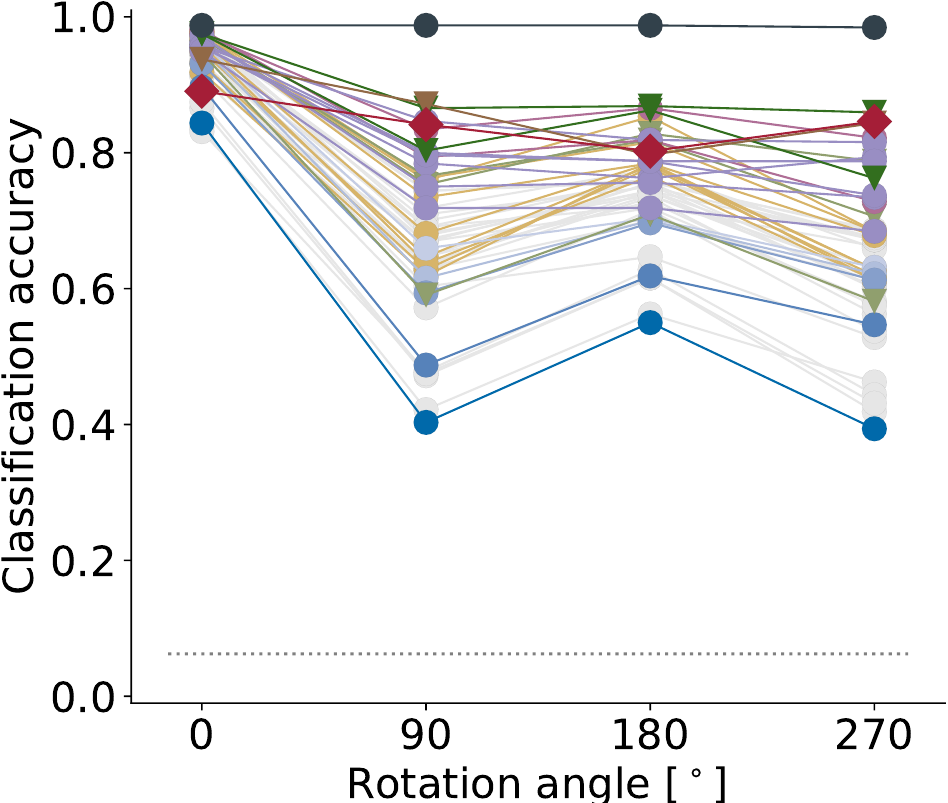}
	    	\vspace{\captionspace}
	    	\caption{Rotation}
	    \end{subfigure}\hfill
	
		\begin{subfigure}{\figwidth}
			\centering
			\includegraphics[width=\linewidth]{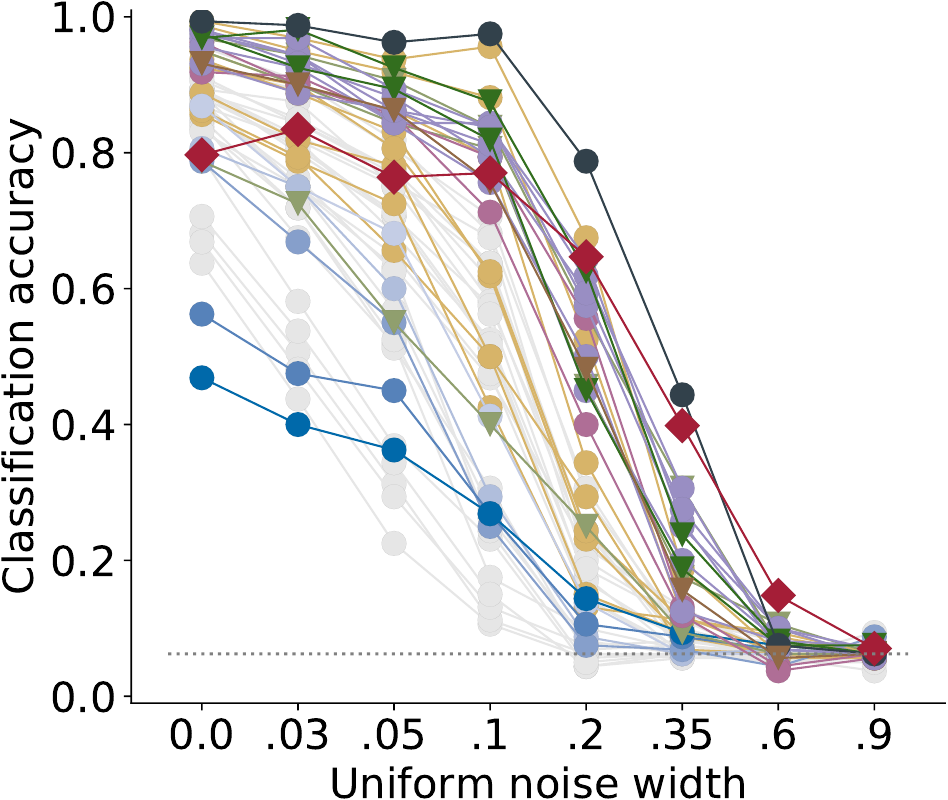}
			\vspace{\captionspace}
			\caption{Uniform noise}
			\vspace{\captionspaceII}
		\end{subfigure}\hfill
		\begin{subfigure}{\figwidth}
			\centering
			\includegraphics[width=\linewidth]{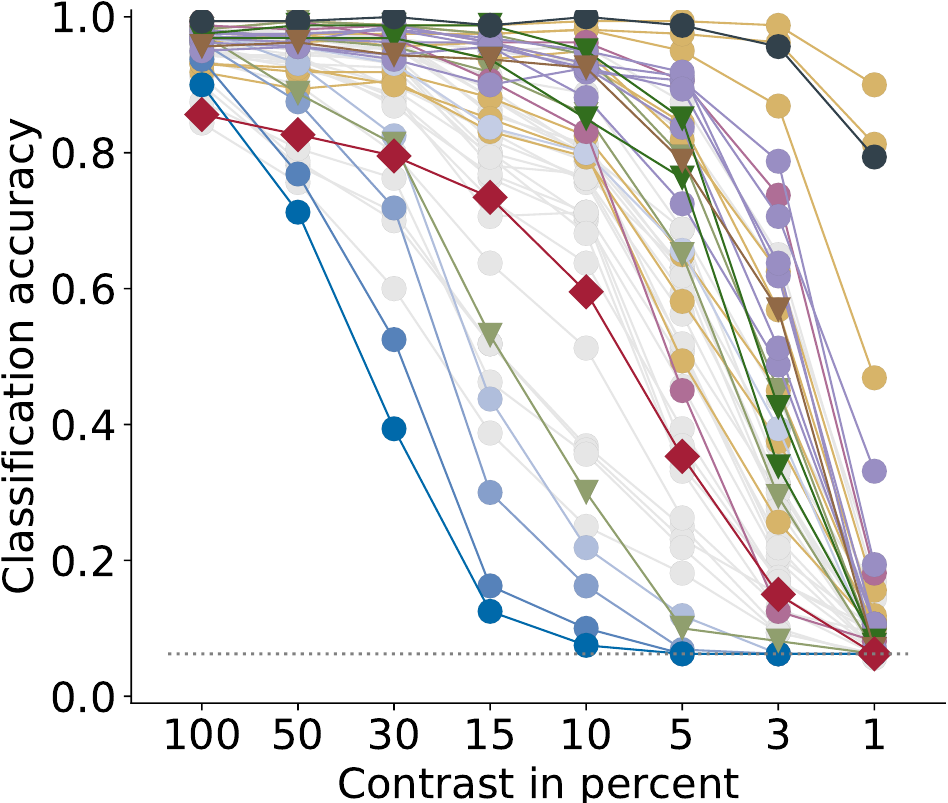}
			\vspace{\captionspace}
			\caption{Contrast}
			\vspace{\captionspaceII}
		\end{subfigure}\hfill
		\begin{subfigure}{\figwidth}
			\centering
			\includegraphics[width=\linewidth]{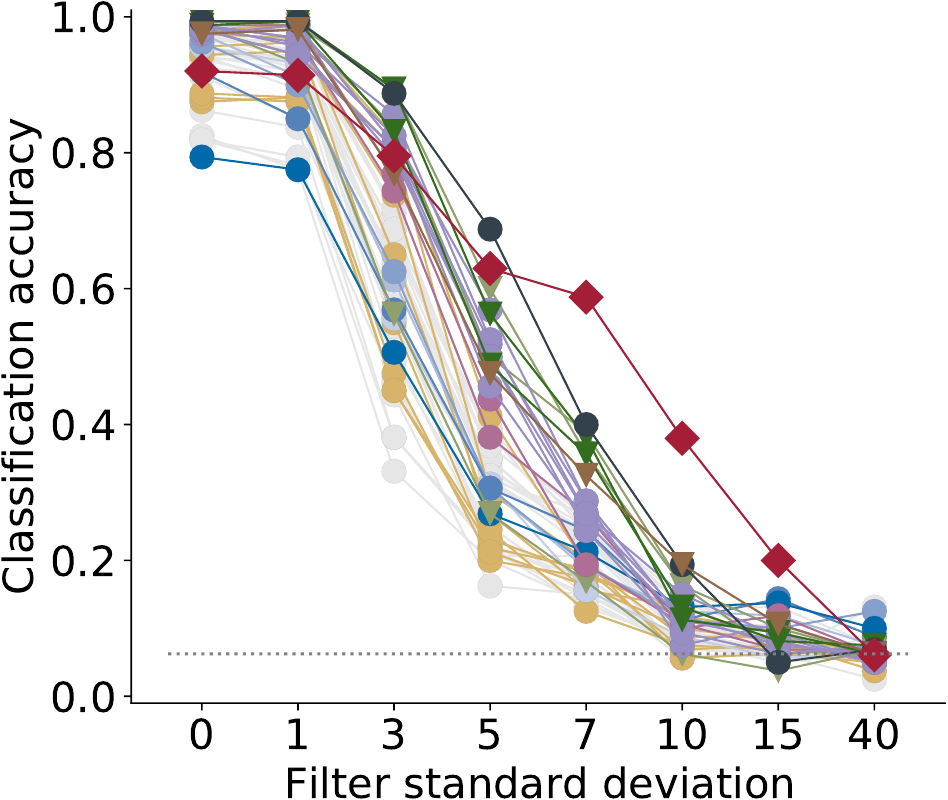}
			\vspace{\captionspace}
			\caption{Low-pass}
			\vspace{\captionspaceII}
		\end{subfigure}\hfill
		\begin{subfigure}{\figwidth}
			\centering
			\includegraphics[width=\linewidth]{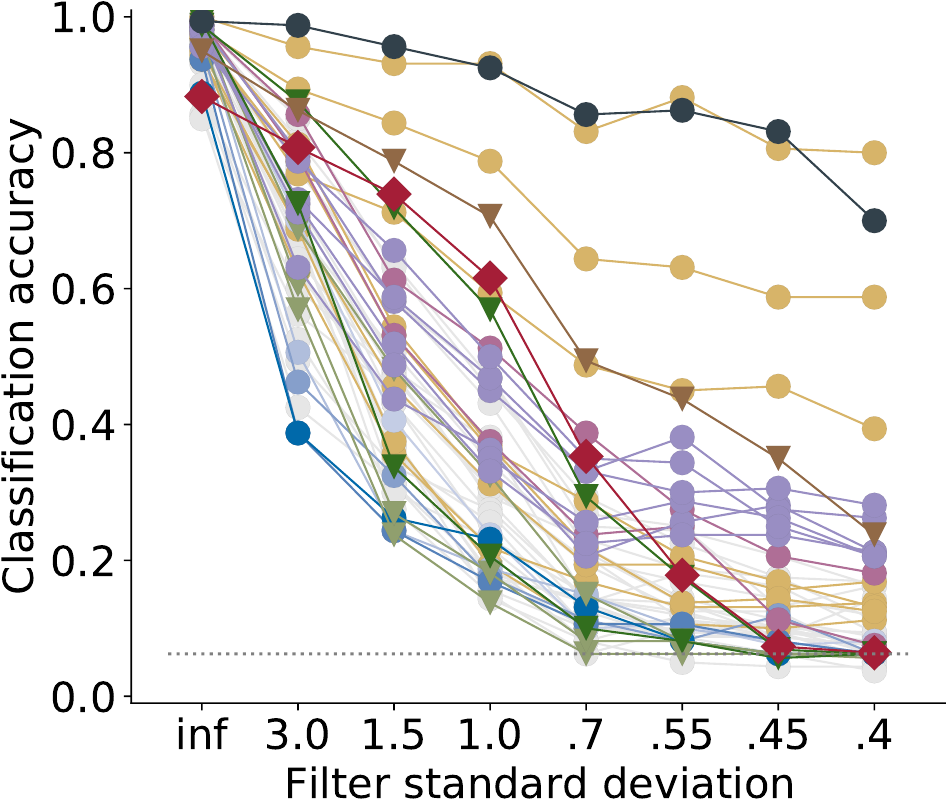}
			\vspace{\captionspace}
			\caption{High-pass}
			\vspace{\captionspaceII}
		\end{subfigure}\hfill
		
		\begin{subfigure}{\figwidth}
			\centering
			\includegraphics[width=\linewidth]{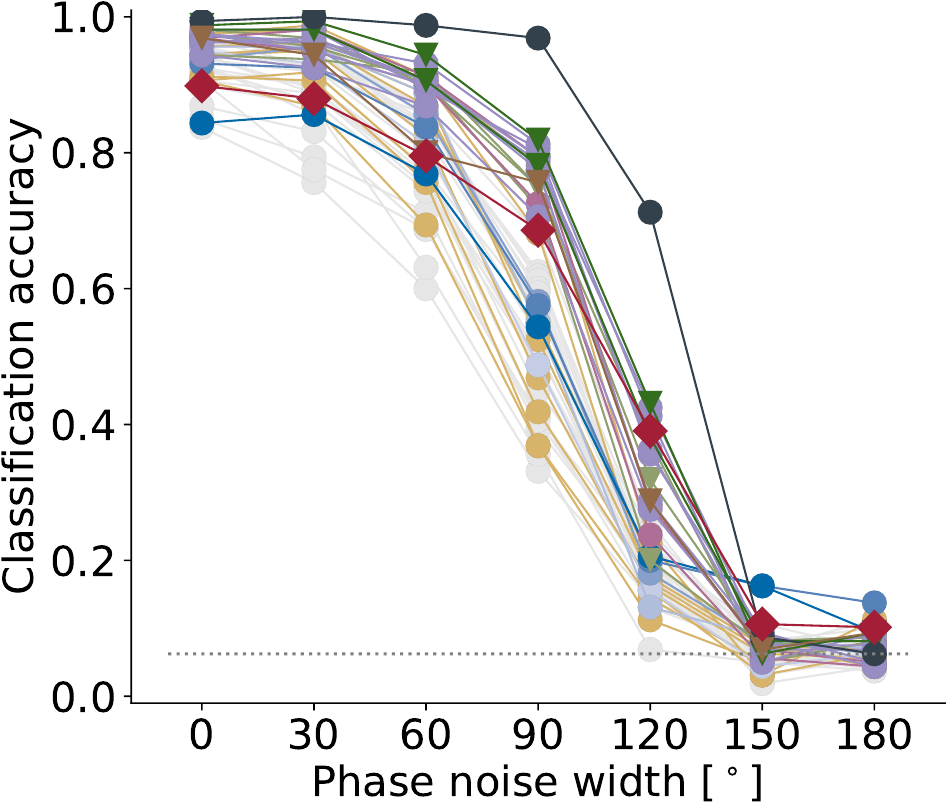}
			\vspace{\captionspace}
			\caption{Phase noise}
			\vspace{\captionspaceII}
		\end{subfigure}\hfill
		\begin{subfigure}{\figwidth}
			\centering
			\includegraphics[width=\linewidth]{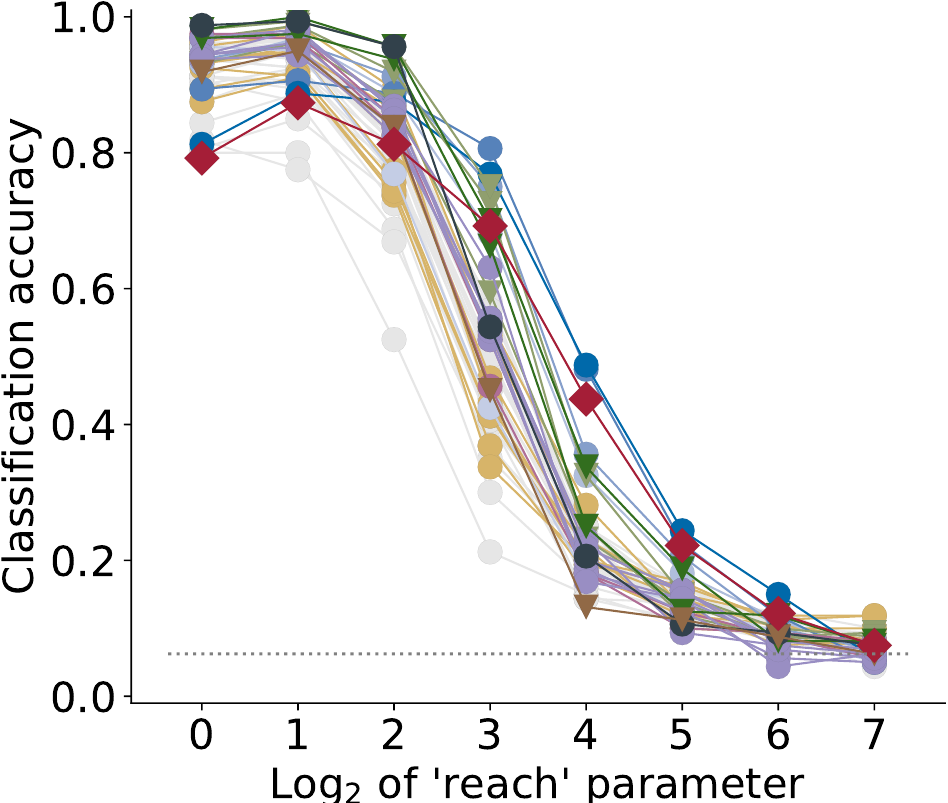}
			\vspace{\captionspace}
			\caption{Eidolon I}
			\vspace{\captionspaceII}
		\end{subfigure}\hfill
    	\begin{subfigure}{\figwidth}
	    	\centering
	    	\includegraphics[width=\linewidth]{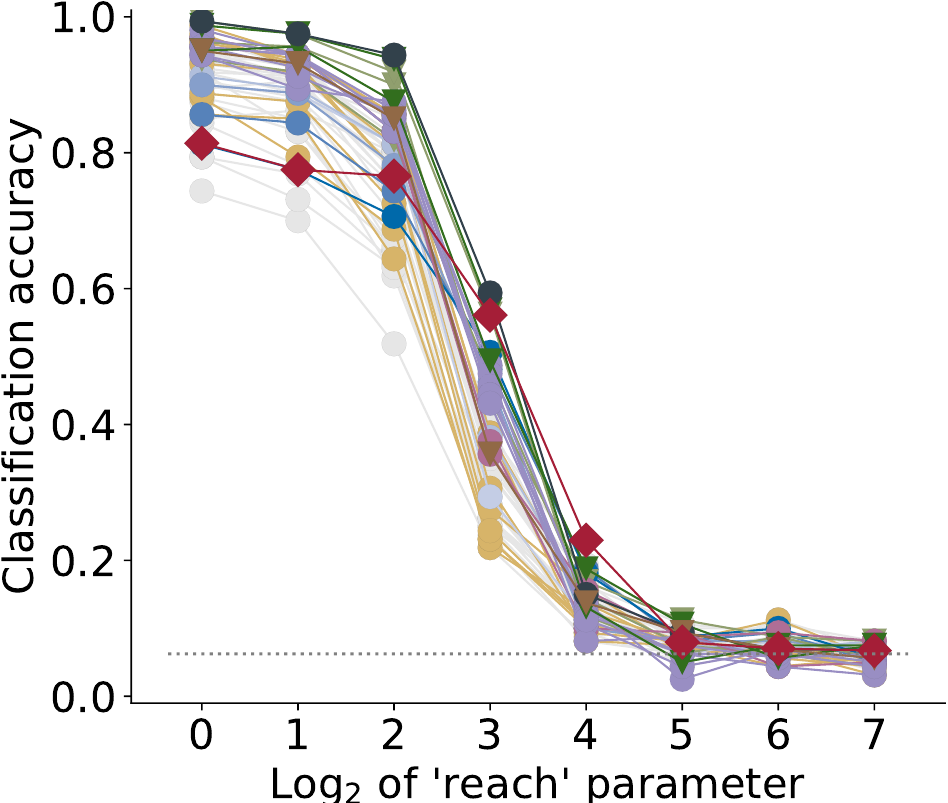}
		    \vspace{\captionspace}
		    \caption{Eidolon II}
		    \vspace{\captionspaceII}
	    \end{subfigure}\hfill
    	\begin{subfigure}{\figwidth}
	    	\centering
		    \includegraphics[width=\linewidth]{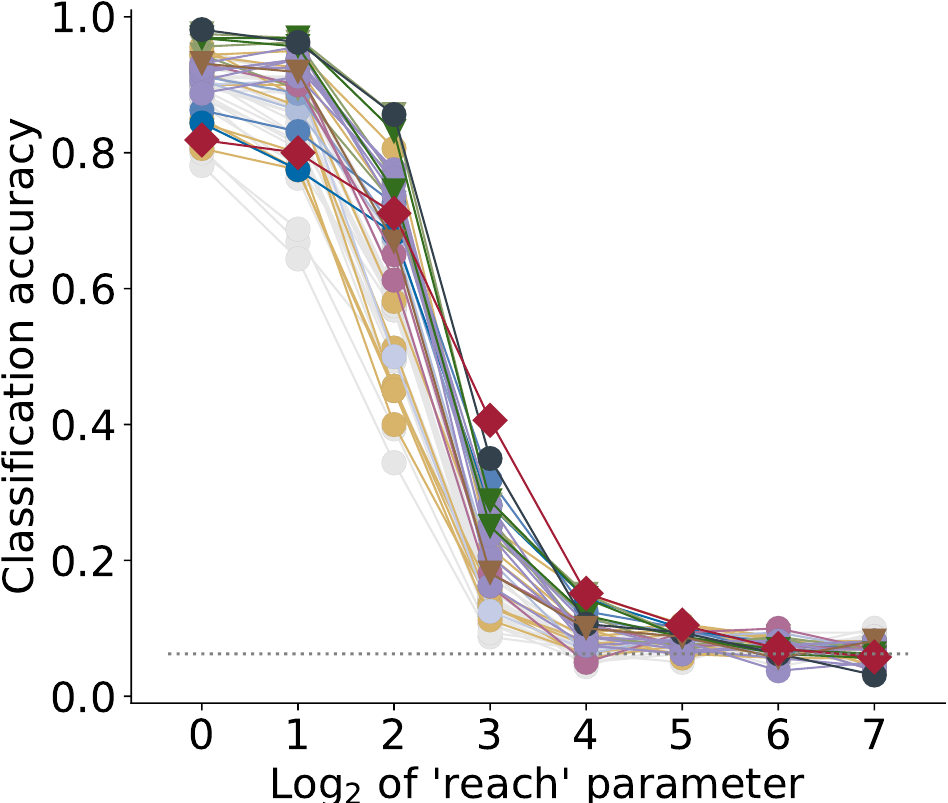}
    		\vspace{\captionspace}
	    	\caption{Eidolon III}
	    \end{subfigure}\hfill
	\caption{The OOD distortion robustness gap between human and machine vision is closing. Robustness towards parametric distortions for \textcolor{red}{humans}, \textcolor{supervised.grey}{standard supervised CNNs}, \textcolor{orange1}{self-supervised models}, \textcolor{blue1}{adversarially trained models}, \textcolor{green1}{vision transformers}, \textcolor{black}{noisy student}, \textcolor{bitm.purple2}{BiT}, \textcolor{purple1}{SWSL}, \textcolor{brown1}{CLIP}. Symbols indicate architecture type ($\ocircle$ convolutional, $\triangledown$ vision transformer, $\lozenge$ human); best viewed on screen.}
	\label{fig:results_accuracy}
\end{figure*}

\paragraph{\textcolor{orange1}{Self-supervised models}}
\label{subsec:self_supervised}
\textit{``If intelligence is a cake, the bulk of the cake is unsupervised learning, the icing on the cake is supervised learning and the cherry on the cake is reinforcement learning''}, Yann LeCun said in 2016 \citep{lecun2016unsupervised}. A few years later, the entire cake is finally on the table---the representations learned via self-supervised learning\footnote{``Unsupervised learning'' and ``self-supervised learning'' are sometimes used interchangeably. We use the term ``self-supervised learning''' since those methods use (label-free) supervision.} now compete with supervised methods on ImageNet \citep{chen2020simple} and outperform supervised pre-training for object detection \citep{misra2020self}. But how do recent self-supervised models differ from their supervised counterparts in terms of their behaviour? Do they bring machine vision closer to human vision? Humans, too, rapidly learn to recognise new objects without requiring hundreds of labels per instance; additionally a number of studies reported increased similarities between self-supervised models and human perception \citep{lotter2020neural, orhan2020self, konkle2020instance, zhuang2020unsupervised, storrs2021unsupervised}. Figure~\ref{fig:results_accuracy} compares the generalisation behaviour of eight self-supervised models in \textcolor{orange1}{orange} (PIRL, MoCo, MoCoV2, InfoMin, InsDis, SimCLR-x1, SimCLR-x2, SimCLR-x4)---with 24 standard supervised models (\textcolor{supervised.grey}{grey}). We find only marginal differences between self-supervised and supervised models: Across distortion types, self-supervised networks are well within the range of their poorly generalising supervised counterparts. However, there is one exception: the three SimCLR variants show strong generalisation improvements on uniform noise, low contrast, and high-pass images, where they are the three top-performing self-supervised networks---quite remarkable given that SimCLR models were trained on a different set of augmentations (random crop with flip and resize, colour distortion, and Gaussian blur). Curious by the outstanding performance of SimCLR, we asked whether the self-supervised objective function or the choice of training data augmentations was the defining factor. When comparing self-supervised SimCLR models with augmentation-matched baseline models trained in the standard supervised fashion (Figure~\ref{fig:results_accuracy_entropy_simclr_baseline} in the Appendix), we find that the augmentation scheme (rather than the self-supervised objective) indeed made the crucial difference: supervised baselines show just the same generalisation behaviour, a finding that fits well with \citep{hermann2020the}, who observed that the influence of training data augmentations is stronger than the role of architecture or training objective. In conclusion, our analyses indicate that the ``cake'' of contrastive self-supervised learning currently (and disappointingly) tastes much like the ``icing''.

\begin{figure}%
    \centering
    \includegraphics[width=\linewidth]{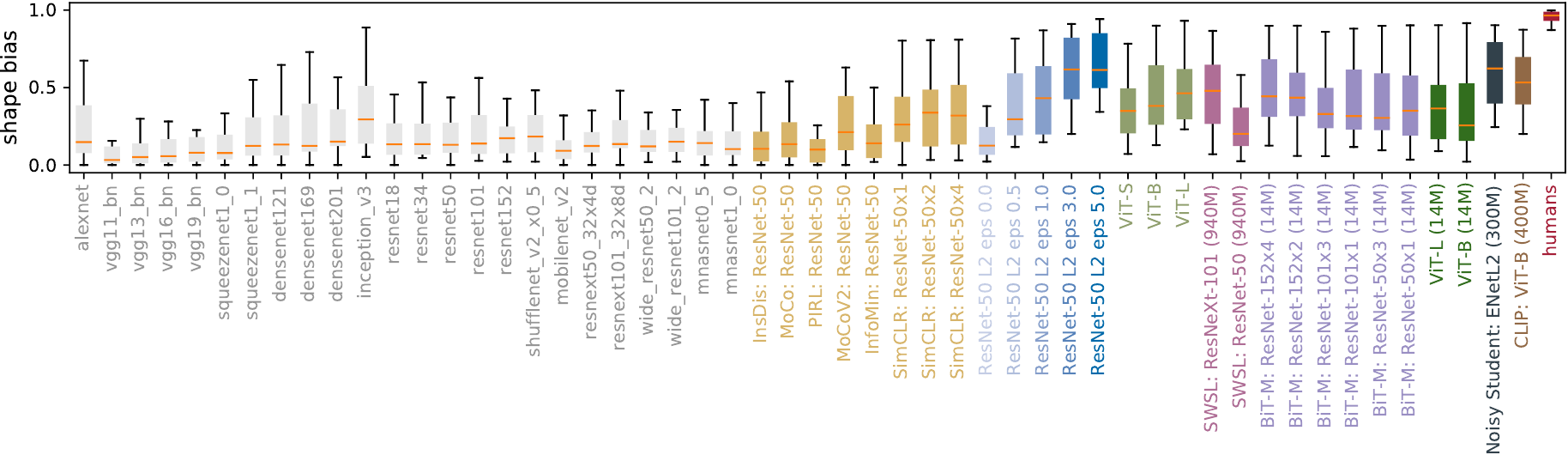}
    \caption{Shape vs.\ texture biases of different models. While human shape bias is not yet matched, several approaches improve over vanilla CNNs. Box plots show category-dependent distribution of shape / texture biases (shape bias: high values, texture bias: low values).}
    \label{fig:shape_bias}
\end{figure}

\paragraph{\textcolor{blue1}{Adversarially trained models}}
\label{subsec:adversarially_trained}
The vulnerability of CNNs to adversarial input perturbations is, arguably, one of the most striking shortcomings of this model class compared to robust human perception. A successful method to increase adversarial robustness is \emph{adversarial training} \citep[e.g.][]{goodfellow2014explaining, huang2015learning}. The resulting models were found to transfer better, have meaningful gradients \citep{kaur2019perceptually}, and enable interpolating between two input images \citep{santurkar2019computer}: ``robust optimization can actually be viewed as inducing a \textit{human prior} over the features that models are able to learn'' \citep[p.\ 10]{engstrom2019adversarial}. Therefore, we include five models with a ResNet-50 architecture and different accuracy-robustness tradeoffs, adversarially trained on ImageNet with Microsoft-scale resources by \citep{salman2020adversarially} to test whether models with ``perceptually-aligned representations'' also show human-aligned OOD generalisation behaviour---as we would hope. This is not the case: the stronger the model is trained adversarially (darker shades of \textcolor{blue1}{blue} in Figure~\ref{fig:results_accuracy}), the more susceptible it becomes to (random) image degradations. Most strikingly, a simple rotation by 90 degrees leads to a 50\% drop in classification accuracy. Adversarial robustness seems to come at the cost of increased vulnerability to large-scale perturbations.\footnote{This might be related to \citep{tramer2020fundamental}, who studied a potentially related tradeoff between selectivity and invariance.} On the other hand, there is a silver lining: when testing whether models are biased towards texture or shape by testing them on cue conflict images (Figure~\ref{fig:shape_bias}), in accordance with \citep{zhang2019interpreting, chen2020shape} we observe a perfect relationship between shape bias and the degree of adversarial training, a big step in the direction of human shape bias (and a stronger shape bias than nearly all other models).

\vspace{-0.27cm}
\paragraph{\textcolor{green1}{Vision transformers}}
\label{subsec:vision_transformers}
In computer vision, convolutional networks have become by far the dominating model class over the last decade. Vision transformers \citep{dosovitskiy2020image} break with the long tradition of using convolutions and are rapidly gaining traction \citep{han2020survey}. We find that the best vision transformer (ViT-L trained on 14M images) even \emph{exceeds} human OOD accuracy (Figure~\ref{subfig:benchmark_a} shows the average across 17 datasets). There appears to be an additive effect of architecture and data: vision transformers trained on 1M images (light green) are already better than standard convolutional models; training on 14M images (dark green) gives another performance boost. In line with \citep{naseer2021intriguing, tuli2021convolutional}, we observe a higher shape bias compared to most standard CNNs.

\paragraph{Standard models trained on more data: \textcolor{bitm.purple2}{BiT-M}, \textcolor{purple1}{SWSL}, \textcolor{black}{Noisy Student}}
\label{subsec:big_data_models}
Interestingly, the biggest effect on OOD robustness we find simply comes from training on larger datasets, not from advanced architectures. When standard models are combined with large-scale training (14M images for BiT-M, 300M for Noisy Student and a remarkable 940M for SWSL), OOD accuracies reach levels not known from standard ImageNet-trained models; these models even outperform a more powerful architecture (vision transformer ViT-S) trained on less data (1M) as shown in Figure~\ref{subfig:benchmark_a}. Simply training on (substantially) more data substantially narrows the gap to human OOD accuracies (\ref{subfig:benchmark_b}), a finding that we quantified in Appendix~\ref{app:regression_model} by means of a regression model. (The regression model also revealed a significant interaction between dataset size and objective function, as well as a significant main effect for transformers over CNNs.) Noisy Student in particular outperforms humans by a large margin overall (Figure~\ref{subfig:benchmark_a})---the beginning of a new human-machine gap, this time in favour of machines?

\paragraph{\textcolor{brown1}{CLIP}}
\label{subsec:clip}
CLIP is special: trained on 400M images\footnote{The boundary between IID and OOD data is blurry for networks trained on big proprietary datasets. We consider it unlikely that CLIP was exposed to many of the exact distortions used here (e.g.\ eidolon or cue conflict images), but CLIP likely had greater exposure to some conditions such as grayscale or low-contrast images.} (more data) with joint language-image supervision (novel objective) and a vision transformer backbone (non-standard architecture), it scores close to humans across all of our metrics presented in Figure~\ref{fig:benchmark_barplots}; most strikingly in terms of error consistency (which will be discussed in the next section). We tested a number of hypotheses to disentangle why CLIP appears ``special''. \textit{H1: because CLIP is trained on a lot of data?} Presumably no: Noisy Student---a model trained on a comparably large dataset of 300M images---performs very well on OOD accuracy, but poorly on error consistency. A caveat in this comparison is the quality of the labels: while Noisy Student uses pseudolabeling, CLIP receives web-based labels for all images. \textit{H2: because CLIP receives higher-quality labels?} About 6\% of ImageNet labels are plainly wrong \citep{northcutt2021pervasive}. Could it be the case that CLIP simply performs better since it doesn't suffer from this issue? In order to test this, we used CLIP to generate new labels for all 1.3M ImageNet images: (a) hard labels, i.e.\ the top-1 class predicted by CLIP; and (b) soft labels, i.e.\ using CLIP's full posterior distribution as a target. We then trained ResNet-50 from scratch on CLIP hard and soft labels (for details see Appendix~\ref{app:training_clip_labels}). However, this does not show any robustness improvements over a vanilla ImageNet-trained ResNet-50, thus different/better labels are not a likely root cause. \textit{H3:~because CLIP has a special image+text loss?} Yes and no: CLIP training on ResNet-50 leads to astonishingly poor OOD results, so training a standard model with CLIP loss alone is insufficient. However, while neither architecture nor loss alone sufficiently explain why CLIP is special, we find a clear interaction between architecture and loss (described in more detail in the Appendix along with the other ``CLIP ablation'' experiments mentioned above).

\begin{figure*}[ht]
    \centering
    \includegraphics[width=0.94\linewidth]{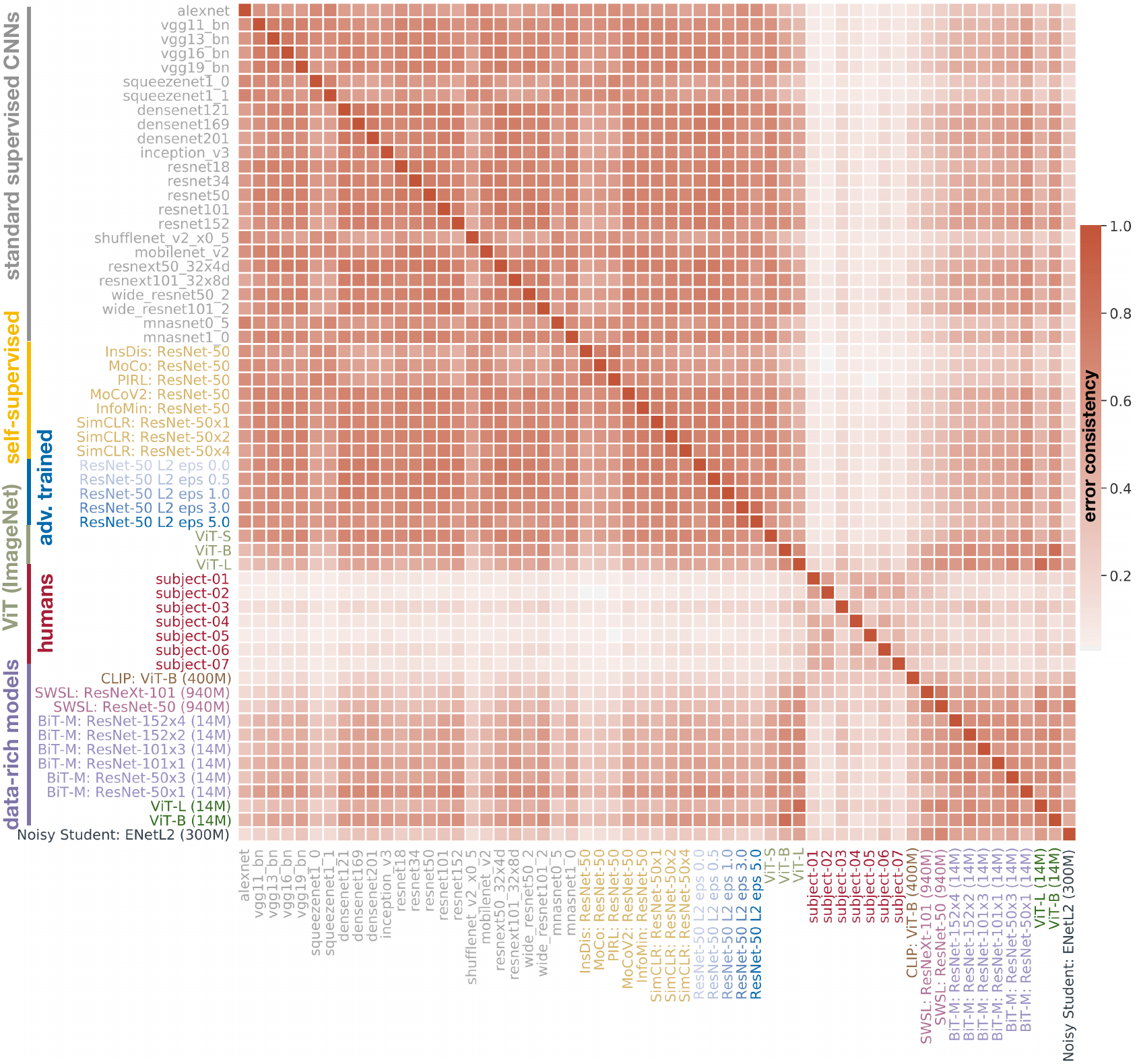}
    \vspace{-0.2cm}
    \caption{Data-rich models narrow the substantial image-level consistency gap between humans and machines. Error consistency analysis on a single dataset (sketch images; for other datasets see Appendix, Figures~\ref{fig:results_accuracy_error_consistency}, \ref{fig:error_consistency_matrix_stylized_by_mean}, \ref{fig:error_consistency_matrix_edges_by_mean}, \ref{fig:error_consistency_matrix_silhouettes_by_mean}, \ref{fig:error_consistency_matrix_cue-conflict_by_mean}) shows that most models cluster (dark red = highly consistent errors) irrespective of their architecture and objective function; humans cluster differently (high human-to-human consistency, low human-to-model consistency); but some data-rich models including \textcolor{brown1}{CLIP} and \textcolor{purple1}{SWSL} blur the boundary, making more human-like errors than standard models.}
    \label{fig:error_consistency_matrix_sketch}
\end{figure*}

\section{Consistency between models: data-rich models narrow the substantial image-level consistency gap between human and machine vision}
\label{sec:consistency_between_models}

In the previous section we have seen that while self-supervised and adversarially trained models lack OOD distortion robustness, models based on vision transformers and/or trained on large-scale datasets now match or exceed human feedforward performance on most datasets. Behaviourally, a natural follow-up question is to ask not just how many, but \emph{which} errors models make---i.e., do they make errors on the same individual images as humans on OOD data (an important characteristic of a ``human-like'' model, cf.~\citep{rajalingham2018large, geirhos2020beyond})? This is quantified via \emph{error consistency} (defined in Section~\ref{sec:methods}); which additionally allows us to compare models with each other, asking e.g.\ which model classes make similar errors. In Figure~\ref{fig:error_consistency_matrix_sketch}, we compare all models with each other and with humans, asking whether they make errors on the same images. On this particular dataset (sketch images), we can see one big model cluster. Irrespective of whether one takes a standard supervised model, a self-supervised model, an adversarially trained model or a vision transformer, all those models make highly systematic errors (which extends the results of \citep{mania2019model, geirhos2020beyond} who found similarities between standard vanilla CNNs). Humans, on the other hand, show a very different pattern of errors. Interestingly, the boundary between humans and some data-rich models at the bottom of the figure---especially CLIP (400M images) and SWSL (940M)---is blurry: some (but not all) data-rich models much more closely mirror the patterns of errors that humans make, and we identified the first models to achieve higher error consistency with humans than with other (standard) models. Are these promising results shared across datasets, beyond the sketch images? In Figures~\ref{subfig:benchmark_c} and \ref{subfig:benchmark_d}, aggregated results over 17 datasets are presented. Here, we can see that data-rich models approach human-to-human observed consistency, but not error consistency. Taken in isolation, \emph{observed} consistency is not a good measure of image-level consistency since it does not take consistency by chance into account; \emph{error} consistency tracks whether there is consistency beyond chance; here we see that there is still a substantial image-level \emph{consistency gap} between human and machine vision. However, several models improve over vanilla CNNs, especially BiT-M (trained on 14M images) and CLIP (400M images). This progress is non-trivial; at the same time, there is ample room for future improvement.

\begin{figure}[t]
    \centering
    \begin{subfigure}{0.45\textwidth}
        \centering
        \includegraphics[width=\linewidth]{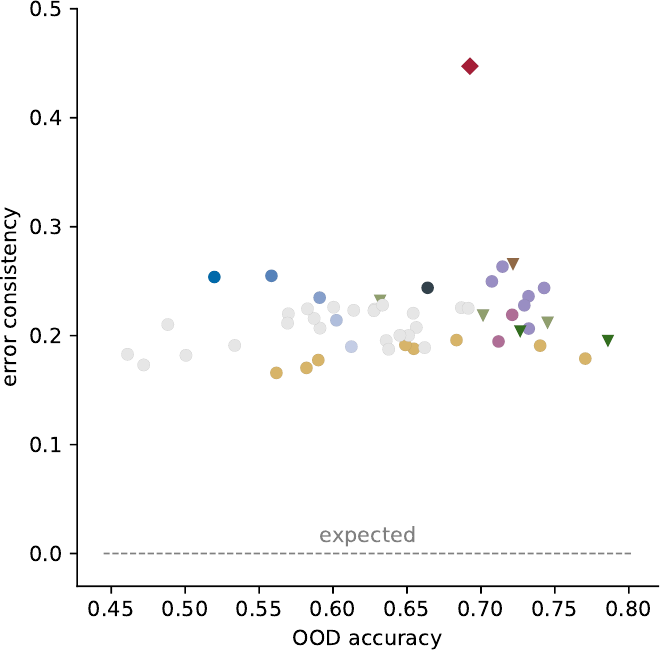}
        \caption{12 ``disappointing'' datasets}
        \label{subfig:error_consistency_12_datasets}
    \end{subfigure}
    \begin{subfigure}{0.45\textwidth}
        \centering
        \includegraphics[width=\linewidth]{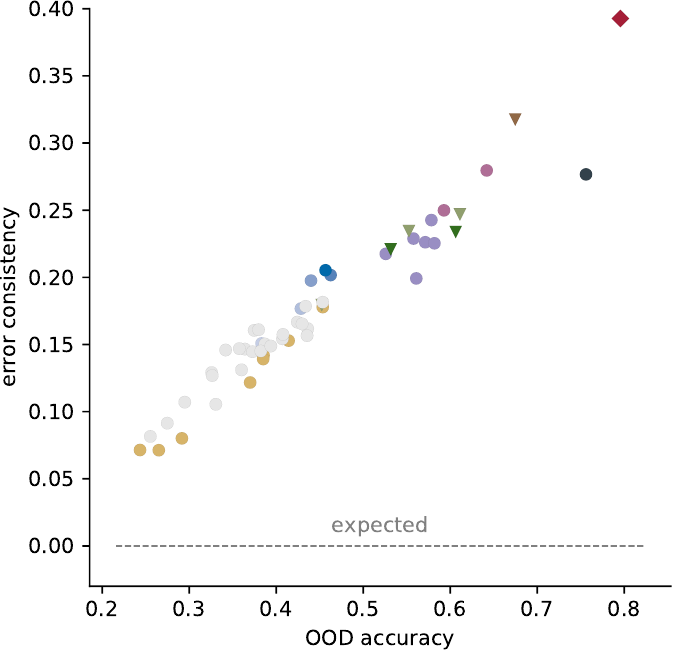}
        \caption{5 ``promising'' datasets}
        \label{subfig:error_consistency_5_datasets}
    \end{subfigure}
    \label{fig:error_consistency_12_and_5_datasets}
    \vspace{-0.1cm}
    \caption{Partial failure, partial success: Error consistency with humans aggregated over multiple datasets. Left: 12 datasets where model accuracies exceed human accuracies; here, there is still a substantial image-level consistency gap to humans. Right: 5 datasets (sketch, silhouette, edge, cue conflict, low-pass) where humans are more robust. Here, OOD accuracy is a near-perfect predictor of image-level consistency; especially data-rich models (e.g.\ \textcolor{brown1}{CLIP}, \textcolor{purple1}{SWSL}, \textcolor{bitm.purple2}{BiT}) narrow the consistency gap to humans. Symbols indicate architecture type ($\ocircle$ convolutional, $\triangledown$ vision transformer, $\lozenge$ human).}
\end{figure}

How do the findings from Figure~\ref{fig:error_consistency_matrix_sketch} (showing nearly human-level error consistency for sketch images) and from Figure~\ref{subfig:benchmark_d} (showing a substantial consistency gap when aggregating over 17 datasets) fit together? Upon closer inspection, we discovered that there are two distinct cases. On 12 datasets (stylized, colour/greyscale, contrast, high-pass, phase-scrambling, power-equalisation, false colour, rotation, eidolonI, -II and -III as well as uniform noise), the human-machine gap is large; here, more robust models do not show improved error consistency (as can be see in Figure~\ref{subfig:error_consistency_12_datasets}). On the other hand, for five datasets (sketch, silhouette, edge, cue conflict, low-pass filtering), there is a completely different result pattern: Here, OOD accuracy is a near-perfect predictor of error consistency, which means that improved generalisation robustness leads to more human-like errors (Figure~\ref{subfig:error_consistency_5_datasets}). Furthermore, training on large-scale datasets leads to considerable improvements along both axes for standard CNNs. Within models trained on larger datasets, CLIP scores best; but models with a standard architecture (SWSL: based on ResNet-50 and ResNeXt-101) closely follow suit.

It remains an open question why the training dataset appears to have the most important impact on a model's decision boundary as measured by error consistency (as opposed to other aspects of a model's inductive bias). Datasets contain various shortcut opportunities \citep{geirhos2020shortcut}, and if two different models are trained on similar data, they might converge to a similar solution simply by exploiting the same shortcuts---which would also fit well to the finding that adversarial examples typically transfer very well between different models \citep{szegedy2013intriguing, tramer2017space}. Making models more flexible (such as transformers, a generalisation of CNNs) wouldn't change much in this regard, since flexible models can still exploit the same shortcuts. Two predictions immediately follow from this hypothesis: (1.) error consistency between two identical models trained on very different datasets, such as ImageNet vs.\ Stylized-ImageNet, is much lower than error consistency between very different models (ResNet-50 vs.\ VGG-16) trained on the same dataset. (2.) error consistency between ResNet-50 and a highly flexible model (e.g., a vision transformer) is much higher than error consistency between ResNet-50 and a highly constrained model like BagNet-9 \citep{brendel2019approximating}. We provide evidence for both predictions in Appendix~\ref{app:error_consistency_predictions}, which makes the shortcut hypothesis of model similarity a potential starting point for future analyses. Looking forward, it may be worth exploring the links between shortcut learning and image difficulty, such as understanding whether many ``trivially easy'' images in common datasets like ImageNet causes models to expoit the same characteristics irrespective of their architecture \citep{meding2021trivial}.

\section{Discussion}
\label{sec:outlook}
\vspace{-0.2cm}

\paragraph{Summary} We set out to answer the question: \emph{Are we making progress in closing the gap between human and machine vision?} In order to quantify progress, we performed large-scale psychophysical experiments on 17 out-of-distribution distortion datasets (open-sourced along with evaluation code as a benchmark to track future progress). We then investigated models that push the boundaries of traditional deep learning (different objective functions, architectures, and dataset sizes ranging from 1M to 1B), asking how they perform relative to human visual perception. We found that the OOD distortion robustness gap between human and machine vision is closing, as the best models now match or exceed human accuracies. At the same time, an image-level consistency gap remains; however, this gap that is at least in some cases narrowing for models trained on large-scale datasets.

\paragraph{Limitations} Model robustness is studied from many different viewpoints, including adversarial robustness \citep{szegedy2013intriguing}, theoretical robustness guarantees \citep[e.g.][]{hein2017formal}, or label noise robustness \citep[e.g.][]{vahdat2017toward}. The focus of our study is robustness towards non-adversarial out-of-distribution data, which is particularly well-suited for comparisons with humans. Since we aimed at a maximally fair comparison between feedforward models and human perception, presentation times for human observers were limited to 200 ms in order to limit the influence of recurrent processing. Therefore, human ceiling performance might be higher still (given more time); investigating this would mean going beyond ``core object recognition'', which happens within less than 200 ms during a single fixation \citep{DiCarlo2012}. Furthermore, human and machine vision can be compared in many different ways. This includes comparing against neural data \cite{Yamins2014, kubilius2019brain}, contrasting Gestalt effects \citep[e.g.][]{kim2021neural}, object similarity judgments \citep{hebart2020revealing}, or mid-level properties \citep{storrs2021unsupervised} and is of course not limited to studying object recognition. By no means do we mean to imply that our behavioural comparison is the only feasible option---on the contrary, we believe it will be all the more exciting to investigate whether our behavioural findings have implications for other means of comparison!

\paragraph{Discussion} We have to admit that we view our results concerning the benefits of increasing dataset size by one-to-three orders of magnitude with mixed feelings. On the one hand, ``simply'' training standard models on (a lot) more data certainly has an intellectually disappointing element---particularly given many rich ideas in the cognitive science and neuroscience literature on which architectural changes might be required to bring machine vision closer to human vision \citep{kriegeskorte2015deep, lake2017building, tavanaei2019deep, sinz2019engineering, dapello2020simulating, evans2021biological}. Additionally, large-scale training comes with infrastructure demands that are hard to meet for many academic researchers. On the other hand, we find it truly exciting to see that machine models are closing not just the OOD distortion robustness gap to humans, but that also, at least for some datasets, those models are actually making more human-like decisions on an individual image level; image-level response consistency is a much stricter behavioural requirement than just e.g.\ matching overall accuracies. Taken together, our results give reason to celebrate partial success in closing the gap between human and machine vision. In those cases where there is still ample room for improvement, our psychophysical \href{https://github.com/bethgelab/model-vs-human/}{benchmark datasets and toolbox} may prove useful in quantifying future progress.

\subsubsection*{Acknowledgments and disclosure of funding}
\begin{footnotesize}
\vspace{-0.2cm}
We thank Andreas Geiger, Simon Kornblith, Kristof Meding, Claudio Michaelis and Ludwig Schmidt for helpful discussions regarding different aspects of this work; Lukas Huber, Maximus Mutschler, David-Elias Künstle for feedback on the manuscript; Ken Kahn for pointing out typos; Santiago Cadena for sharing a PyTorch implementation of SimCLR; Katherine Hermann and her collaborators for providing supervised SimCLR baselines; Uli Wannek and Silke Gramer for infrastructure/administrative support; the many authors who made their models publicly available; and our anonymous reviewers for many valuable suggestions.

Furthermore, we are grateful to the International Max Planck Research School for Intelligent Systems (IMPRS-IS) for supporting R.G.; the Collaborative Research Center (Projektnummer 276693517---SFB 1233: Robust Vision) for supporting M.B. and F.A.W. This work was supported by the German Federal Ministry of Education and Research (BMBF): Tübingen AI Center, FKZ: 01IS18039A (W.B. and M.B.). F.A.W. is a member of the Machine Learning Cluster of Excellence, EXC number 2064/1---Project number 390727645. M.B. and W.B. acknowledge funding from the MICrONS program of the Intelligence Advanced Research Projects Activity (IARPA) via Department of Interior/Interior Business Center (DoI/IBC) contract number D16PC00003. W.B. acknowledges financial support via the Emmy Noether Research Group on The Role of Strong Response Consistency for Robust and Explainable Machine Vision funded by the German Research Foundation (DFG) under grant no. BR 6382/1-1.
\end{footnotesize}

\subsubsection*{Author contributions}
\begin{footnotesize}
\vspace{-0.2cm}
Project idea: R.G.\ and W.B.; project lead: R.G.; coding toolbox and model evaluation pipeline: R.G., K.N.\ and B.M.\ based on a prototype by R.G.; training models: K.N.\ with input from R.G., W.B.\ and M.B.; data visualisation: R.G., B.M.\ and K.N.\ with input from M.B., F.A.W. and W.B.; psychophysical data collection: T.T.\ (12 datasets) and B.M.\ (2 datasets) under the guidance of R.G.\ and F.A.W.; curating stimuli: R.G.; interpreting analyses and findings: R.G., M.B., F.A.W. and W.B.; guidance, feedback, infrastructure \& funding acquisition: M.B., F.A.W.\ and W.B.; paper writing: R.G.\ with help from F.A.W.\ and W.B.\ and input from all other authors.
\end{footnotesize}

\begin{small}
\bibliographystyle{unsrtnat}
\bibliography{refs}
\end{small}

\newpage
\appendix

\section*{Appendix}
\begin{figure}[h]
	\begin{subfigure}{0.19\linewidth}
			\centering
			\includegraphics[width=\linewidth]{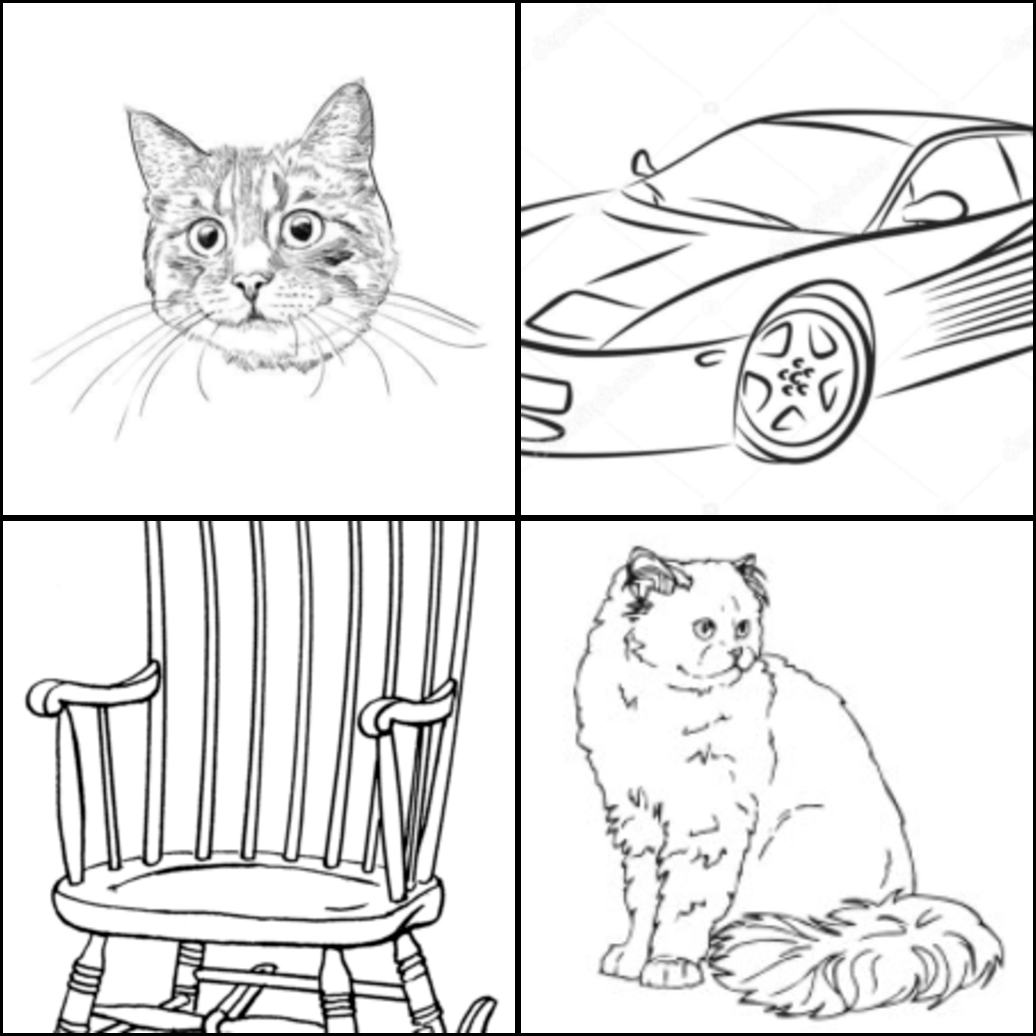}
			\caption{Sketch}
	\end{subfigure}\hfill
	\begin{subfigure}{0.19\linewidth}
			\centering
			\includegraphics[width=\linewidth]{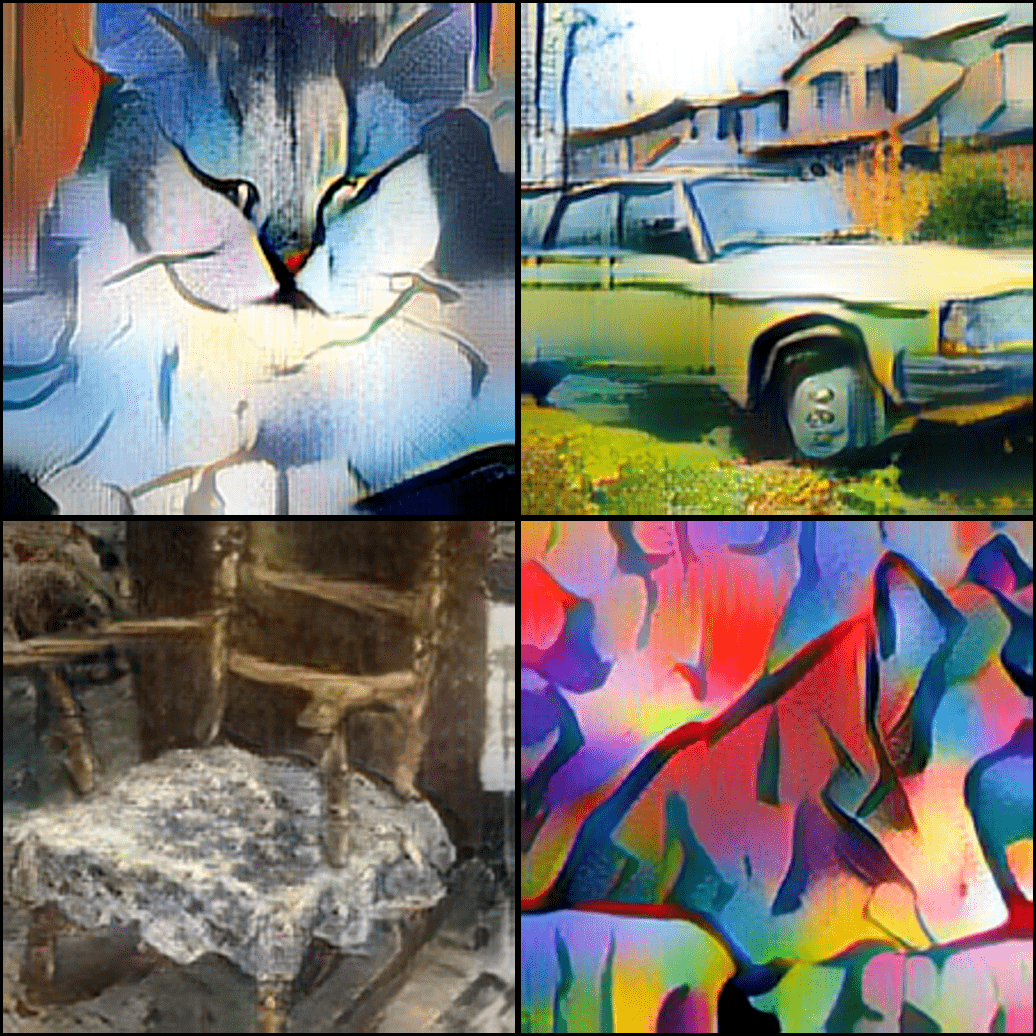}
			\caption{Stylized}
	\end{subfigure}\hfill
	\begin{subfigure}{0.19\linewidth}
			\centering
			\includegraphics[width=\linewidth]{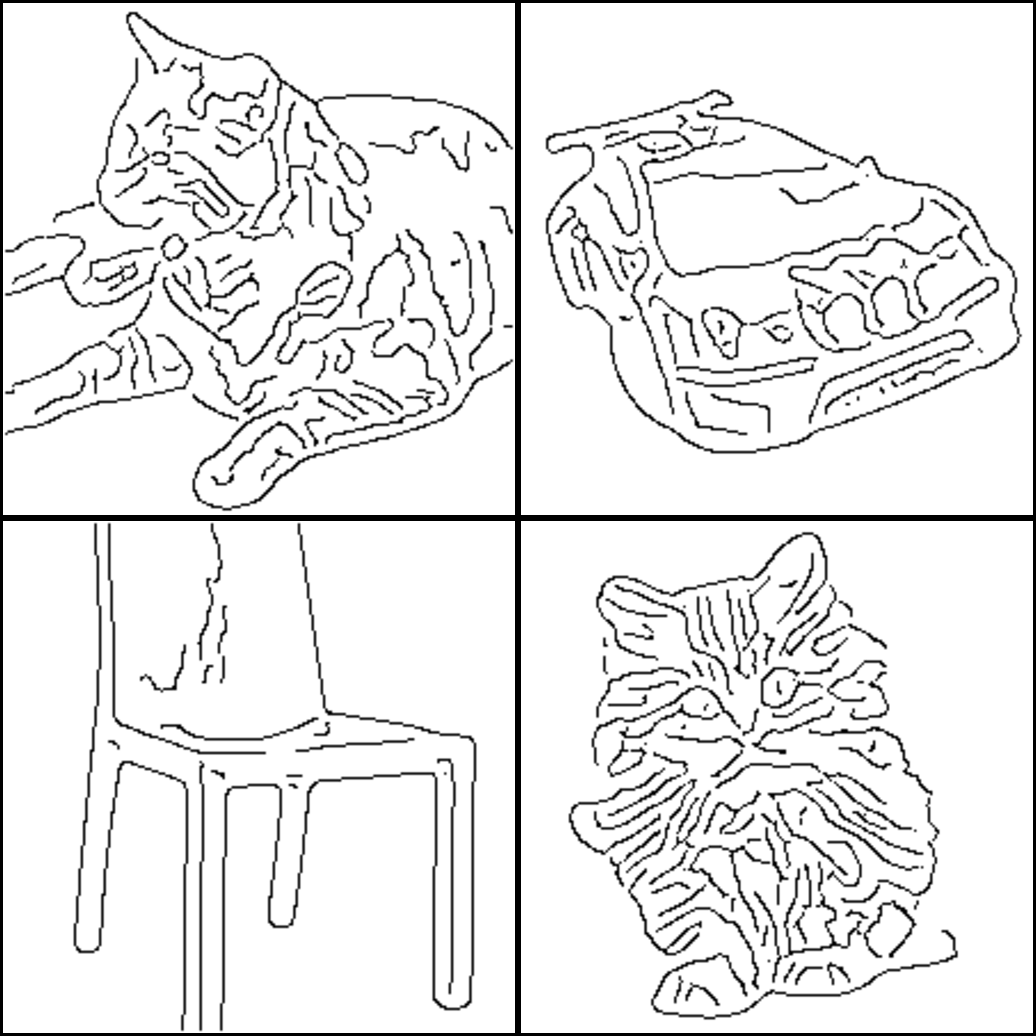}
			\caption{Edge}
	\end{subfigure}\hfill
	\begin{subfigure}{0.19\linewidth}
			\centering
			\includegraphics[width=\linewidth]{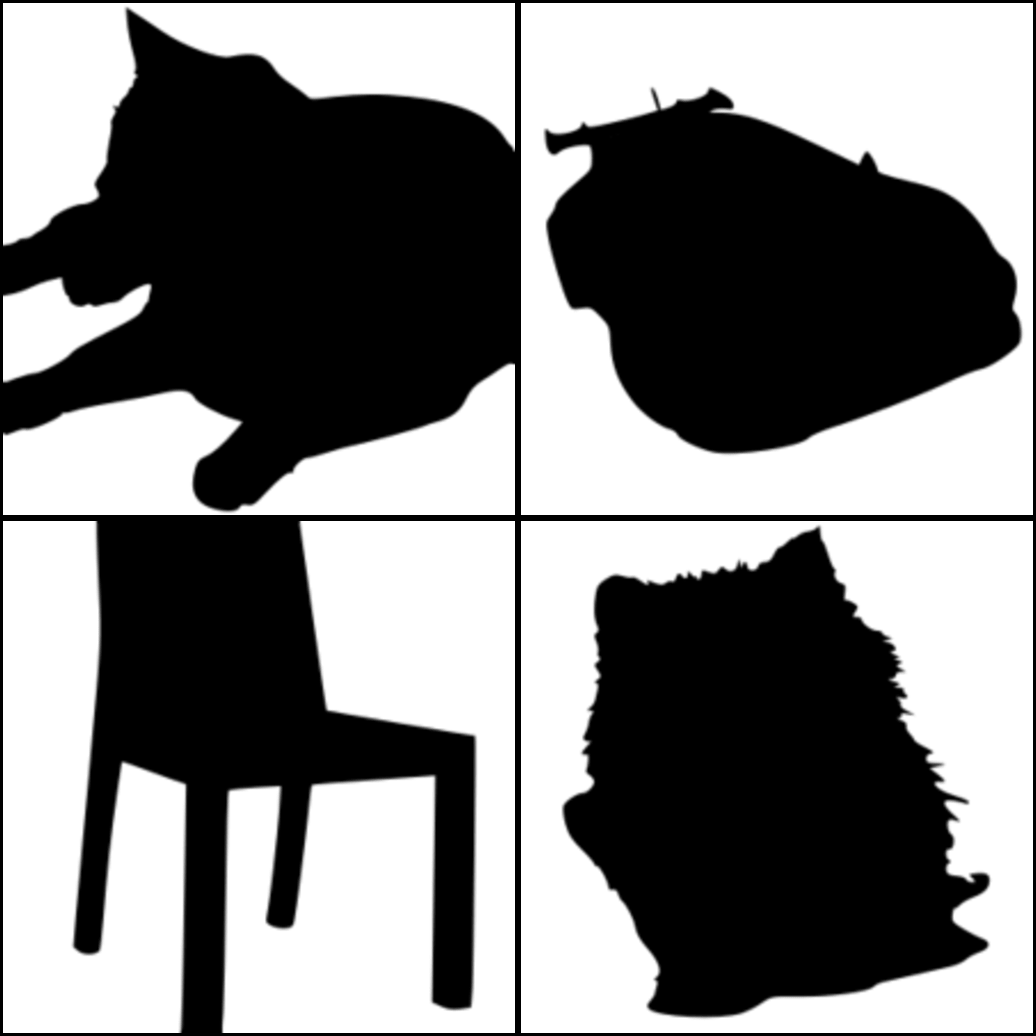}
			\caption{Silhouette}
	\end{subfigure}\hfill
	\begin{subfigure}{0.19\linewidth}
			\centering
			\includegraphics[width=\linewidth]{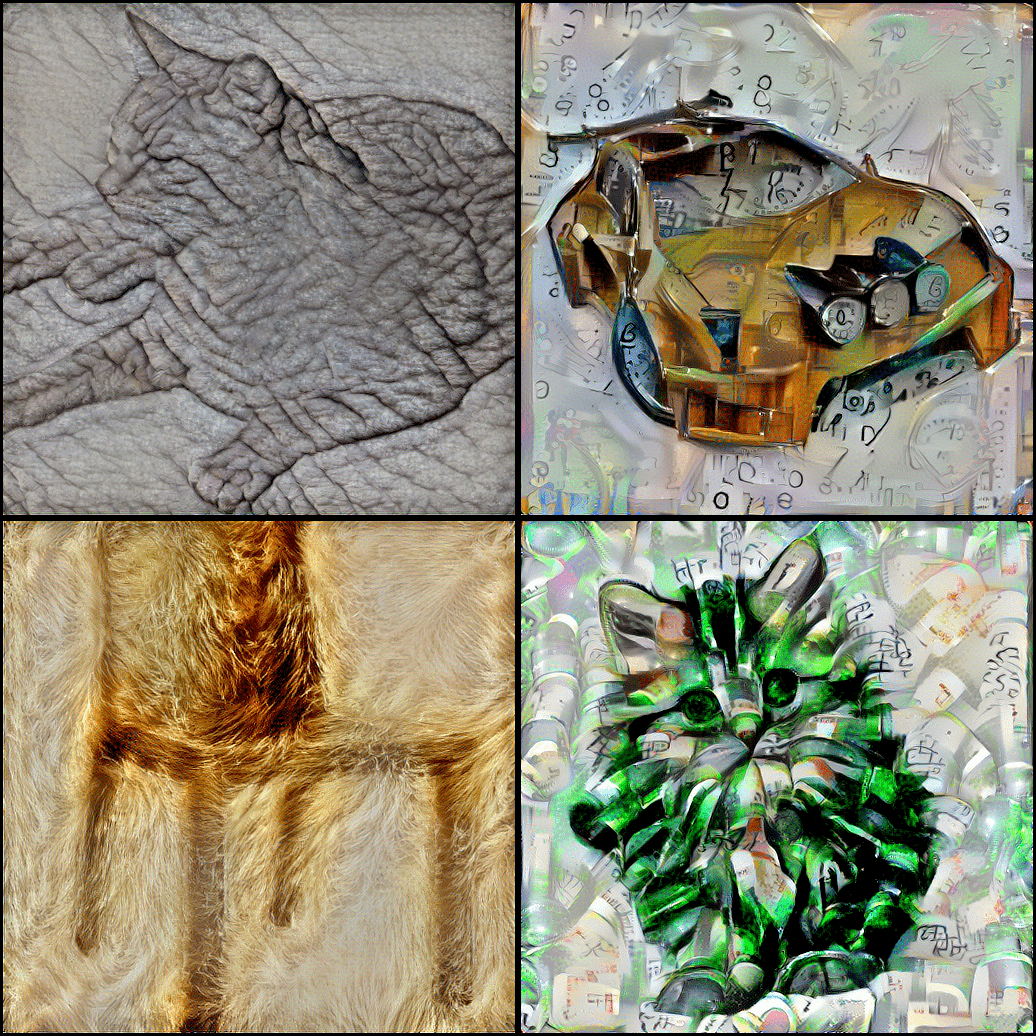}
			\caption{Cue conflict}
	\end{subfigure}\hfill
	\caption{Exemplary stimuli (nonparametric image manipulations) for the following datasets: sketch (7 observers, 800 trials each), stylized (5 observers, 800 trials each), edge (10 observers, 160 trials each), silhouette (10 observers, 160 trials each), and cue conflict (10 observers, 1280 trials each). Figures c--e reprinted from \citep{geirhos2020beyond} with permission from the authors. \citep{geirhos2020beyond} also analyzed ``diagnostic'' images, i.e.\ stimuli that most humans correctly classified (but few networks) and vice-versa.}
	\label{fig:methods_nonparametric_stimuli}
\end{figure}

\begin{figure*}[h]
	\centering
	\includegraphics[width=\linewidth]{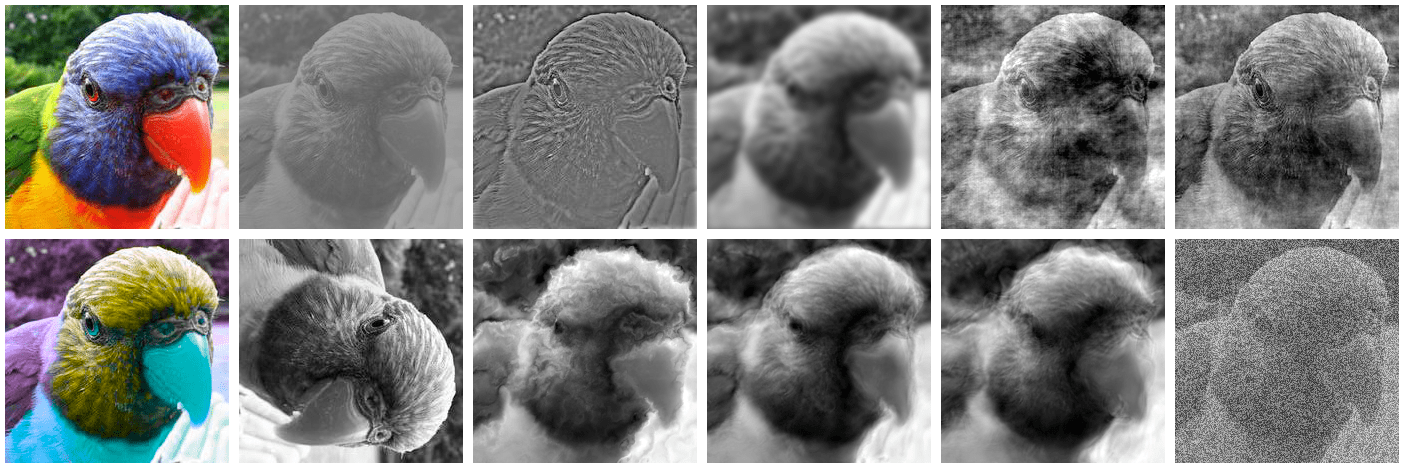}
	\caption{Exemplary stimuli (parametric image manipulations). Manipulations are either binary (e.g.\ colour vs.\ grayscale) or they have a parameter (such as the degree of rotation, or the contrast level). Top row: colour vs.\ grayscale (4 observers, 1280 trials each), low contrast (4 observers, 1280 trials each), high-pass (4 observers, 1280 trials each), low-pass/blurring (4 observers, 1280 trials each), phase noise (4 observers, 1120 trials each), true power spectrum vs.\ power equalisation (4 observers, 1120 trials each). Bottom row: true vs.\ opponent colour (4 observers, 1120 trials each), rotation (4 observers, 1280 trials each), Eidolon~I (4 observers, 1280 trials each), Eidolon II (4 observers, 1280 trials each), Eidolon III (4 observers, 1280 trials each, additive uniform noise (4 observers, 1280 trials each). Figure adapted from \citep{geirhos2018generalisation} with permission from the authors.}
	\label{fig:methods_parametric_stimuli}
\end{figure*}

We here provide details on models (\ref{app:model_details}), describe additional predictions and experiments regarding error consistency mentioned in Section~\ref{sec:consistency_between_models} (\ref{app:error_consistency_predictions}), report experimental details regarding our psychphysical experiments (\ref{app:experimental_details}), provide license information (\ref{app:license}), and describe training with ImageNet labels provided by CLIP (\ref{app:training_clip_labels}) as well as experiments with supervised SimCLR baseline models (\ref{app:supervised_simclr_baseline_models}), provide overall benchmark scores ranking different models (\ref{app:benchmark_scores}), describe a regression model (\ref{app:regression_model}) and motivate the choice of behavioural response mapping (\ref{app:mapping_decisions}). Stimuli are visualized in Figures~\ref{fig:methods_nonparametric_stimuli} and \ref{fig:methods_parametric_stimuli}.

Our Python library,``modelvshuman'', to test and benchmark models against high-quality human psychophyiscal data is available from \url{https://github.com/bethgelab/model-vs-human/}.

\section{Model details}
\label{app:model_details}

\paragraph{Standard supervised models.}
We used all 24 available pre-trained models from the \href{https://pytorch.org/docs/1.4.0/model_zoo.html}{PyTorch model zoo version 1.4.0} (VGG: with batch norm).

\paragraph{Self-supervised models.} InsDis \citep{wu2018unsupervised}, MoCo \citep{he2020momentum}, MoCoV2 \citep{chen2020improved}, PIRL \citep{misra2020self} and InfoMin \citep{tian2020makes} were obtained as pre-trained models from the \href{https://github.com/HobbitLong/PyContrast/}{PyContrast model zoo}. We trained one linear classifier per model on top of the self-supervised representation. A PyTorch \citep{paszke2019pytorch} implementation of SimCLR \citep{chen2020simple} was obtained via \href{https://github.com/tonylins/simclr-converter}{simclr-converter}. All self-supervised models use a ResNet-50 architecture and a different training approach within the framework of contrastive learning \citep[e.g.][]{oord2018representation}.

\paragraph{Adversarially trained models.} We obtained five adversarially trained models \citep{salman2020adversarially} from the \href{https://github.com/microsoft/robust-models-transfer}{robust-models-transfer} repository. All of them have a ResNet-50 architecture, but a different accuracy-L2-robustness tradeoff indicated by $\epsilon$. Here are the five models that we used, in increasing order of adversarial robustness: $\epsilon=0, 0.5, 1.0, 3.0, 5.0$.

\paragraph{Vision transformers.} Three ImageNet-trained vision transformer (ViT) models \citep{dosovitskiy2020image} were obtained from \href{https://github.com/rwightman/pytorch-image-models}{pytorch-image-models} \citep{rw2019timm}. Specifically, we used vit\_small\_patch16\_224, vit\_base\_patch16\_224 and vit\_large\_patch16\_224. They are referred to as ViT-S, ViT-B and ViT-L throughout the paper. Additionally, we included two transformers that were pre-trained on ImageNet21K \citep{deng2009imagenet}, i.e.\ 14M images with some 21K classes, before they were fine-tuned on ``standard'' ImageNet-1K. These two models are referred to as ViT-L (14M) and ViT-B (14M) in the paper. They were obtained from the \href{https://github.com/lukemelas/PyTorch-Pretrained-ViT}{PyTorch-Pretrained-ViT} repository, where they are called L\_16\_imagenet1k and B\_16\_imagenet1k. (No ViT-S model was available from the repository.) Note that the ``imagenet1k'' suffix in the model names does not mean the model was only trained on ImageNet1K. On the contrary, this indicates fine-tuning on ImageNet; as mentioned above these models were pre-trained on ImageNet21K before fine-tuning.

\paragraph{CLIP.} OpenAI trained a variety of CLIP models using different backbone networks \citep{radford2021learning}. Unfortunately, the best-performing model has not been released so far, and it is not currently clear whether it will be released at some point according to \href{https://github.com/openai/CLIP/issues/2}{issue \#2 of OpenAI's CLIP github repository}. We included the most powerful released model in our analysis, a model with a ViT-B/32 backbone.

\paragraph{Noisy Student} One pre-trained Noisy Student model was obtained from \href{https://github.com/rwightman/pytorch-image-models}{pytorch-image-models} \citep{rw2019timm}, where the model is called tf\_efficientnet\_e2\_ns\_475. This involved the following preprocessing (taken from \citep{rusak2021adapting}):

\begin{tiny}
\begin{lstlisting}[language=Python]
from PIL.Image import Image
from torchvision.transforms import Compose, Resize, CenterCrop, ToTensor, Normalize

def get_noisy_student_preprocessing():
    normalize = Normalize(mean=[0.485, 0.456, 0.406],
                          std=[0.229, 0.224, 0.225])
    img_size = 475
    crop_pct = 0.936
    scale_size = img_size / crop_pct
    return Compose([
        Resize(scale_size, interpolation=PIL.Image.BICUBIC),
        CenterCrop(img_size),
        ToTensor(),
        normalize,
    ])
\end{lstlisting}
\end{tiny}

\paragraph{SWSL} Two pre-trained SWSL (semi-weakly supervised) models were obtained from \href{https://github.com/facebookresearch/semi-supervised-ImageNet1K-models}{semi-supervised-ImageNet1K-models}, one with a ResNet-50 architecture and one with a ResNeXt101\_32x16d architecture.

\paragraph{BiT-M} Six pre-trained Big Transfer models were obtained from \href{https://github.com/rwightman/pytorch-image-models}{pytorch-image-models} \citep{rw2019timm}, where they are called resnetv2\_50x1\_bitm, resnetv2\_50x3\_bitm, resnetv2\_101x1\_bitm, resnetv2\_101x3\_bitm, resnetv2\_152x2\_bitm and resnetv2\_152x4\_bitm.

\paragraph{Linear classifier training procedure.}
The \href{https://github.com/HobbitLong/PyContrast}{PyContrast repository} by Yonglong Tian contains a Pytorch implementation of unsupervised representation learning methods, including pre-trained representation weights. The repository provides training and evaluation pipelines, but it supports only multi-node distributed training and does not (currently) provide weights for the classifier. We have used the repository's linear classifier evaluation pipeline to train classifiers for InsDis \citep{wu2018unsupervised}, MoCo \citep{he2020momentum}, MoCoV2 \citep{chen2020improved}, PIRL \citep{misra2020self} and InfoMin \citep{tian2020makes} on ImageNet. Pre-trained weights of the model representations (without classifier) were taken from the provided \href{https://www.dropbox.com/sh/87d24jqsl6ra7t2/AABdDXZZBTnMQCBrg1yrQKVCa?dl=0}{Dropbox link} and we then ran the training pipeline on a NVIDIA TESLA P100 using the default parameters configured in the pipeline. Detailed documentation about running the pipeline and parameters can be found in the \href{https://github.com/HobbitLong/PyContrast}{PyContrast repository} (commit \#3541b82).

\section{Error consistency predictions}
\label{app:error_consistency_predictions}

\begin{table*}[ht]
\caption{Error consistency across all five non-parametric datasets. Specifically, this comparison compares the influence of dataset vs.\ architecture (top) and the influence of flexibility vs.\ constraints (bottom). Results are described in Section~\ref{app:error_consistency_predictions}.}
\label{tab:error_consistency_predictions}
\begin{tabular}{@{}llllll@{}}
\toprule
                            & sketch & stylized & edge & silhouette & cue conflict \\ \midrule
ResNet-50 vs. VGG-16        & \textbf{0.74}   & \textbf{0.56}     & \textbf{0.68}  & \textbf{0.71}        & \textbf{0.59}         \\
ResNet-50 vs. ResNet-50 trained on Stylized-ImageNet & 0.44   & 0.09     & 0.10  & 0.67        & 0.27         \\\midrule
ResNet-50 vs. vision transformer (ViT-S)    & \textbf{0.67}   & \textbf{0.43}     & \textbf{0.41}  & \textbf{0.68}        & \textbf{0.48}         \\
ResNet-50 vs. BagNet-9      & 0.31   & 0.17     & 0.32  & 0.14        & 0.44         \\ \bottomrule
\end{tabular}
\end{table*}

In Section~\ref{sec:consistency_between_models}, we hypothesised that shortcut opportunities in the dataset may be a potential underlying cause of high error consistency between models, since all sufficiently flexible models will pick up on those same shortcuts. We then made two predictions which we test here.

\paragraph{Dataset vs.\ architecture.} \emph{Prediction:} error consistency between two identical models trained on very different datasets, such as ImageNet vs. Stylized-ImageNet, is much lower than error consistency between very different models (ResNet-50 vs.\ VGG-16) trained on the same dataset. \emph{Observation:} According to Table~\ref{tab:error_consistency_predictions}, this is indeed the case---training ResNet-50 on a different dataset, Stylized-ImageNet \citep{geirhos2019imagenettrained}, leads to lower error consistency than comparing two ImageNet-trained CNNs with different architecture. While this relationship is not perfect (e.g., the difference is small for silhouette images), we have confirmed that this is a general pattern not limited to the specific networks in the table.

\paragraph{Flexibility vs.\ constraints.} \emph{Prediction:} error consistency between ResNet-50 and a highly flexible model (e.g., a vision transformer) is much higher than error consistency between ResNet-50 and a highly constrained model like BagNet-9 \citep{brendel2019approximating}. \emph{Observation:} A vision transformer (ViT-S) indeed shows higher error consistency with ResNet-50 than with BagNet-9 (see Table~\ref{tab:error_consistency_predictions}). However, this difference is not large for one out of five datasets (cue conflict). One could imagine different reasons for this: perhaps BagNet-9 is still flexible enough to learn a decision rule close to the one of standard ResNet-50 for cue conflict images; and of course there is also the possibility that the hypothesis is wrong. Further insights could be gained by testing successively more constrained versions of the same base model.

\section{Experimental details regarding psychophysical experiments}
\label{app:experimental_details}

\subsection{Participant instructions and preparation}
\label{app:participant_instruction}
Participants were explained how to respond (via mouse click), instructed to respond as accurately as possible, and to go with their best guess if unsure. In order to rule out any potential misunderstanding, participants were asked to name all 16 categories on the response screen. Prior to the experiment, visual acuity was measured with a Snellen chart to ensure normal or corrected to normal vision. Furthermore, four blocks of 80 practice trials each (320 practice trials in total) on undistorted colour or greyscale images were conducted (non-overlapping with experimental stimuli) to gain familiarity with the task. During practice trials, but not experimental trials, visual and auditory feedback was provided: the correct category was highlighted and a ``beep'' sound was played for incorrect or missed trials. The experiment itself consisted of blocks of 80 trials each, after each blocks participants were free to take a break. In order to increase participant motivation, aggregated performance over the last block was displayed on the screen.

\subsection{Participant risks}
\label{app:participant_risks}
Our experiment was a standard perceptual experiment, for which no IRB approval was required. The task consisted of viewing corrupted images and clicking with a computer mouse. In order to limit participant risks related to a COVID-19 infection, we implemented the following measures: (1.)~The experimenter was tested for corona twice per week. (2.)~Prior to participation in our experiments, participants were explained that they could perform a (cost-free) corona test next to our building, and that if they choose to do so, we would pay them 10€/hour for the time spent doing the test and waiting for the result (usually approx.\ 15--30min). (3.)~Experimenter and participant adhered to a strict distance of at least 1.5m during the entire course of the experiment, including instructions and practice trials. During the experiment itself, the participant was the only person in the room; the experimenter was seated in an adjacent room. (4.)~Wearing a medical mask was mandatory for both experimenter and participant. (5.)~Participants were asked to disinfect their hands prior to the experiment; additionally the desk, mouse etc.\ were disinfected after completion of an experiment. (6.)~Participants were tested in a room where high-performance ventilation was installed; in order to ensure that the ventilation was working as expected we performed a one-time safety check measuring CO2 parts-per-million before we decided to go ahead with the experiments.

\subsection{Participant remuneration}
\label{app:participant_remuneration}
Participants were paid 10€ per hour or granted course credit. Additionally, an incentive bonus of up to 15€ could be earned on top of the standard remuneration. This was meant to further motivate our participants to achieve their optimal performance. The minimum performance for receiving a bonus was set as 15\% below the mean of the previous experiments accuracy. The bonus then was linearly calculated with the maximal bonus being given from 15\% above the previous experiments mean. The total amount spent on participant compensation amounts to 647,50€.

\subsection{Participant declaration of consent}
Participants were asked to review and sign the following declaration of consent (of which they received a copy):\textit{\\
    \\
    \textbf{Psychophysical study}\\
    Your task consists of viewing visual stimuli on a computer monitor and evaluating them by pressing a key.
    Participation in a complete experimental session is remunerated at 10 Euros/hour.\\
   \textbf{ Declaration of consent}\\
    Herewith I agree to participate in a behavioural experiment to study visual perception. My participation in the study is voluntary. I am informed that I can stop the experiment at any time and without giving any reason without incurring any disadvantages. I know that I can contact the experimenter at any time with questions
    about the research project.\\
   \textbf{ Declaration of consent for data processing and data publication}\\
    Herewith I agree that the experimental data obtained in the course of the experiment may be used in semianonymised form for scientific evaluation and publication.
    I agree that my personal data (e.g. name, phone number, address) may be stored in digital form; they will not
    be used for any other purpose than for contacting me. This personal data will remain exclusively within
    the Wichmannlab and will not be passed on to third parties at any time.}

\section{Licenses}
\label{app:license}
Licenses for datasets, code and models are included in our code (see directory ``licenses/'', file ``LICENSES\_OVERVIEW.md'' of \url{https://github.com/bethgelab/model-vs-human}.

\section{Training with CLIP labels}
\label{app:training_clip_labels}

\begin{figure}[h]
	\begin{subfigure}{0.49\linewidth}
			\centering
			\includegraphics[width=\linewidth]{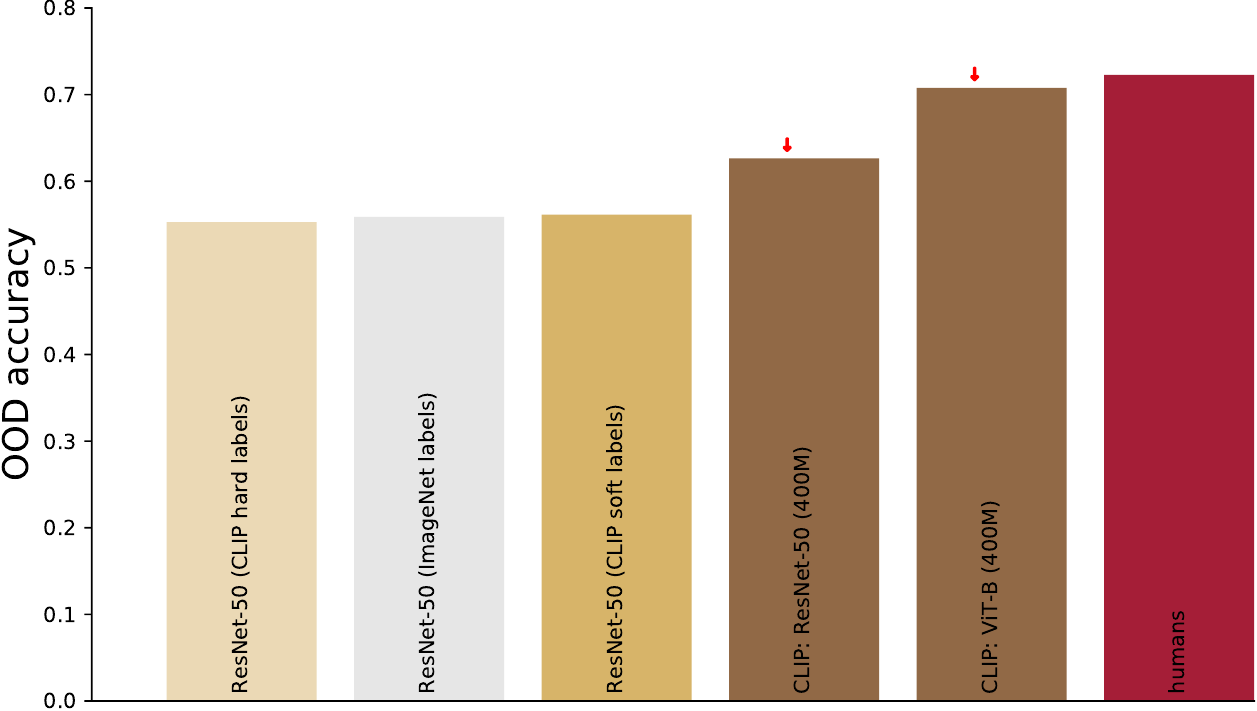}
			\caption{OOD accuracy (higher = better).}
			\vspace{\captionspaceII}
	\end{subfigure}\hfill
	\begin{subfigure}{0.49\linewidth}
			\centering
			\includegraphics[width=\linewidth]{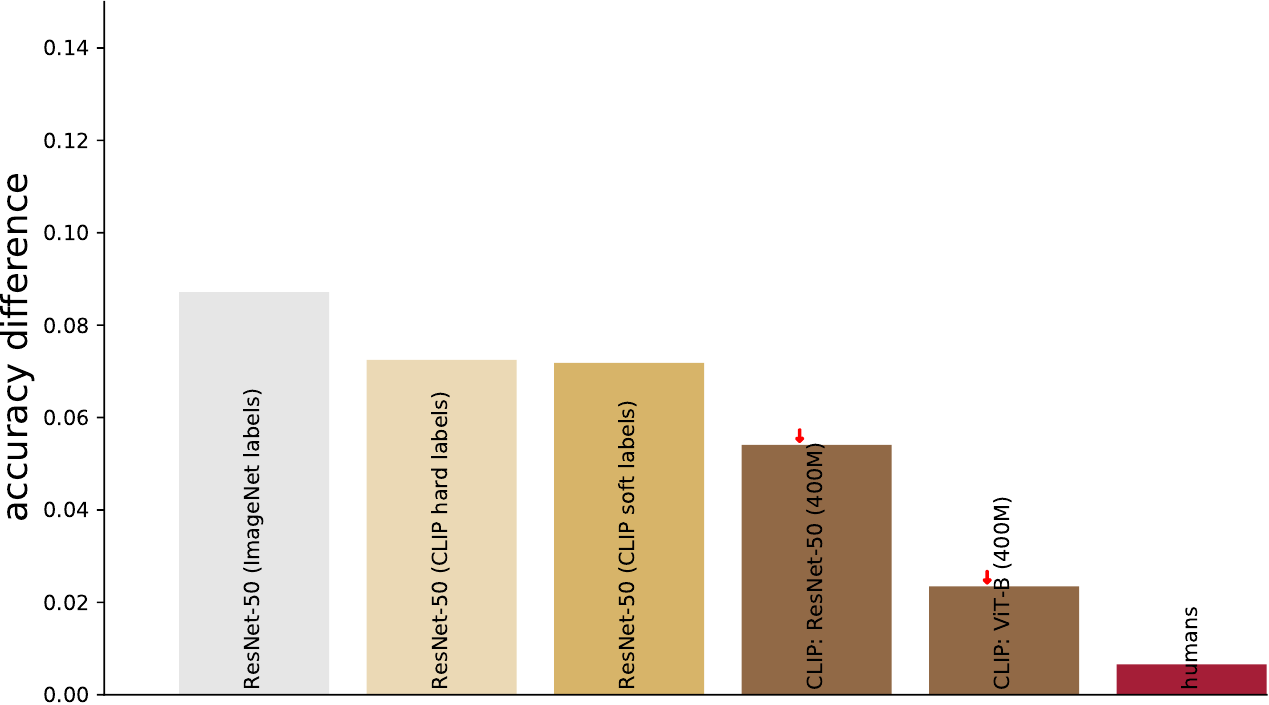}
			\caption{Accuracy difference (lower = better).}
			\vspace{\captionspaceII}
	\end{subfigure}\hfill
	\begin{subfigure}{0.49\linewidth}
			\centering
			\includegraphics[width=\linewidth]{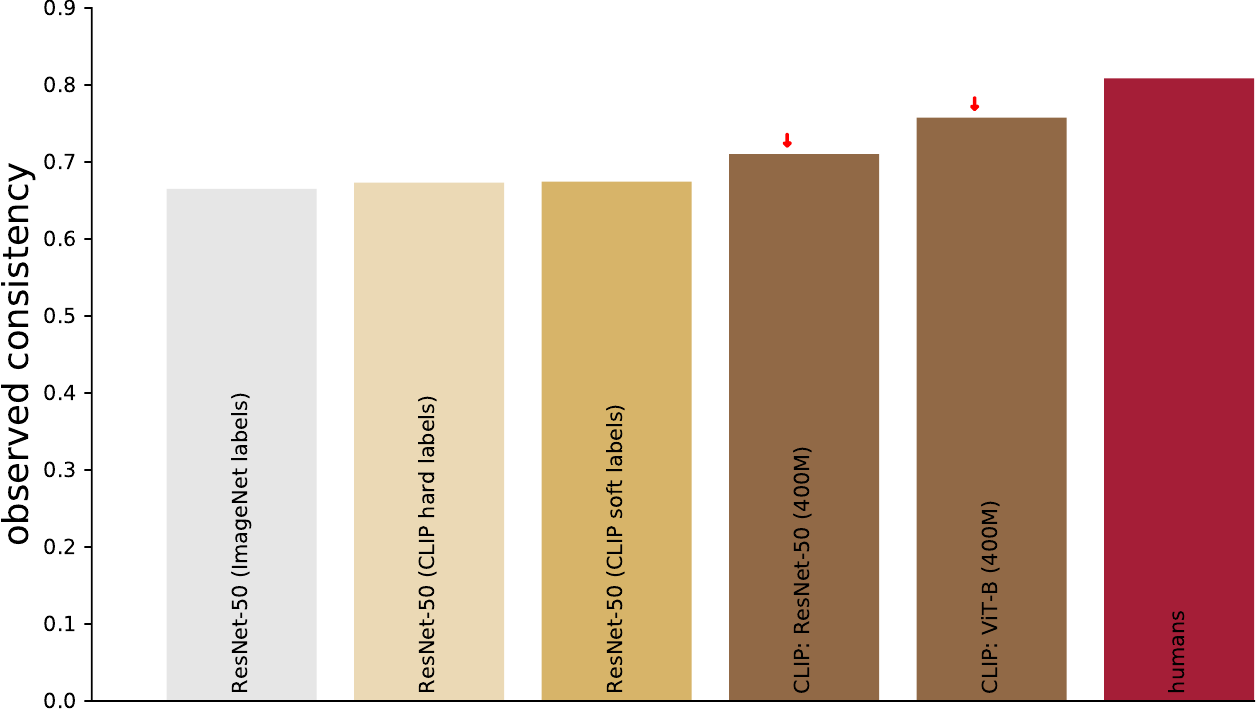}
			\caption{Observed consistency (higher = better).}
			\vspace{\captionspaceII}
	\end{subfigure}\hfill
	\begin{subfigure}{0.49\linewidth}
			\centering
			\includegraphics[width=\linewidth]{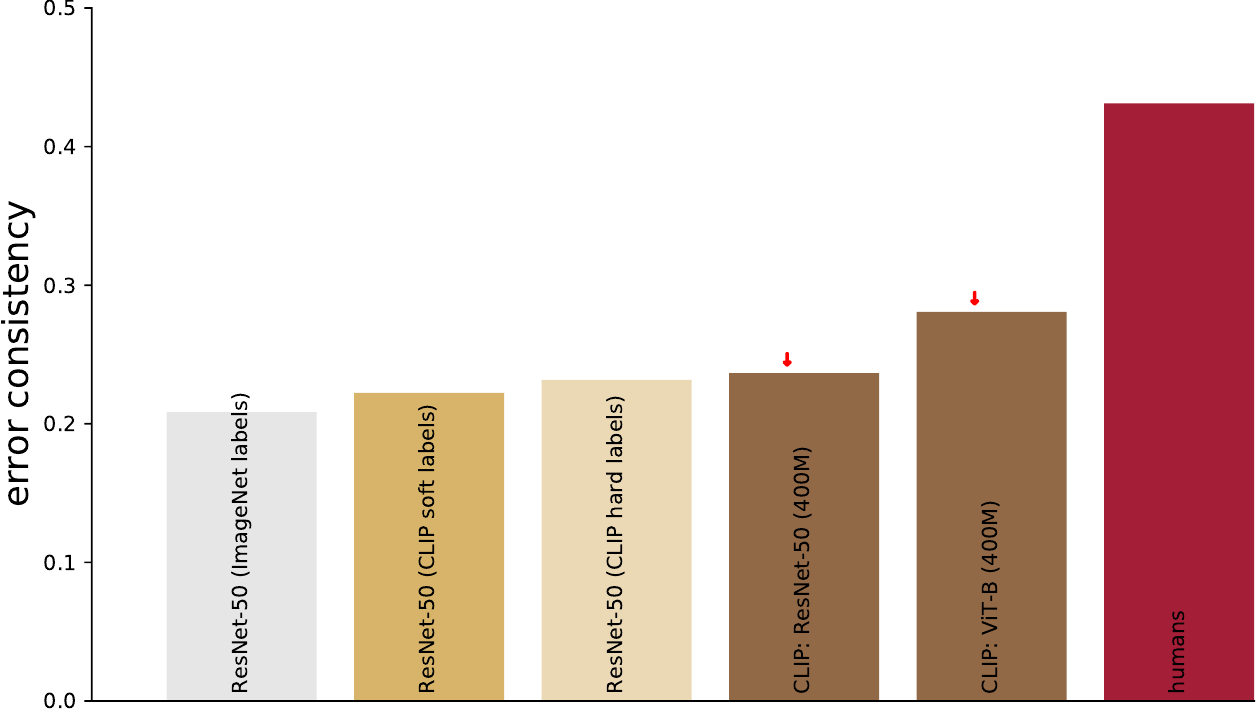}
			\caption{Error consistency (higher = better).}
			\vspace{\captionspaceII}
	\end{subfigure}\hfill
	\caption{Aggregated results comparing models with and without CLIP-provided labels. Comparison of standard ResNet-50 (\textcolor{supervised.grey}{light grey}), CLIP with vision transformer backend (\textcolor{brown1}{brown}), CLIP with ResNet-50 backend (\textcolor{brown1}{brown}), and standard ResNet-50 with hard labels (\textcolor{clip.hard.labels}{bright yellow}) vs.\ soft labels (\textcolor{clip.soft.labels}{dark yellow}) provided by evaluating standard CLIP on ImageNet; as well as humans \textcolor{red}{(red diamonds)} for comparison. Detailed performance across datasets in Figure~\ref{fig:CLIP_results_accuracy_error_consistency}.}
	\label{fig:CLIP_benchmark_barplots}
\end{figure}

As CLIP performed very well across metrics, we intended to obtain a better understanding for why this might be the case. One hypothesis is that CLIP might just receive better labels: About 6\% of ImageNet validation images are mis-labeled according to \citet{northcutt2021pervasive}. We therefore designed an experiment where we re-labeled the entire ImageNet training and validation dataset using CLIP predictions as ground truth (\url{https://github.com/kantharajucn/CLIP-imagenet-evaluation}). Having re-labeled ImageNet, we then trained a standard ResNet-50 model from scratch on this dataset using the standard \href{https://github.com/pytorch/examples/tree/master/imagenet}{PyTorch ImageNet training script}. Training was performed on our on-premise cloud using four RTX 2080 Ti GPUs for five days. We ran the training pipeline in distributed mode with an nccl backed using the default parameters configured in the script, except for the number of workers which we changed to 25. Cross-entropy loss was used to train two models, once with CLIP hard labels (the top-1 class predicted by CLIP) and once with CLIP soft labels (using CLIP's full posterior distribution as training target). The accuracies on the original ImageNet validation dataset of the resulting models ResNet50-CLIP-hard-labels and ResNet-50-CLIP-soft-labels are 63.53 (top-1), 86.97 (top-5) and 64.63 (top-1), 88.60 (top-5) respectively. In order to make sure that the model trained on soft labels had indeed learned to approximate CLIP's posterior distribution on ImageNet, we calculated the KL divergence between CLIP soft labels and probability distributions from ResNet-50 trained on the CLIP soft labels. The resulting value of 0.001 on both ImageNet training and validation dataset is sufficiently small to conclude that the model had successfully learned to approximate CLIP's posterior distribution on ImageNet. The results are visualised in Figure~\ref{fig:CLIP_benchmark_barplots}. The results indicate that simply training a standard ResNet-50 model with labels provided by CLIP does not lead to strong improvements on any metric, which means that ImageNet label errors are unlikely to hold standard models back in terms of OOD accuracy and consistency with human responses.

\begin{table}[ht]
\caption{Benchmark table of model results. The three metrics ``accuracy difference'' ``observed consistency'' and ``error consistency'' (plotted in Figure~\ref{fig:benchmark_barplots}) each produce a different model ranking. The mean rank of a model across those three metrics is used to rank the models on our benchmark.}
\label{tab:benchmark_table_humanlike}
\centering
\begin{tabular}{lrrrr}
\toprule
                        model & accuracy diff. $\downarrow$ & obs. consistency $\uparrow$ & error consistency $\uparrow$ & mean rank $\downarrow$ \\
\midrule
           CLIP: ViT-B (400M) &              \textbf{0.023} &                       0.758 &               \textbf{0.281} &         \textbf{1.333} \\
     SWSL: ResNeXt-101 (940M) &                       0.028 &                       0.752 &                        0.237 &                  4.000 \\
    BiT-M: ResNet-101x1 (14M) &                       0.034 &                       0.733 &                        0.252 &                  4.333 \\
    BiT-M: ResNet-152x2 (14M) &                       0.035 &                       0.737 &                        0.243 &                  5.000 \\
                        ViT-L &                       0.033 &                       0.738 &                        0.222 &                  6.667 \\
    BiT-M: ResNet-152x4 (14M) &                       0.035 &                       0.732 &                        0.233 &                  7.667 \\
     BiT-M: ResNet-50x3 (14M) &                       0.040 &                       0.726 &                        0.228 &                  9.333 \\
     BiT-M: ResNet-50x1 (14M) &                       0.042 &                       0.718 &                        0.240 &                  9.667 \\
                  ViT-L (14M) &                       0.035 &                       0.744 &                        0.206 &                  9.667 \\
       SWSL: ResNet-50 (940M) &                       0.041 &                       0.727 &                        0.211 &                 11.667 \\
                        ViT-B &                       0.044 &                       0.719 &                        0.223 &                 12.000 \\
    BiT-M: ResNet-101x3 (14M) &                       0.040 &                       0.720 &                        0.204 &                 14.333 \\
                  densenet201 &                       0.060 &                       0.695 &                        0.212 &                 15.000 \\
                  ViT-B (14M) &                       0.049 &                       0.717 &                        0.209 &                 15.000 \\
                        ViT-S &                       0.066 &                       0.684 &                        0.216 &                 16.667 \\
                  densenet169 &                       0.065 &                       0.688 &                        0.207 &                 17.333 \\
                inception\_v3 &                       0.066 &                       0.677 &                        0.211 &                 17.667 \\
 Noisy Student: ENetL2 (300M) &                       0.040 &              \textbf{0.764} &                        0.169 &                 18.000 \\
         ResNet-50 L2 eps 1.0 &                       0.079 &                       0.669 &                        0.224 &                 21.000 \\
         ResNet-50 L2 eps 3.0 &                       0.079 &                       0.663 &                        0.239 &                 22.000 \\
           wide\_resnet101\_2 &                       0.068 &                       0.676 &                        0.187 &                 24.333 \\
          SimCLR: ResNet-50x4 &                       0.071 &                       0.698 &                        0.179 &                 24.667 \\
          SimCLR: ResNet-50x2 &                       0.073 &                       0.686 &                        0.180 &                 25.333 \\
         ResNet-50 L2 eps 0.5 &                       0.078 &                       0.668 &                        0.203 &                 25.333 \\
                  densenet121 &                       0.077 &                       0.671 &                        0.200 &                 25.333 \\
                    resnet101 &                       0.074 &                       0.671 &                        0.192 &                 25.667 \\
                    resnet152 &                       0.077 &                       0.675 &                        0.190 &                 25.667 \\
            resnext101\_32x8d &                       0.074 &                       0.674 &                        0.182 &                 26.667 \\
         ResNet-50 L2 eps 5.0 &                       0.087 &                       0.649 &                        0.240 &                 27.000 \\
                     resnet50 &                       0.087 &                       0.665 &                        0.208 &                 28.667 \\
                     resnet34 &                       0.084 &                       0.662 &                        0.205 &                 29.333 \\
                    vgg19\_bn &                       0.081 &                       0.660 &                        0.200 &                 30.000 \\
             resnext50\_32x4d &                       0.079 &                       0.666 &                        0.184 &                 30.333 \\
          SimCLR: ResNet-50x1 &                       0.080 &                       0.667 &                        0.179 &                 32.000 \\
                     resnet18 &                       0.091 &                       0.648 &                        0.201 &                 34.667 \\
                    vgg16\_bn &                       0.088 &                       0.651 &                        0.198 &                 34.667 \\
            wide\_resnet50\_2 &                       0.084 &                       0.663 &                        0.176 &                 35.667 \\
            MoCoV2: ResNet-50 &                       0.083 &                       0.660 &                        0.177 &                 36.000 \\
                mobilenet\_v2 &                       0.092 &                       0.645 &                        0.196 &                 37.000 \\
         ResNet-50 L2 eps 0.0 &                       0.086 &                       0.654 &                        0.178 &                 37.333 \\
                  mnasnet1\_0 &                       0.092 &                       0.646 &                        0.189 &                 38.333 \\
                    vgg11\_bn &                       0.106 &                       0.635 &                        0.193 &                 38.667 \\
           InfoMin: ResNet-50 &                       0.086 &                       0.659 &                        0.168 &                 39.333 \\
                    vgg13\_bn &                       0.101 &                       0.631 &                        0.180 &                 41.000 \\
                  mnasnet0\_5 &                       0.110 &                       0.617 &                        0.173 &                 45.000 \\
              MoCo: ResNet-50 &                       0.107 &                       0.617 &                        0.149 &                 47.000 \\
                      alexnet &                       0.118 &                       0.597 &                        0.165 &                 47.333 \\
               squeezenet1\_1 &                       0.131 &                       0.593 &                        0.175 &                 47.667 \\
              PIRL: ResNet-50 &                       0.119 &                       0.607 &                        0.141 &                 48.667 \\
        shufflenet\_v2\_x0\_5 &                       0.126 &                       0.592 &                        0.160 &                 49.333 \\
            InsDis: ResNet-50 &                       0.131 &                       0.593 &                        0.138 &                 50.667 \\
               squeezenet1\_0 &                       0.145 &                       0.574 &                        0.153 &                 51.000 \\
\bottomrule
\end{tabular}

\end{table}

\begin{table}[ht]
\caption{Benchmark table of model results (accuracy).}
\label{tab:benchmark_table_accurate}
\centering
\begin{tabular}{lrr}
\toprule
                        model & OOD accuracy $\uparrow$ & rank $\downarrow$ \\
\midrule
 Noisy Student: ENetL2 (300M) &          \textbf{0.829} &    \textbf{1.000} \\
                  ViT-L (14M) &                   0.733 &             2.000 \\
           CLIP: ViT-B (400M) &                   0.708 &             3.000 \\
                        ViT-L &                   0.706 &             4.000 \\
     SWSL: ResNeXt-101 (940M) &                   0.698 &             5.000 \\
    BiT-M: ResNet-152x2 (14M) &                   0.694 &             6.000 \\
    BiT-M: ResNet-152x4 (14M) &                   0.688 &             7.000 \\
    BiT-M: ResNet-101x3 (14M) &                   0.682 &             8.000 \\
     BiT-M: ResNet-50x3 (14M) &                   0.679 &             9.000 \\
          SimCLR: ResNet-50x4 &                   0.677 &            10.000 \\
       SWSL: ResNet-50 (940M) &                   0.677 &            11.000 \\
    BiT-M: ResNet-101x1 (14M) &                   0.672 &            12.000 \\
                  ViT-B (14M) &                   0.669 &            13.000 \\
                        ViT-B &                   0.658 &            14.000 \\
     BiT-M: ResNet-50x1 (14M) &                   0.654 &            15.000 \\
          SimCLR: ResNet-50x2 &                   0.644 &            16.000 \\
                  densenet201 &                   0.621 &            17.000 \\
                  densenet169 &                   0.613 &            18.000 \\
          SimCLR: ResNet-50x1 &                   0.596 &            19.000 \\
            resnext101\_32x8d &                   0.594 &            20.000 \\
                    resnet152 &                   0.584 &            21.000 \\
           wide\_resnet101\_2 &                   0.583 &            22.000 \\
                    resnet101 &                   0.583 &            23.000 \\
                        ViT-S &                   0.579 &            24.000 \\
                  densenet121 &                   0.576 &            25.000 \\
            MoCoV2: ResNet-50 &                   0.571 &            26.000 \\
                inception\_v3 &                   0.571 &            27.000 \\
           InfoMin: ResNet-50 &                   0.571 &            28.000 \\
             resnext50\_32x4d &                   0.569 &            29.000 \\
            wide\_resnet50\_2 &                   0.566 &            30.000 \\
                     resnet50 &                   0.559 &            31.000 \\
                     resnet34 &                   0.553 &            32.000 \\
         ResNet-50 L2 eps 0.5 &                   0.551 &            33.000 \\
         ResNet-50 L2 eps 1.0 &                   0.547 &            34.000 \\
                    vgg19\_bn &                   0.546 &            35.000 \\
         ResNet-50 L2 eps 0.0 &                   0.545 &            36.000 \\
         ResNet-50 L2 eps 3.0 &                   0.530 &            37.000 \\
                    vgg16\_bn &                   0.530 &            38.000 \\
                  mnasnet1\_0 &                   0.524 &            39.000 \\
                     resnet18 &                   0.521 &            40.000 \\
                mobilenet\_v2 &                   0.520 &            41.000 \\
              MoCo: ResNet-50 &                   0.502 &            42.000 \\
         ResNet-50 L2 eps 5.0 &                   0.501 &            43.000 \\
                    vgg13\_bn &                   0.499 &            44.000 \\
                    vgg11\_bn &                   0.498 &            45.000 \\
              PIRL: ResNet-50 &                   0.489 &            46.000 \\
                  mnasnet0\_5 &                   0.472 &            47.000 \\
            InsDis: ResNet-50 &                   0.468 &            48.000 \\
        shufflenet\_v2\_x0\_5 &                   0.440 &            49.000 \\
                      alexnet &                   0.434 &            50.000 \\
               squeezenet1\_1 &                   0.425 &            51.000 \\
               squeezenet1\_0 &                   0.401 &            52.000 \\
\bottomrule
\end{tabular}

\end{table}

\begin{figure}[h]
	\begin{subfigure}{\figwidth}
			\centering
			\textbf{Accuracy}
			\includegraphics[width=\linewidth]{colour_OOD-accuracy.pdf}
			\vspace{\captionspace}
			\caption{Colour vs. greyscale}
			\vspace{\captionspaceII}
		\end{subfigure}\hfill
		\begin{subfigure}{\figwidth}
			\centering
			\textbf{Error consistency}
			\includegraphics[width=\linewidth]{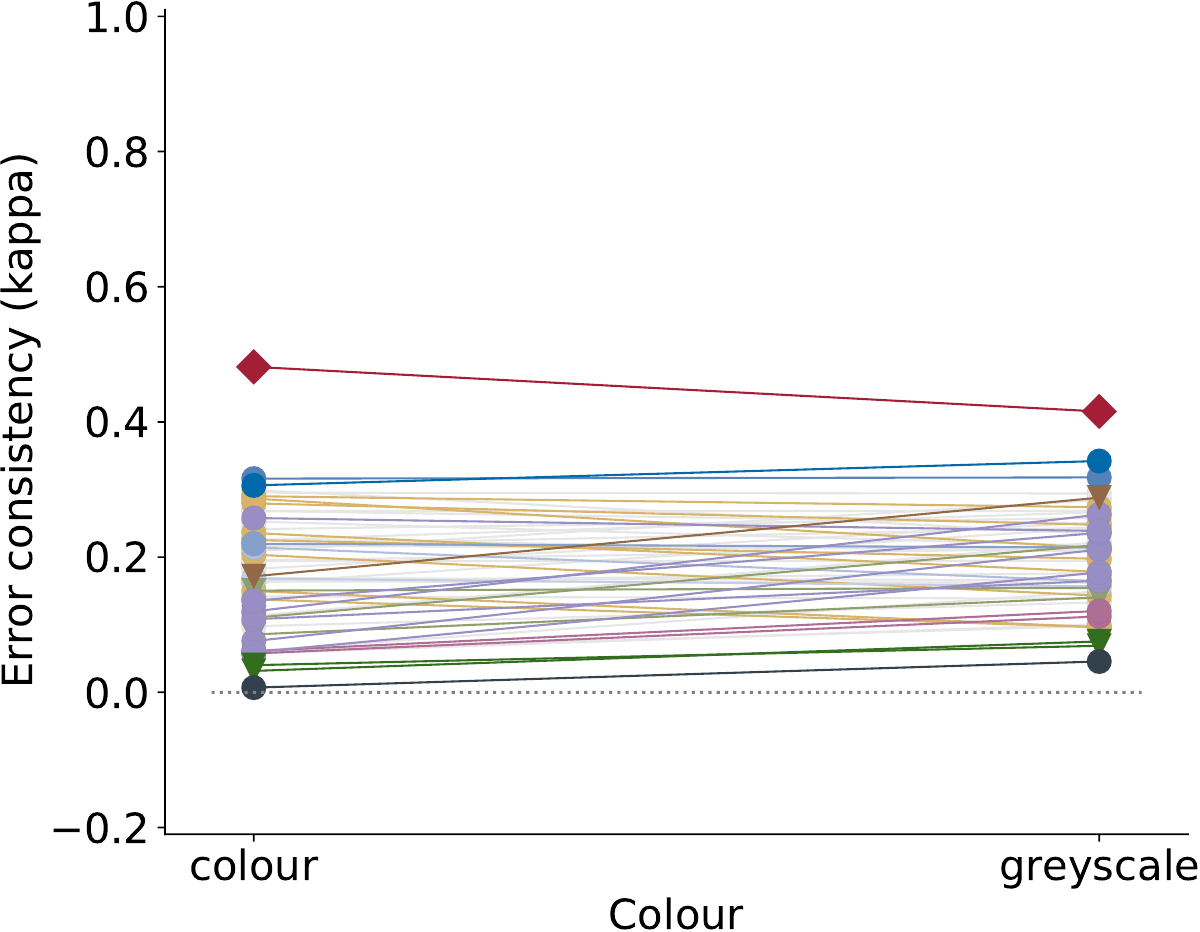}
			\vspace{\captionspace}
			\caption*{}
			\vspace{\captionspaceII}
		\end{subfigure}\hfill
		\begin{subfigure}{\figwidth}
			\centering
			\textbf{Accuracy}
			\includegraphics[width=\linewidth]{false-colour_OOD-accuracy.pdf}
			\vspace{\captionspace}
			\caption{True vs. false colour}
			\vspace{\captionspaceII}
		\end{subfigure}\hfill
		\begin{subfigure}{\figwidth}
			\centering
			\textbf{Error consistency}
			\includegraphics[width=\linewidth]{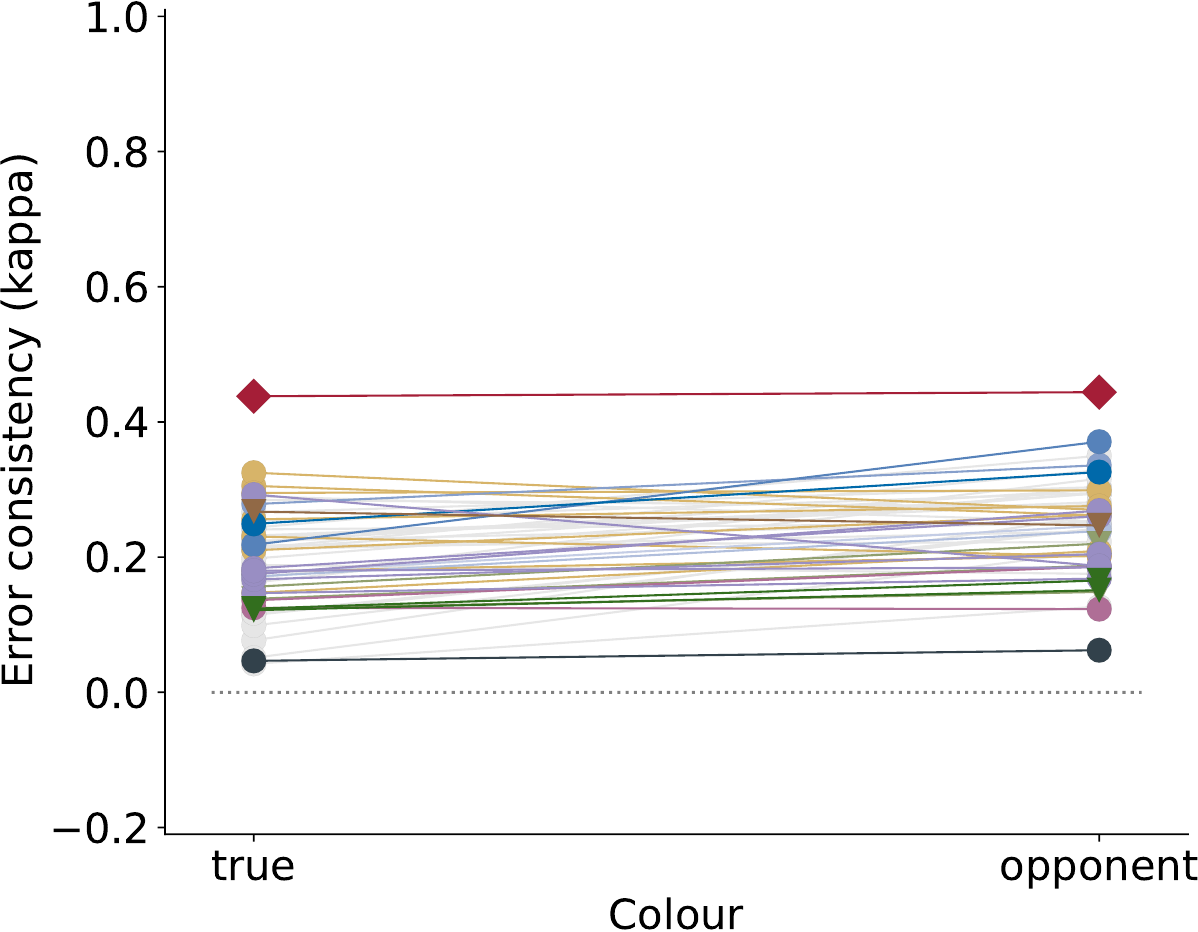}
			\vspace{\captionspace}
			\caption*{}
			\vspace{\captionspaceII}
		\end{subfigure}\hfill

		\begin{subfigure}{\figwidth}
			\centering
			\includegraphics[width=\linewidth]{uniform-noise_OOD-accuracy.pdf}
			\vspace{\captionspace}
			\caption{Uniform noise}
			\vspace{\captionspaceII}
		\end{subfigure}\hfill
		\begin{subfigure}{\figwidth}
			\centering
			\includegraphics[width=\linewidth]{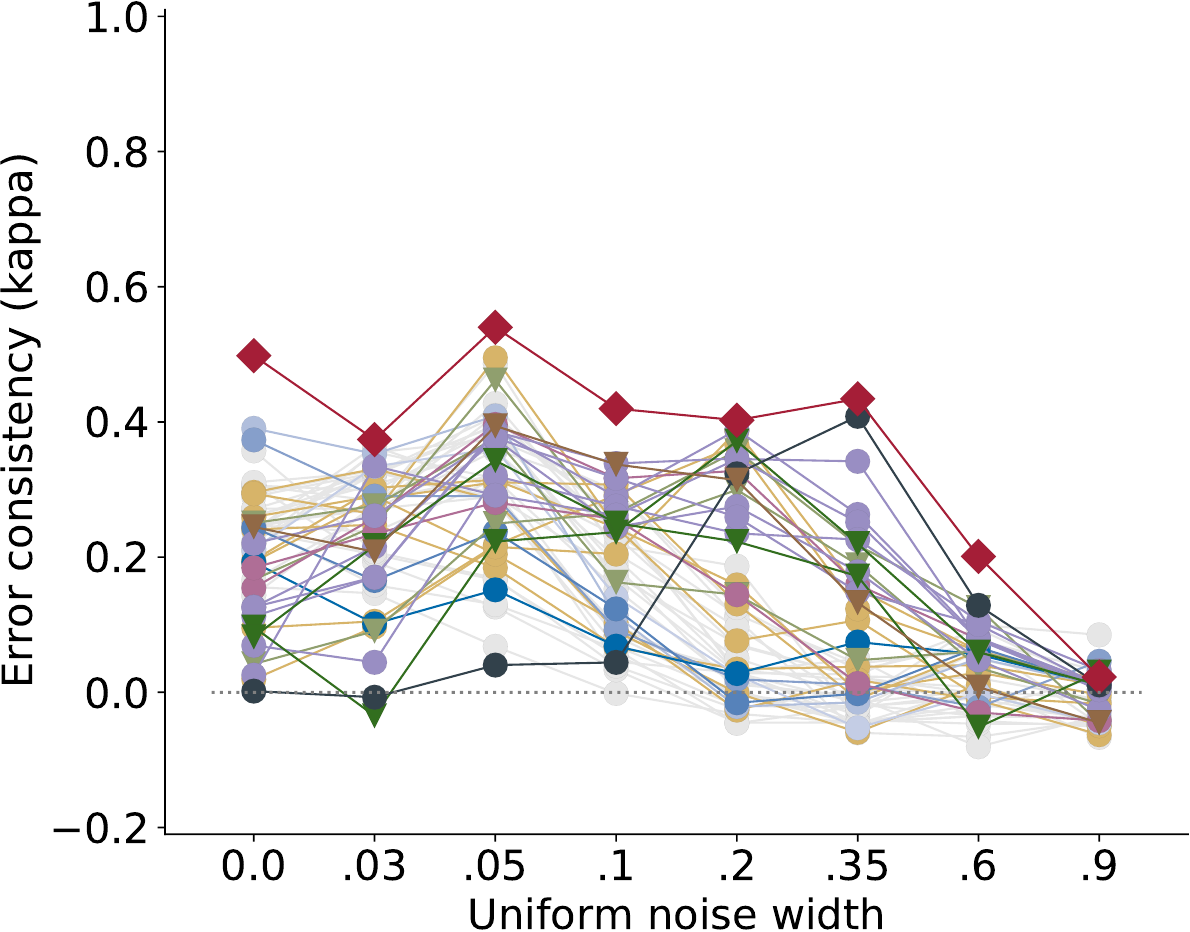}
			\vspace{\captionspace}
			\caption*{}
			\vspace{\captionspaceII}
		\end{subfigure}\hfill
		\begin{subfigure}{\figwidth}
			\centering
			\includegraphics[width=\linewidth]{low-pass_OOD-accuracy.pdf}
			\vspace{\captionspace}
			\caption{Low-pass}
			\vspace{\captionspaceII}
		\end{subfigure}\hfill
		\begin{subfigure}{\figwidth}
			\centering
			\includegraphics[width=\linewidth]{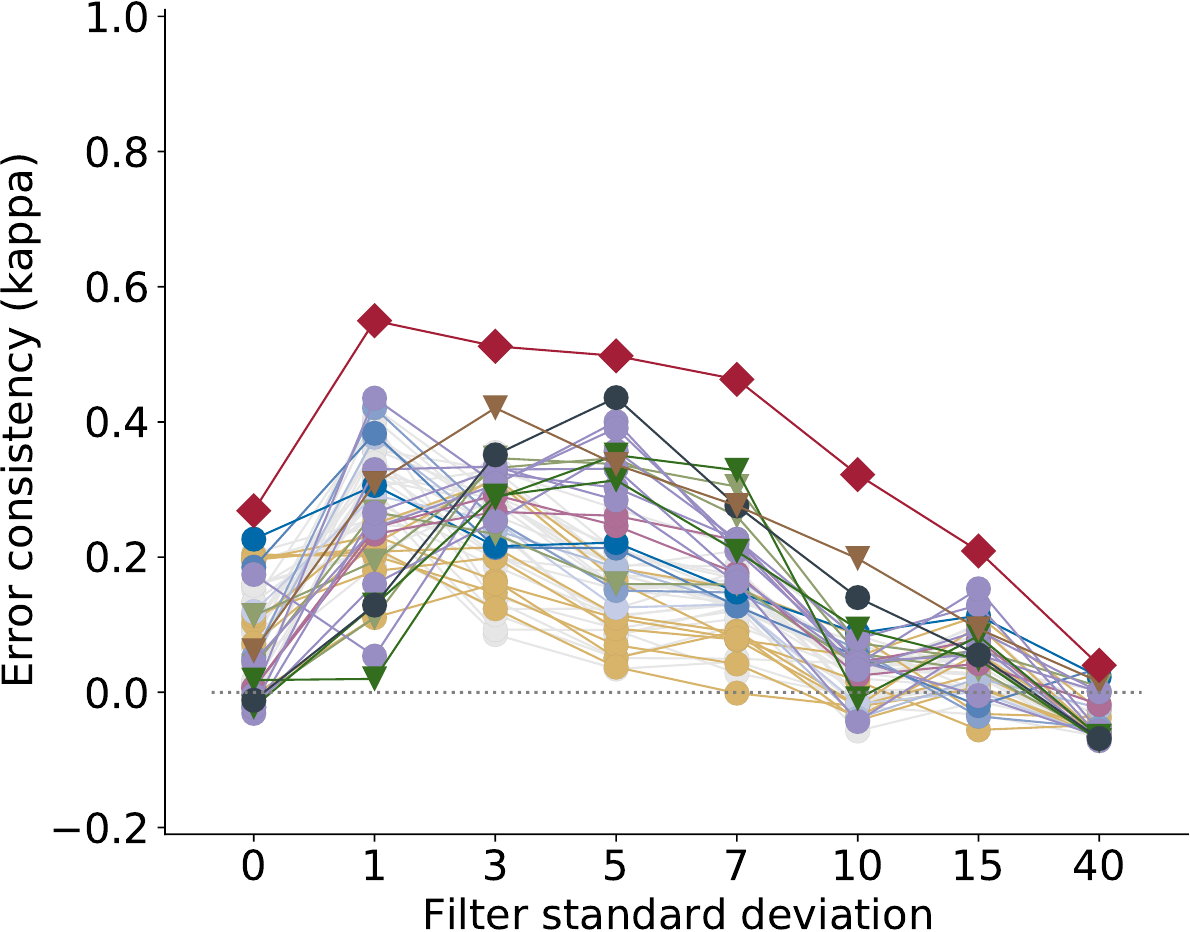}
			\vspace{\captionspace}
			\caption*{}
			\vspace{\captionspaceII}
		\end{subfigure}\hfill

		\begin{subfigure}{\figwidth}
			\centering
			\includegraphics[width=\linewidth]{contrast_OOD-accuracy.pdf}
			\vspace{\captionspace}
			\caption{Contrast}
			\vspace{\captionspaceII}
		\end{subfigure}\hfill
		\begin{subfigure}{\figwidth}
			\centering
			\includegraphics[width=\linewidth]{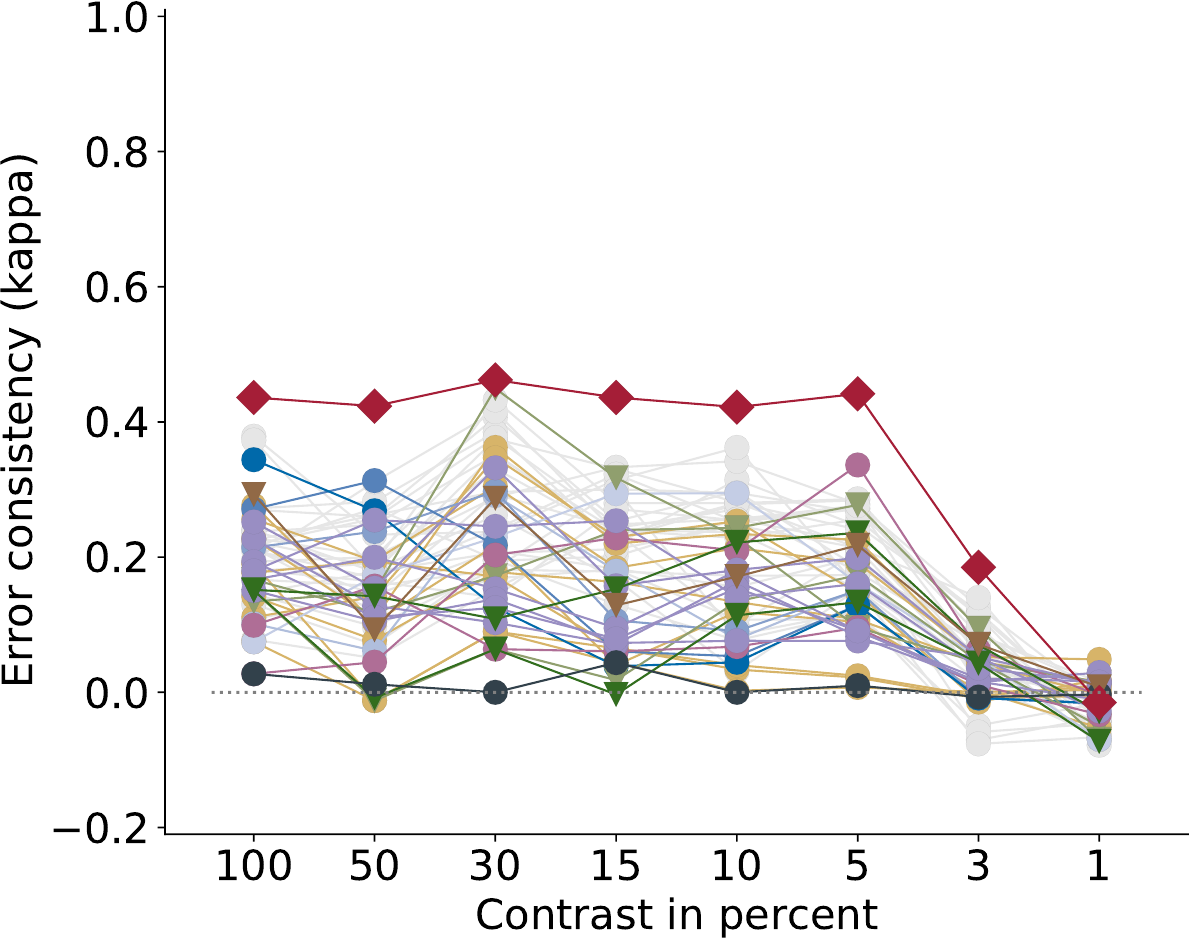}
			\vspace{\captionspace}
			\caption*{}
			\vspace{\captionspaceII}
		\end{subfigure}\hfill
		\begin{subfigure}{\figwidth}
			\centering
			\includegraphics[width=\linewidth]{high-pass_OOD-accuracy.pdf}
			\vspace{\captionspace}
			\caption{High-pass}
			\vspace{\captionspaceII}
		\end{subfigure}\hfill
		\begin{subfigure}{\figwidth}
			\centering
			\includegraphics[width=\linewidth]{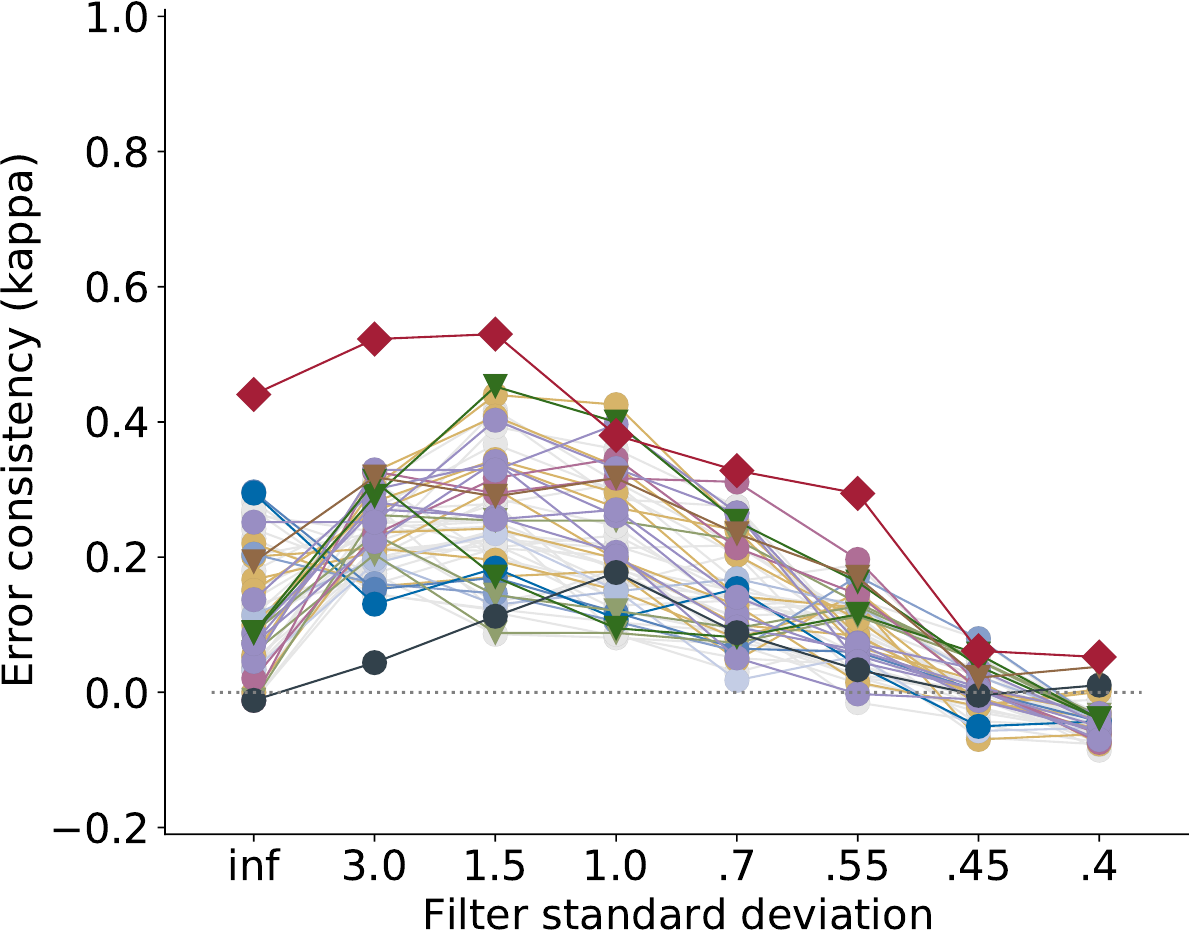}
			\vspace{\captionspace}
			\caption*{}
			\vspace{\captionspaceII}
		\end{subfigure}\hfill

		\begin{subfigure}{\figwidth}
			\centering
			\includegraphics[width=\linewidth]{eidolonI_OOD-accuracy.pdf}
			\vspace{\captionspace}
			\caption{Eidolon I}
			\vspace{\captionspaceII}
		\end{subfigure}\hfill
		\begin{subfigure}{\figwidth}
			\centering        \includegraphics[width=\linewidth]{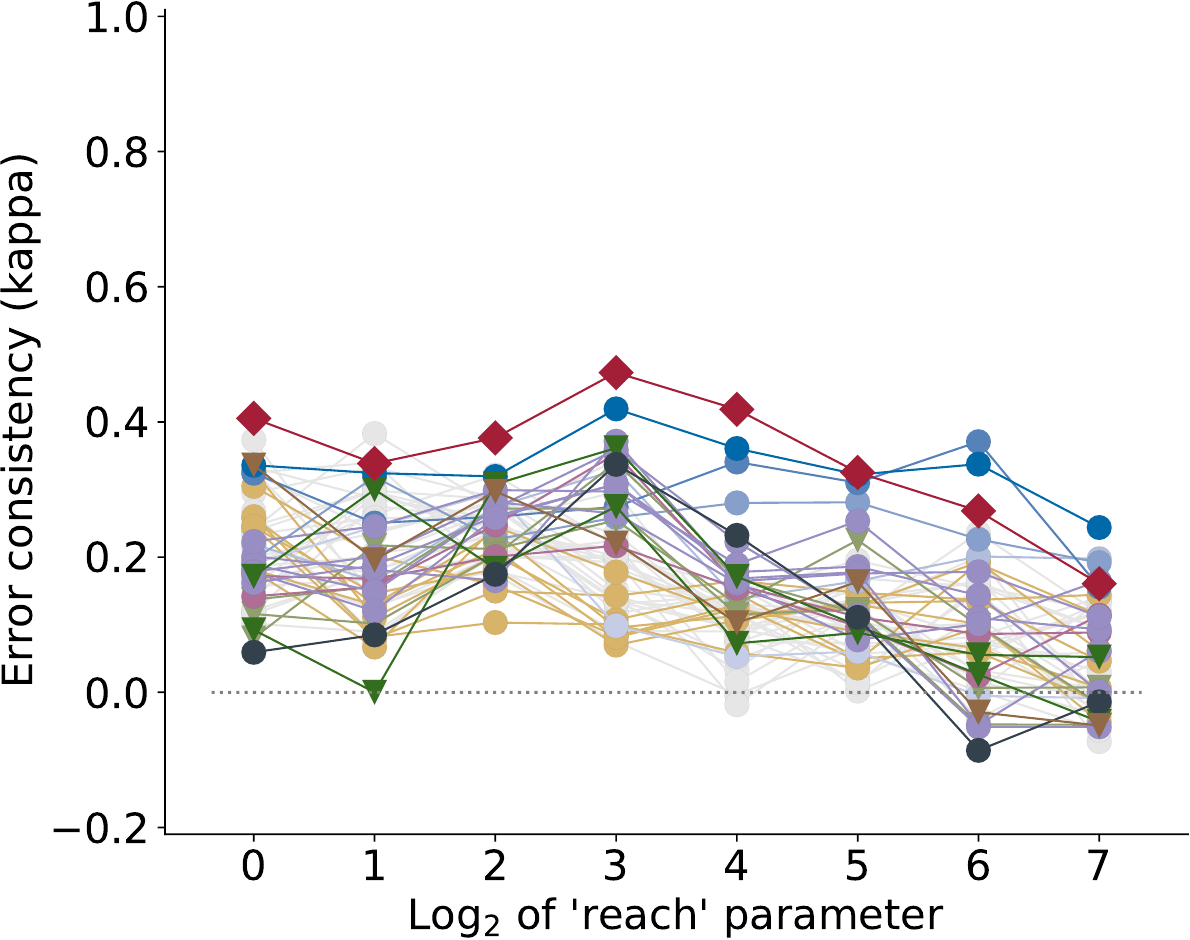}
			\vspace{\captionspace}
			\caption*{}
			\vspace{\captionspaceII}
		\end{subfigure}\hfill
		\begin{subfigure}{\figwidth}
			\centering
			\includegraphics[width=\linewidth]{phase-scrambling_OOD-accuracy.pdf}
			\vspace{\captionspace}
			\caption{Phase noise}
			\vspace{\captionspaceII}
		\end{subfigure}\hfill
		\begin{subfigure}{\figwidth}
			\centering
			\includegraphics[width=\linewidth]{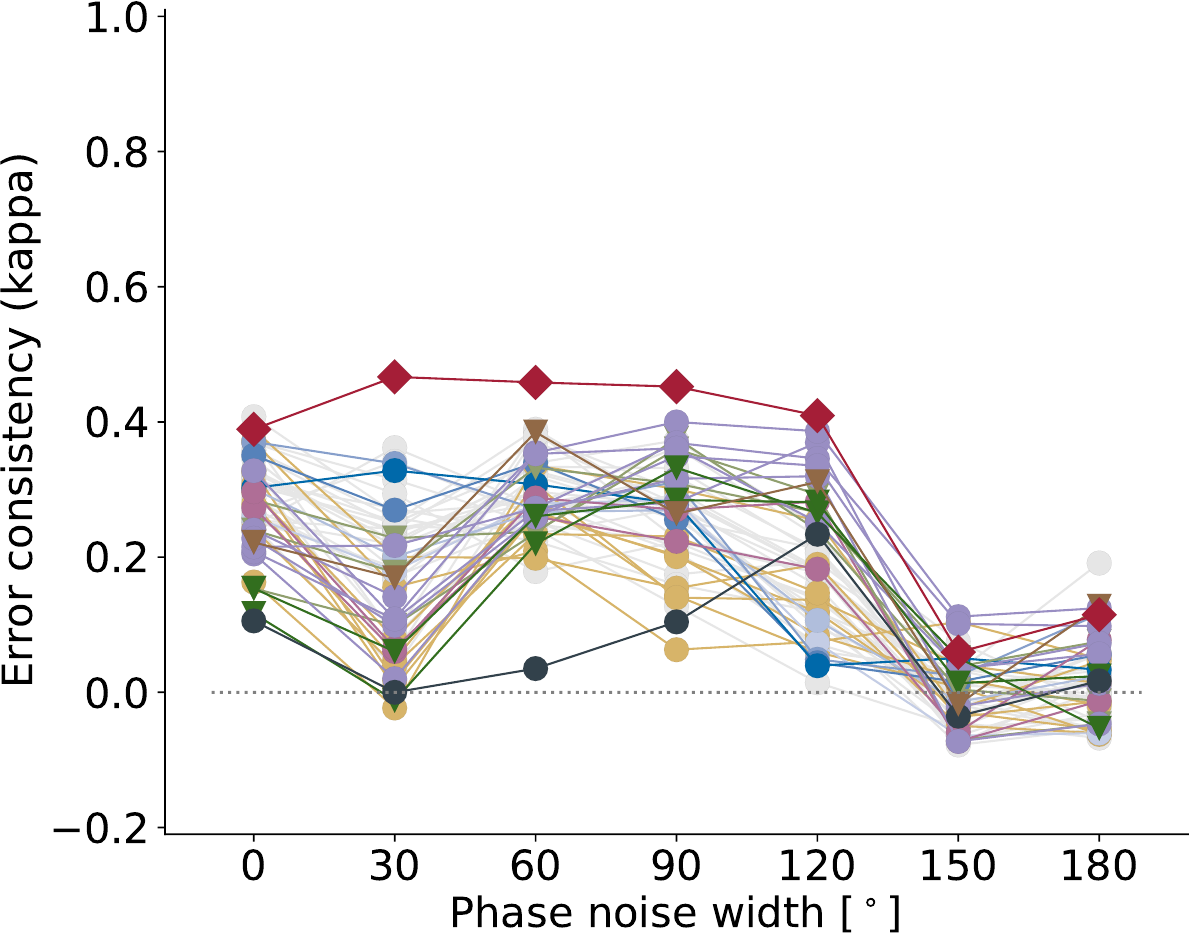}
			\vspace{\captionspace}
			\caption*{}
			\vspace{\captionspaceII}
		\end{subfigure}\hfill

	\begin{subfigure}{\figwidth}
		\centering
		\includegraphics[width=\linewidth]{eidolonII_OOD-accuracy.pdf}
		\vspace{\captionspace}
		\caption{Eidolon II}
		\vspace{\captionspaceII}
	\end{subfigure}\hfill
	\begin{subfigure}{\figwidth}
		\centering
		\includegraphics[width=\linewidth]{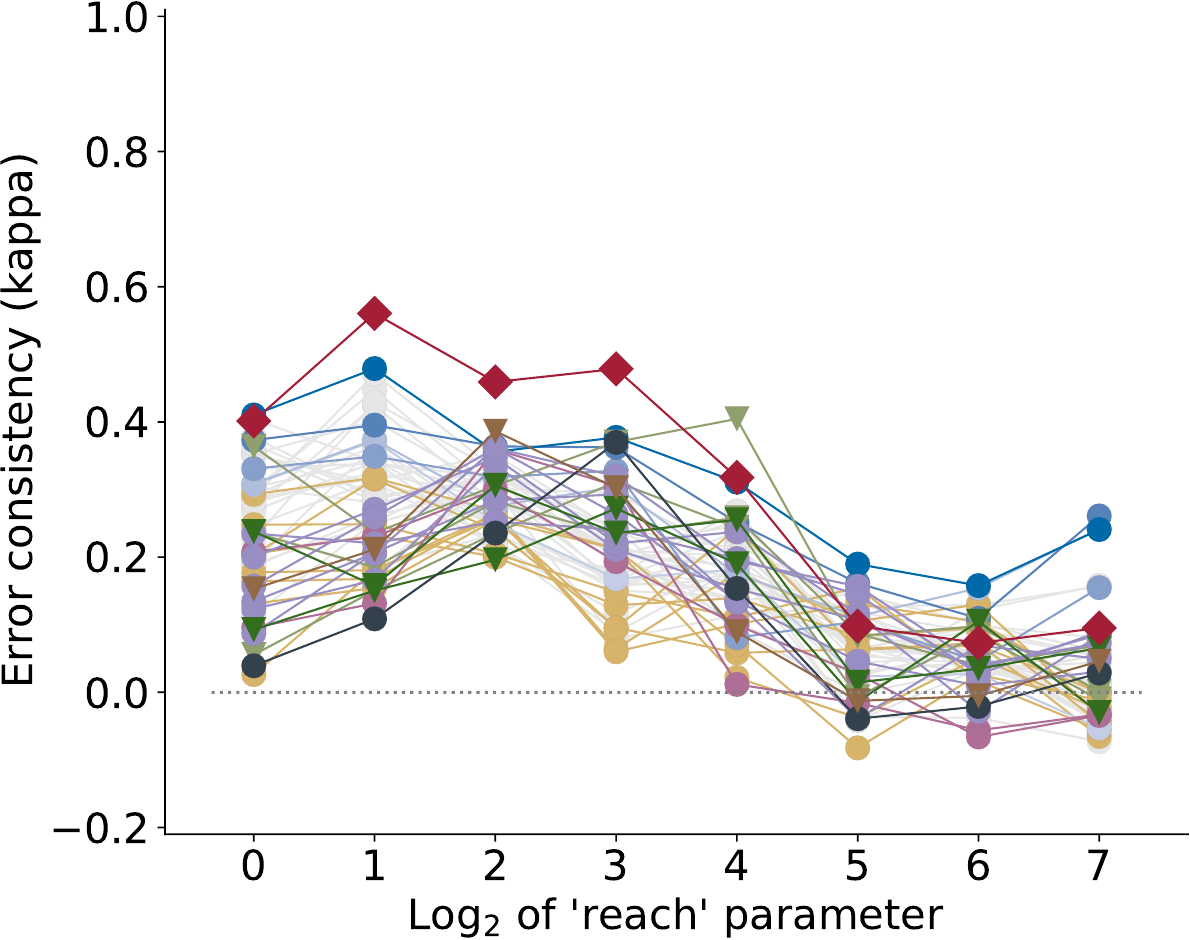}
		\vspace{\captionspace}
		\caption*{}
		\vspace{\captionspaceII}
	\end{subfigure}\hfill
	\begin{subfigure}{\figwidth}
		\centering
		\includegraphics[width=\linewidth]{power-equalisation_OOD-accuracy.pdf}
		\vspace{\captionspace}
		\caption{Power equalisation}
		\vspace{\captionspaceII}
	\end{subfigure}\hfill
	\begin{subfigure}{\figwidth}
		\centering
		\includegraphics[width=\linewidth]{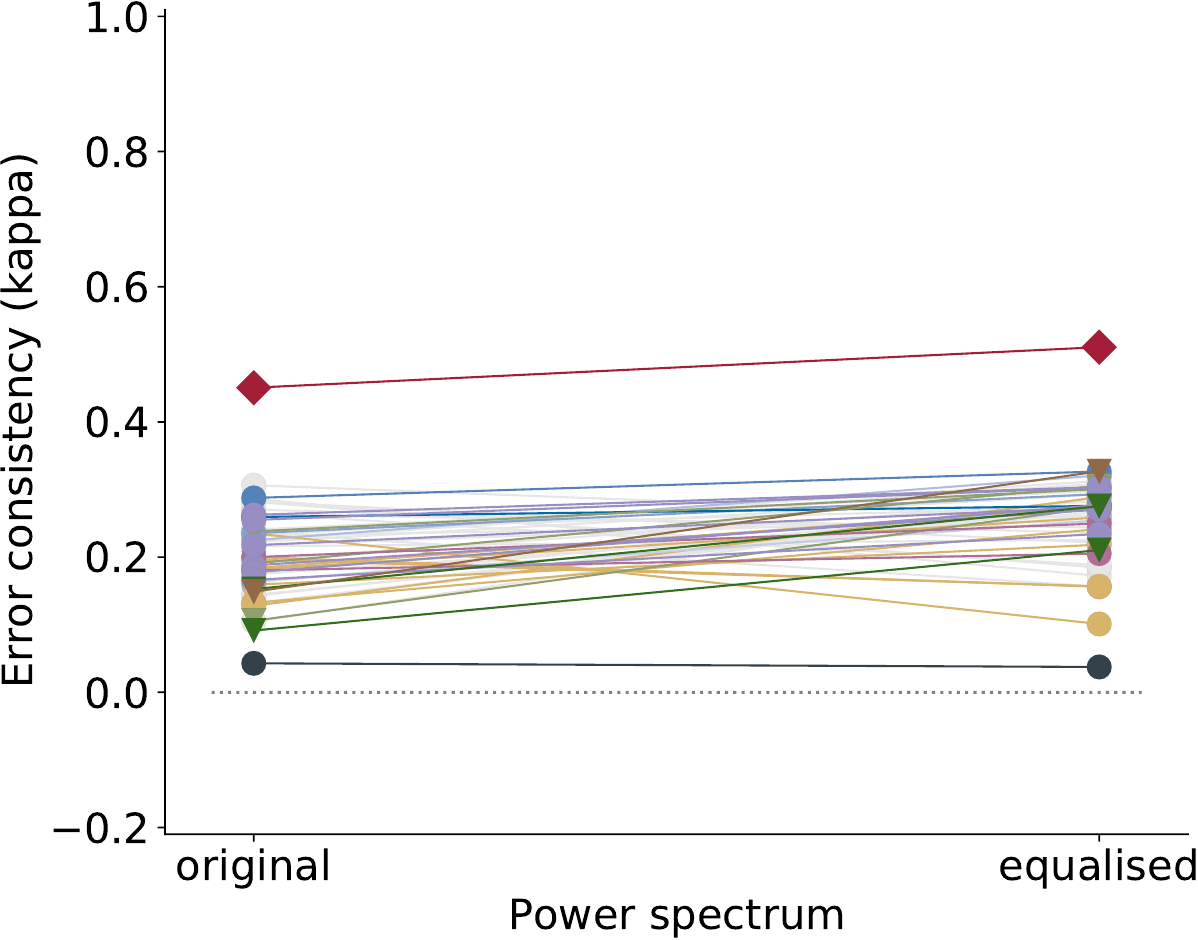}
		\vspace{\captionspace}
		\caption*{}
		\vspace{\captionspaceII}
	\end{subfigure}\hfill

	\begin{subfigure}{\figwidth}
		\centering
		\includegraphics[width=\linewidth]{eidolonIII_OOD-accuracy.pdf}
		\vspace{\captionspace}
		\caption{Eidolon III}
	\end{subfigure}\hfill
	\begin{subfigure}{\figwidth}
		\centering        \includegraphics[width=\linewidth]{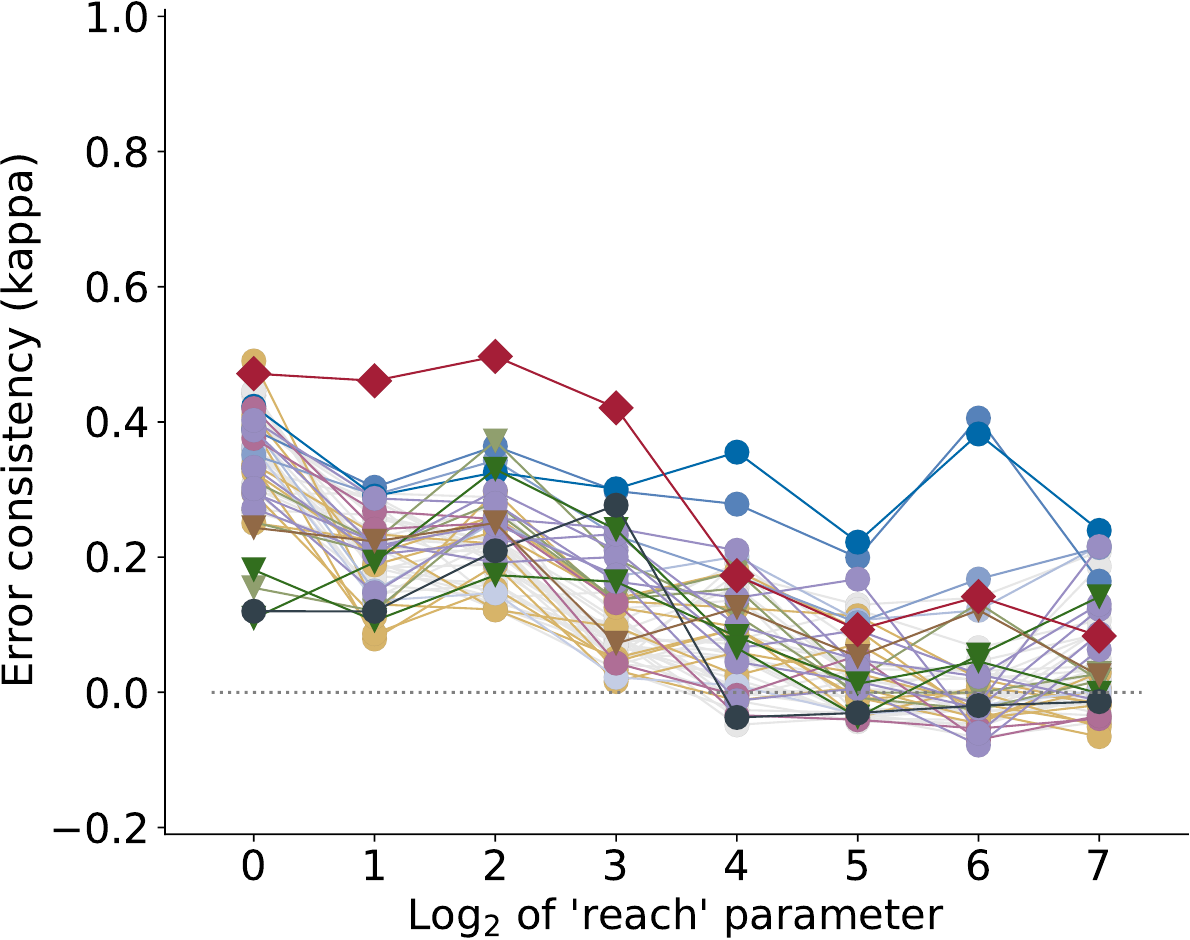}
		\vspace{\captionspace}
		\caption*{}
	\end{subfigure}\hfill
	\begin{subfigure}{\figwidth}
		\centering
		\includegraphics[width=\linewidth]{rotation_OOD-accuracy.pdf}
		\vspace{\captionspace}
		\caption{Rotation}
	\end{subfigure}\hfill
	\begin{subfigure}{\figwidth}
		\centering
		\includegraphics[width=\linewidth]{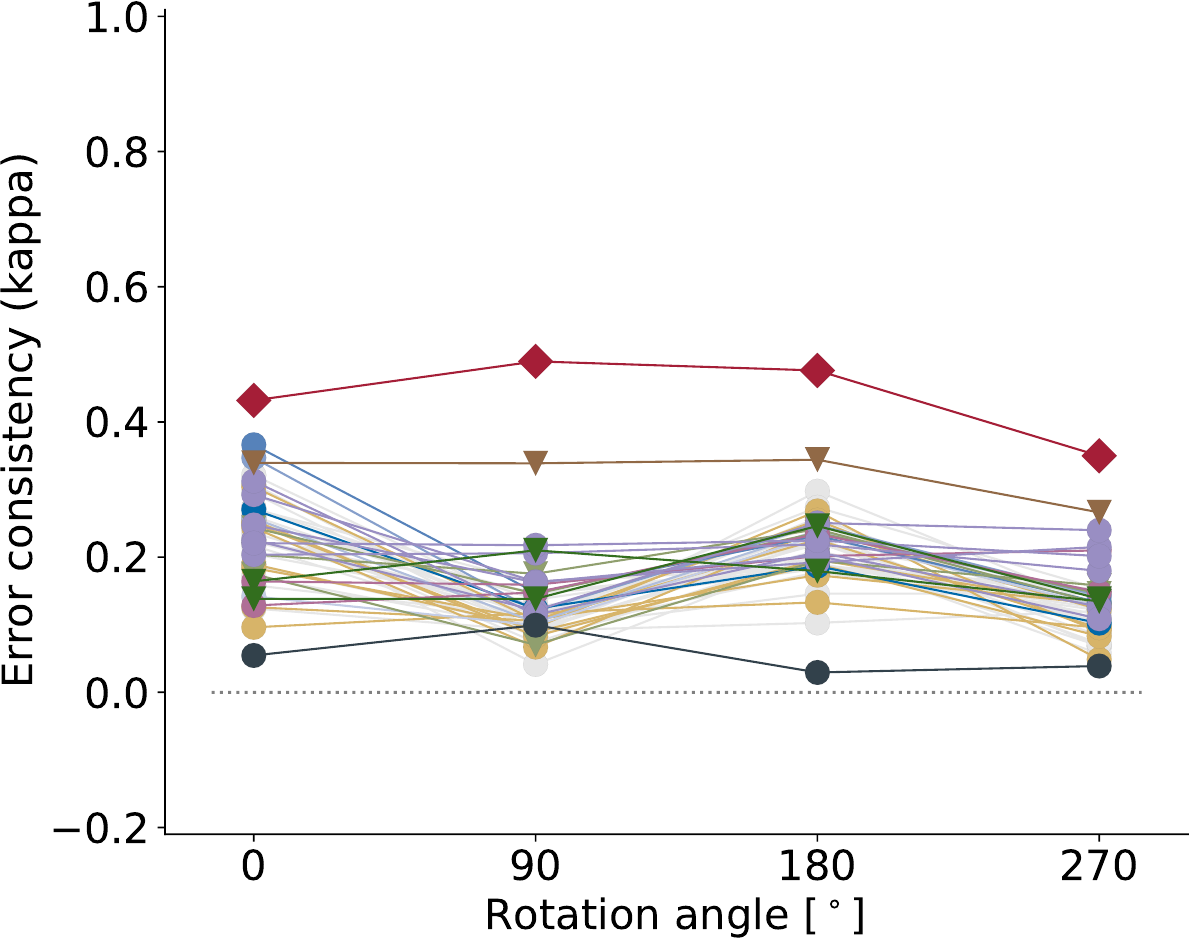}
		\vspace{\captionspace}
		\caption*{}
	\end{subfigure}\hfill
	\caption{OOD generalisation and error consistency results for \textcolor{red}{humans}, \textcolor{supervised.grey}{standard supervised CNNs}, \textcolor{orange1}{self-supervised models}, \textcolor{blue1}{adversarially trained models}, \textcolor{green1}{vision transformers}, \textcolor{black}{noisy student}, \textcolor{bitm.purple2}{BiT}, \textcolor{purple1}{SWSL}, \textcolor{brown1}{CLIP}. Symbols indicate architecture type ($\ocircle$ convolutional, $\triangledown$ vision transformer, $\lozenge$ human); best viewed on screen. `Accuracy' measures recognition performance (higher is better), `error consistency' how closely image-level errors are aligned with humans. Accuracy results are identical to Figure~\ref{fig:results_accuracy} in the main paper. In many cases, human-to-human error consistency increases for moderate distortion levels and drops afterwards.}
	\label{fig:results_accuracy_error_consistency}
\end{figure}

\begin{figure*}
    \centering
    \includegraphics[width=0.8\linewidth]{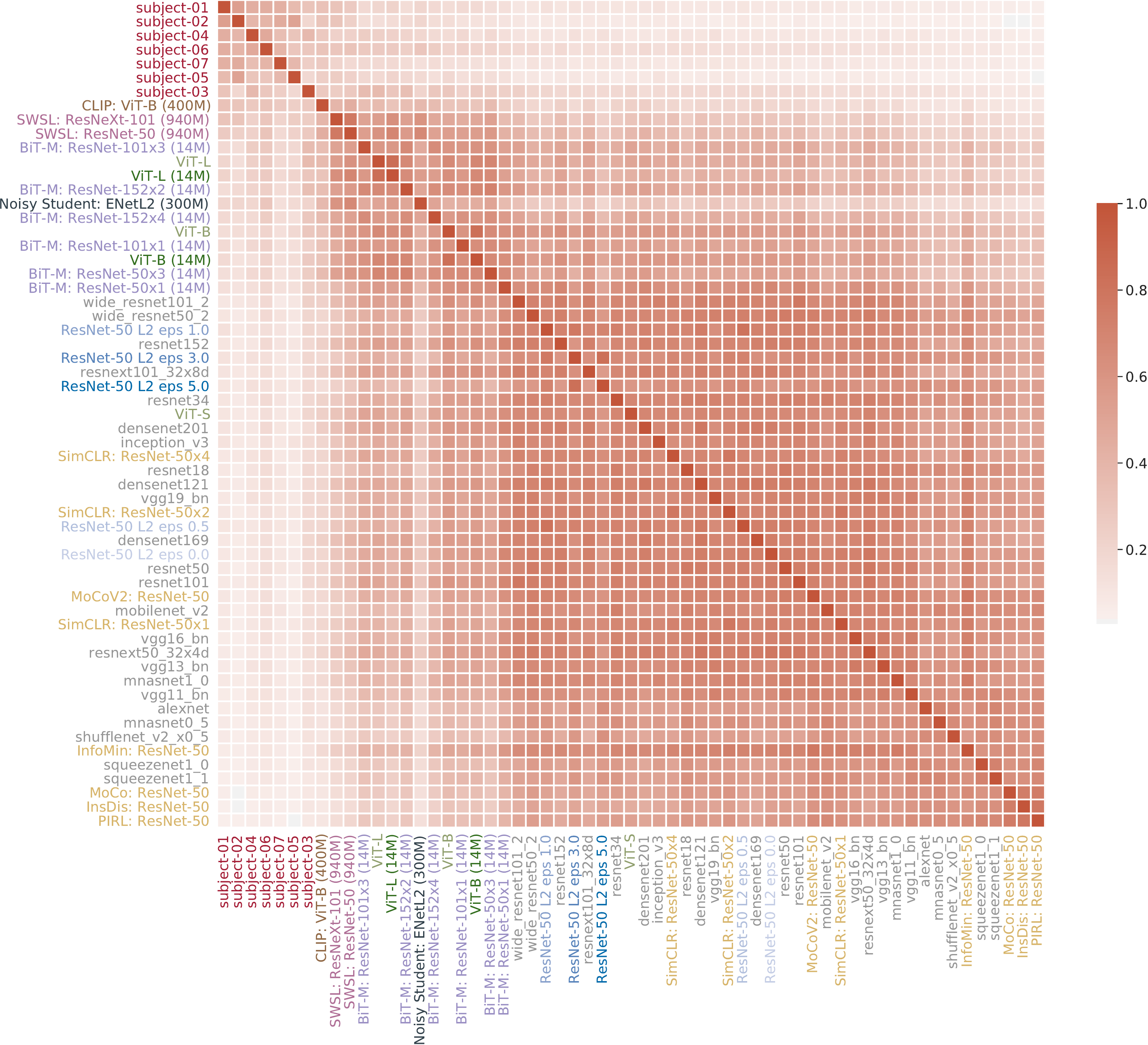}
    \caption{Error consistency for `sketch' images (same as Figure~\ref{fig:error_consistency_matrix_sketch} but sorted w.r.t.\ mean error consistency with humans).}
    \label{fig:error_consistency_matrix_sketch_by_mean}
\end{figure*}

\begin{figure*}
    \centering
    \includegraphics[width=0.8\linewidth]{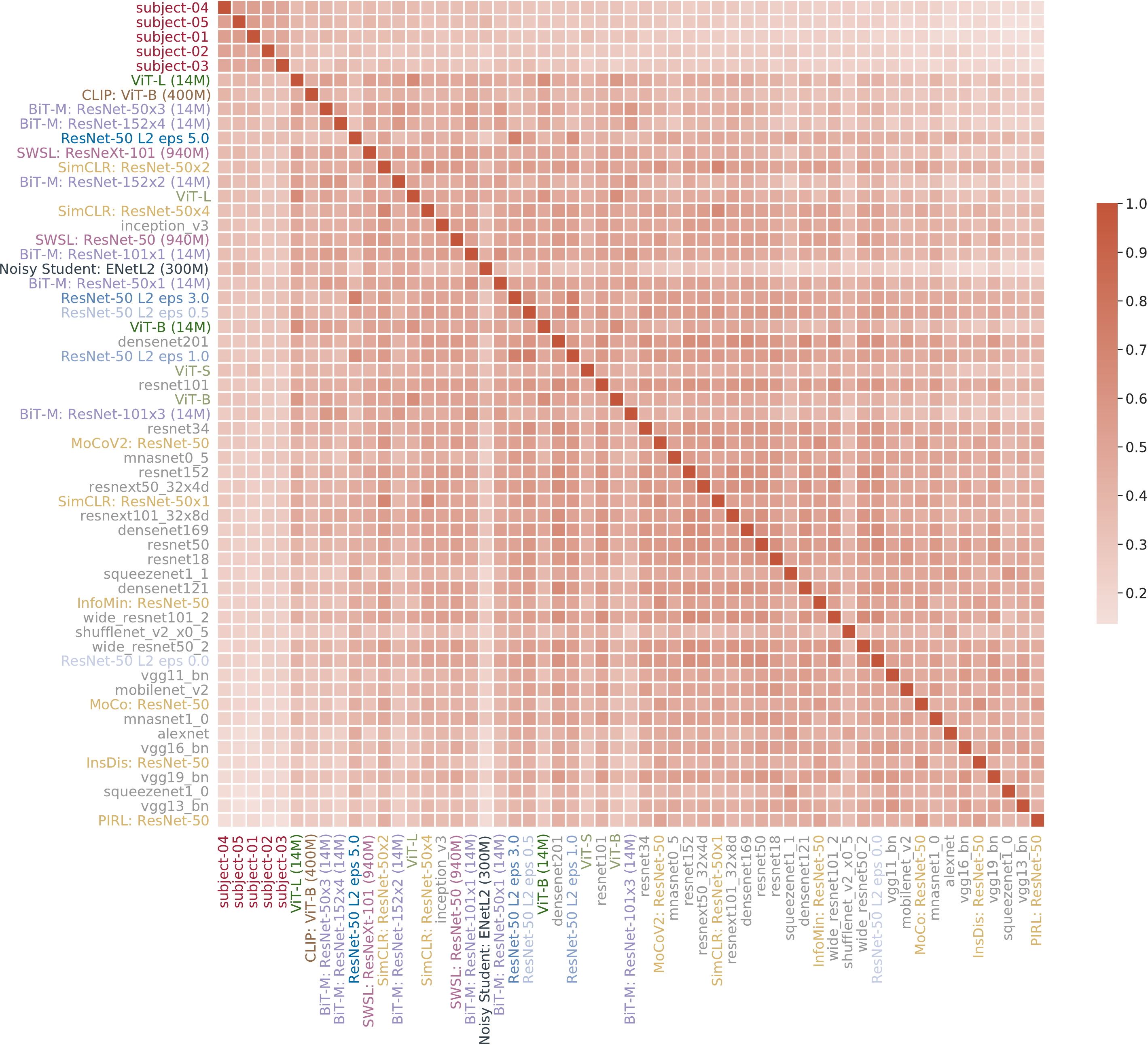}
    \caption{Error consistency for `stylized' images (sorted w.r.t.\ mean error consistency with humans).}
    \label{fig:error_consistency_matrix_stylized_by_mean}
\end{figure*}

\begin{figure*}
	\centering
	\includegraphics[width=0.8\linewidth]{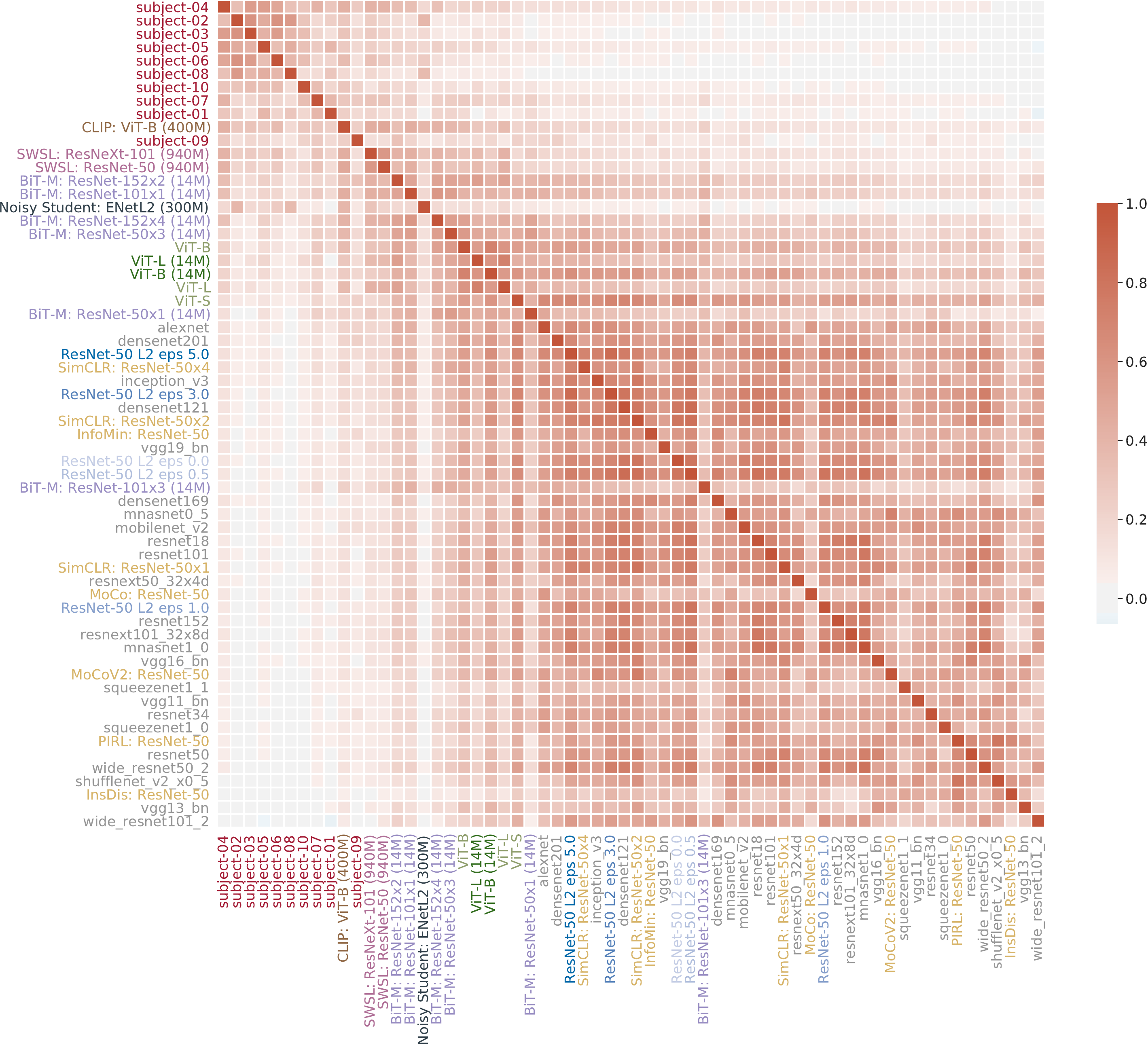}
	\caption{Error consistency for `edge' images (sorted w.r.t.\ mean error consistency with humans).}
	\label{fig:error_consistency_matrix_edges_by_mean}
\end{figure*}

\begin{figure*}
	\centering
	\includegraphics[width=0.8\linewidth]{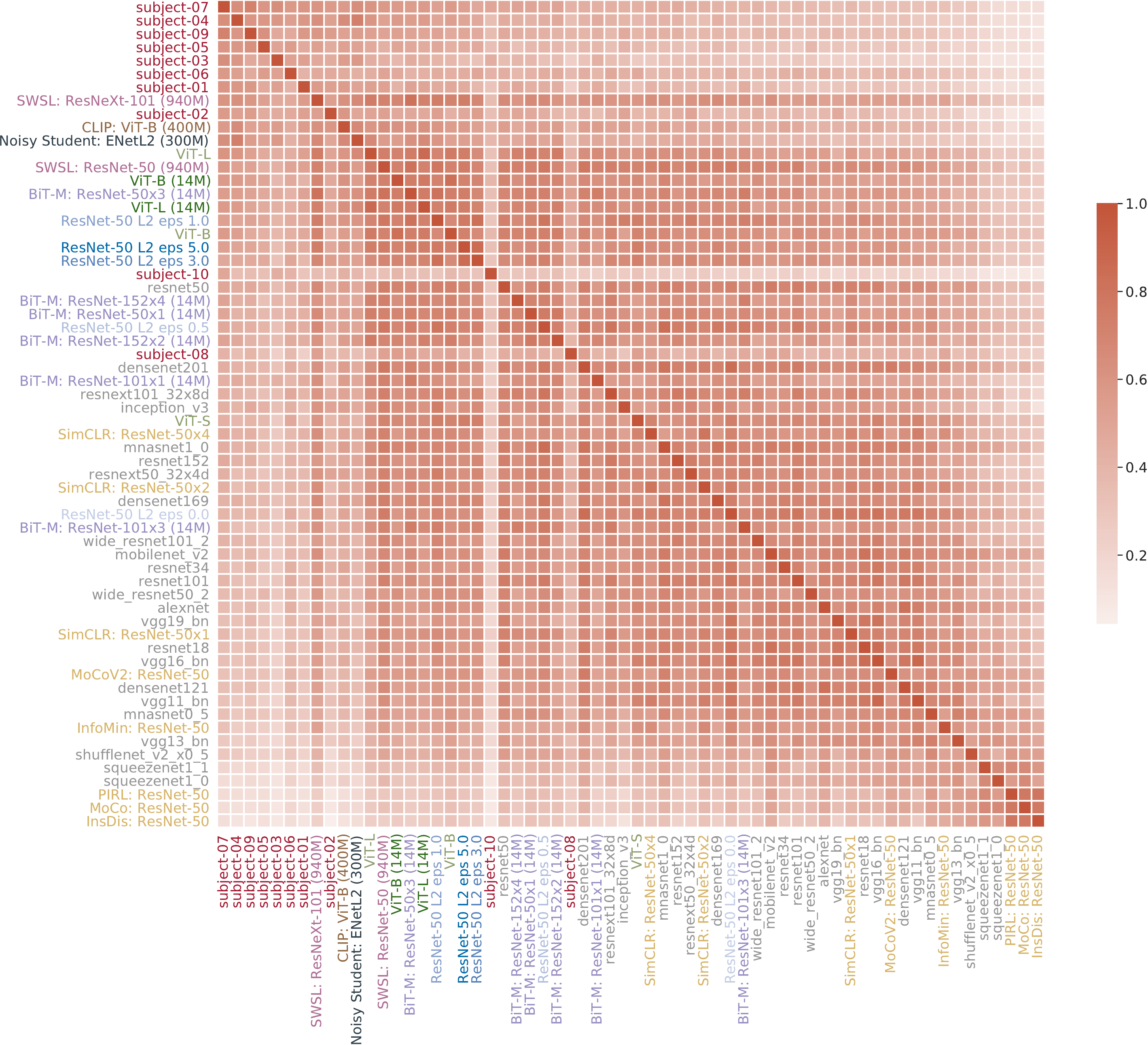}
	\caption{Error consistency for `silhouette' images (sorted w.r.t.\ mean error consistency with humans).}
	\label{fig:error_consistency_matrix_silhouettes_by_mean}
\end{figure*}

\begin{figure*}
	\centering
	\includegraphics[width=0.8\linewidth]{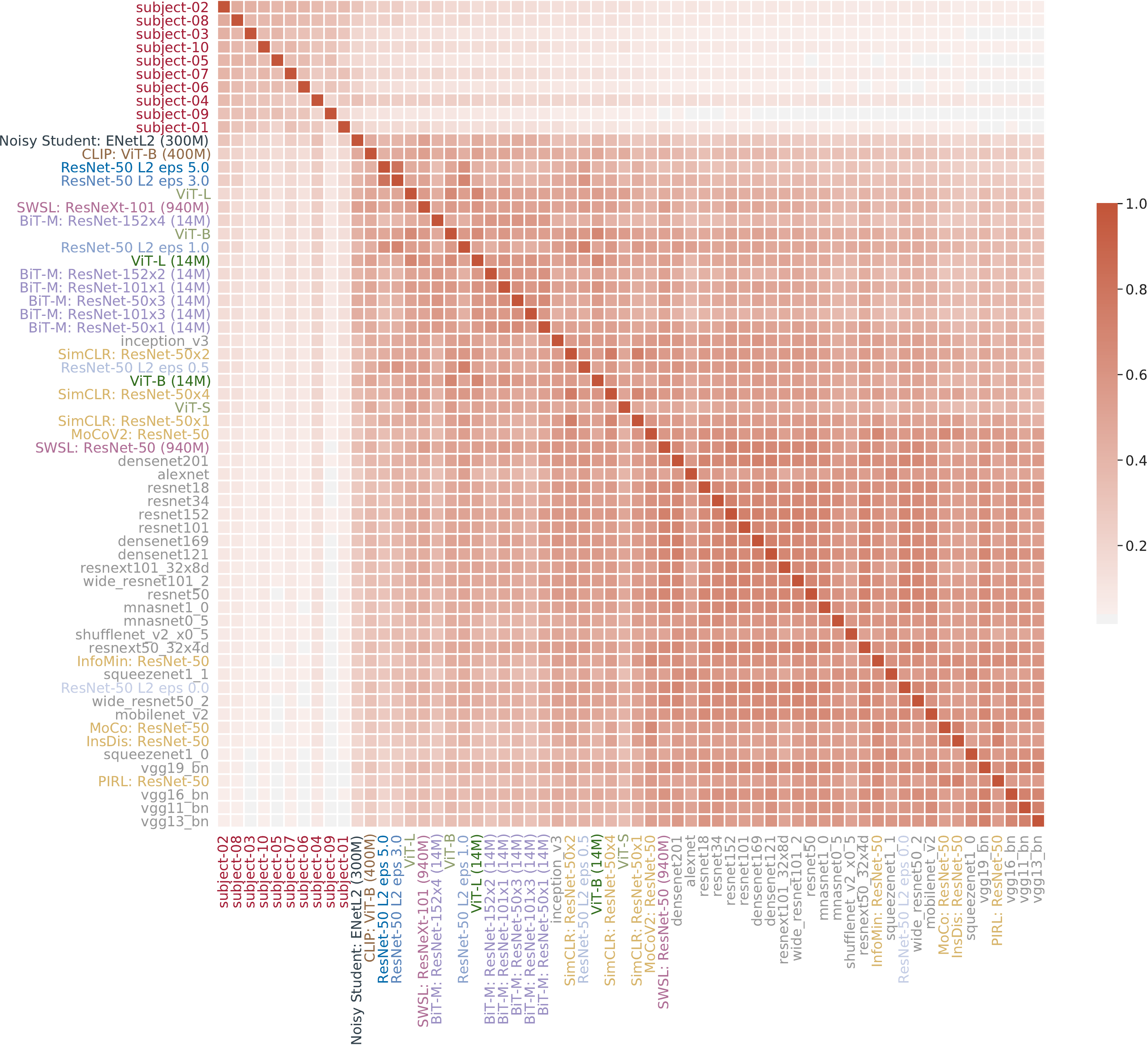}
	\caption{Error consistency for `cue conflict' images (sorted w.r.t.\ mean error consistency with humans).}
	\label{fig:error_consistency_matrix_cue-conflict_by_mean}
\end{figure*}

\section{Supervised SimCLR baseline models}
\label{app:supervised_simclr_baseline_models}
\begin{figure}
	\begin{subfigure}{\figwidth}
			\centering
			\textbf{Accuracy}\\
			\includegraphics[width=\linewidth]{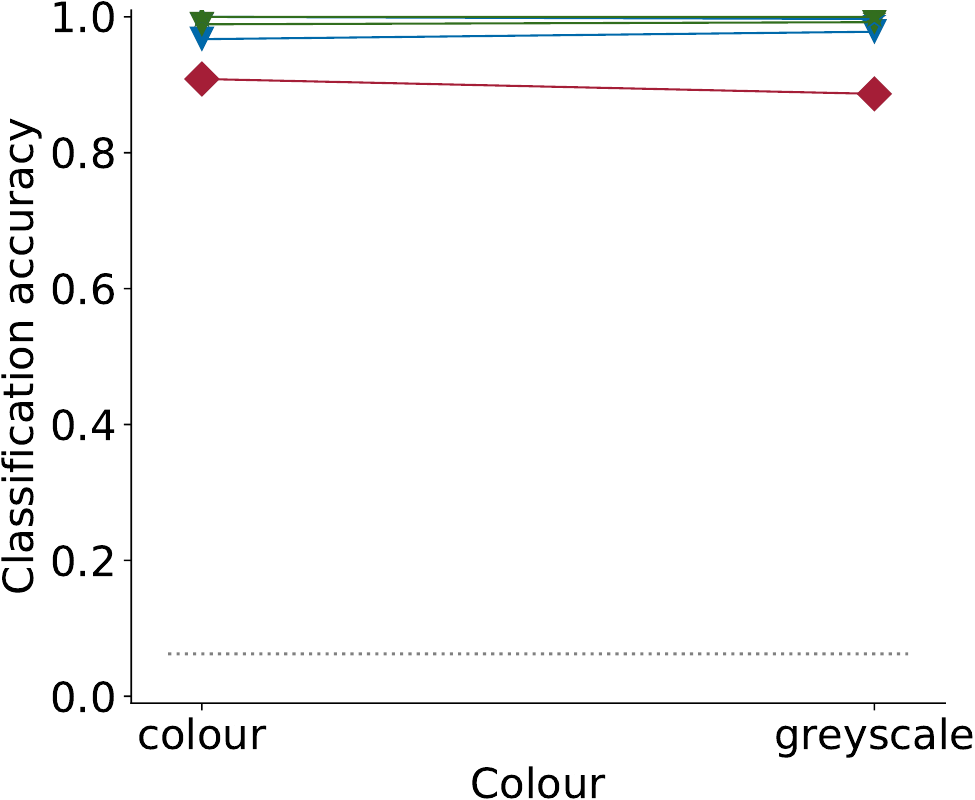}
			\vspace{\captionspace}
			\caption{Colour vs. greyscale}
			\vspace{\captionspaceII}
		\end{subfigure}\hfill
		\begin{subfigure}{\figwidth}
			\centering
			\textbf{Error consistency}\\
			\includegraphics[width=\linewidth]{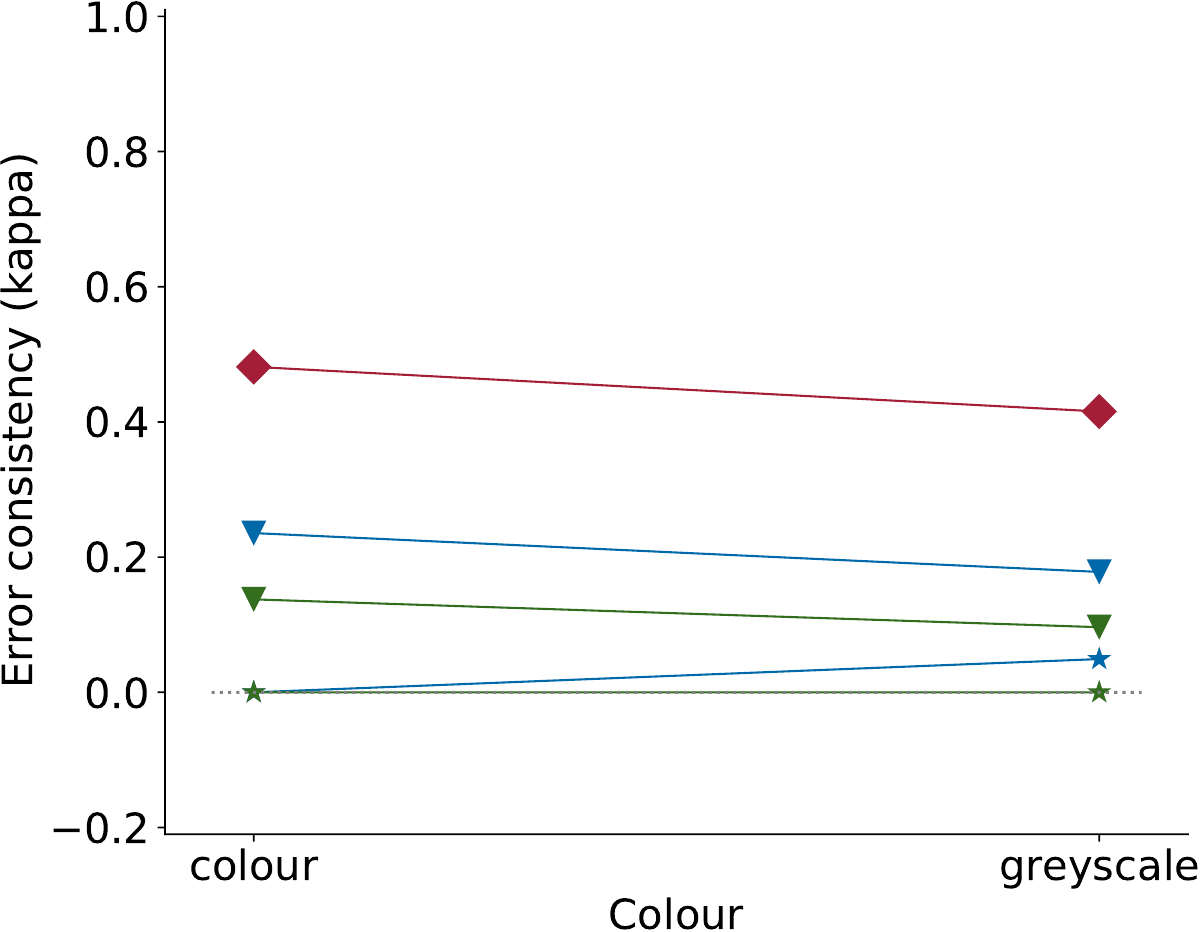}
			\vspace{\captionspace}
			\caption*{}
			\vspace{\captionspaceII}
		\end{subfigure}\hfill
		\begin{subfigure}{\figwidth}
			\centering
			\textbf{Accuracy}\\
			\includegraphics[width=\linewidth]{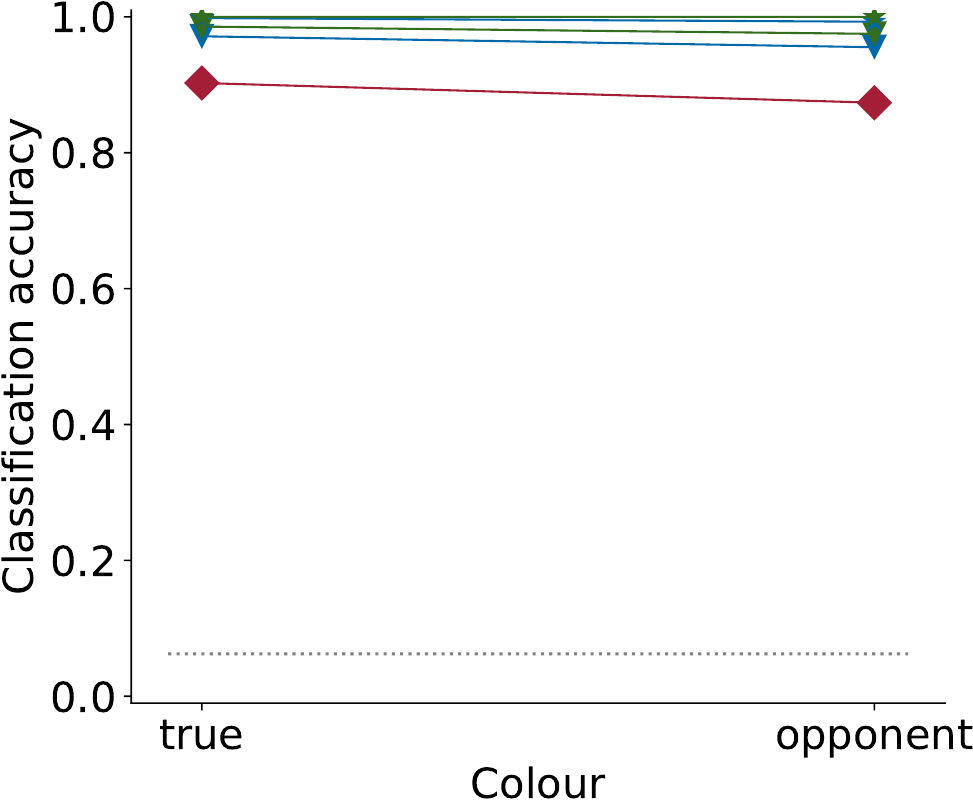}
			\vspace{\captionspace}
			\caption{True vs. false colour}
			\vspace{\captionspaceII}
		\end{subfigure}\hfill
		\begin{subfigure}{\figwidth}
			\centering
			\textbf{Error consistency}\\
			\includegraphics[width=\linewidth]{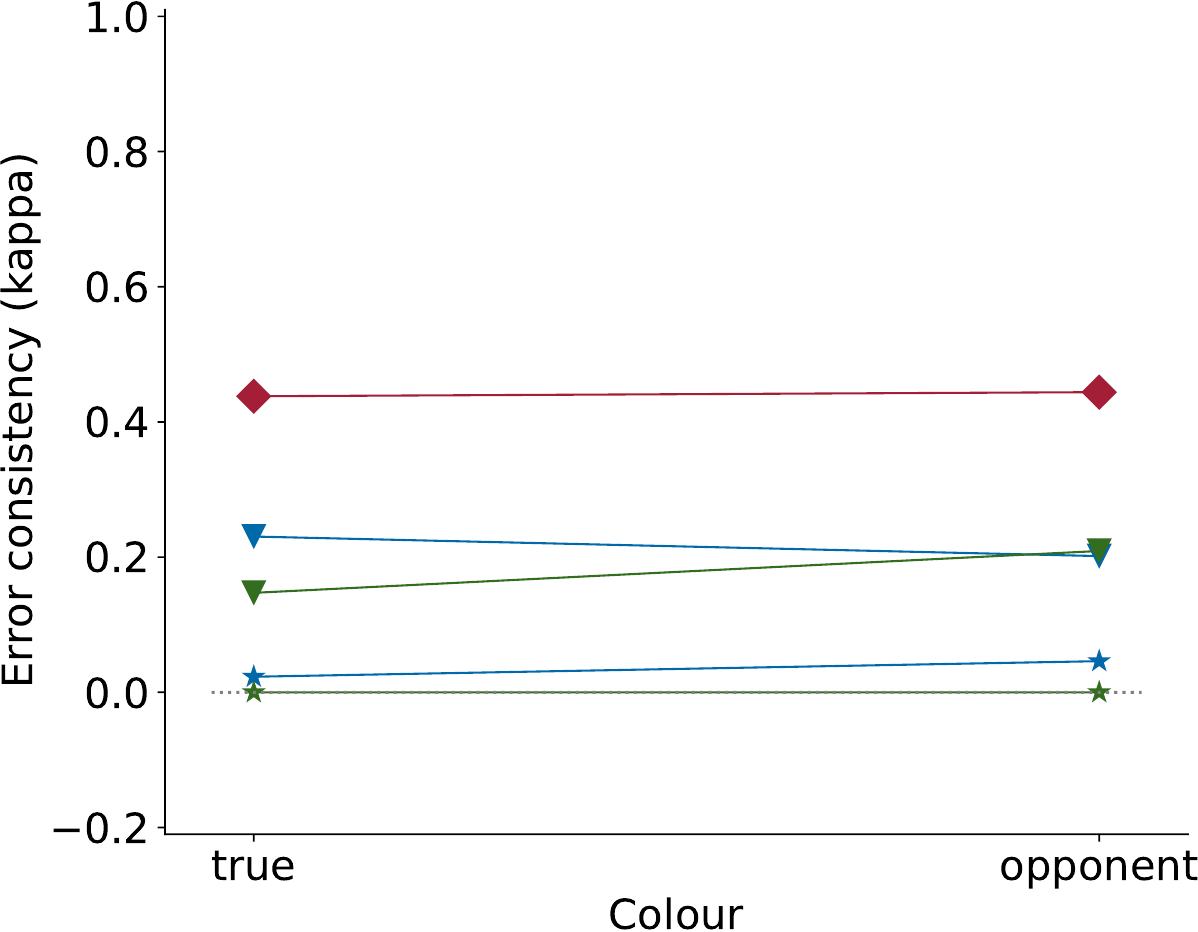}
			\vspace{\captionspace}
			\caption*{}
			\vspace{\captionspaceII}
		\end{subfigure}\hfill

		\begin{subfigure}{\figwidth}
			\centering
			\includegraphics[width=\linewidth]{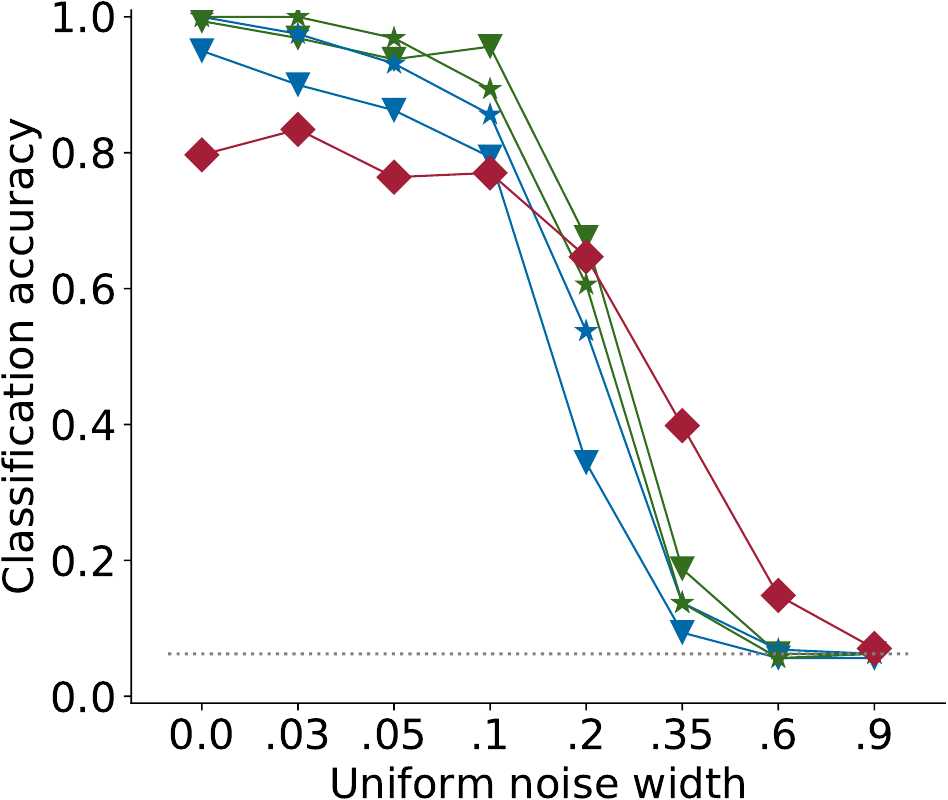}
			\vspace{\captionspace}
			\caption{Uniform noise}
			\vspace{\captionspaceII}
		\end{subfigure}\hfill
		\begin{subfigure}{\figwidth}
			\centering
			\includegraphics[width=\linewidth]{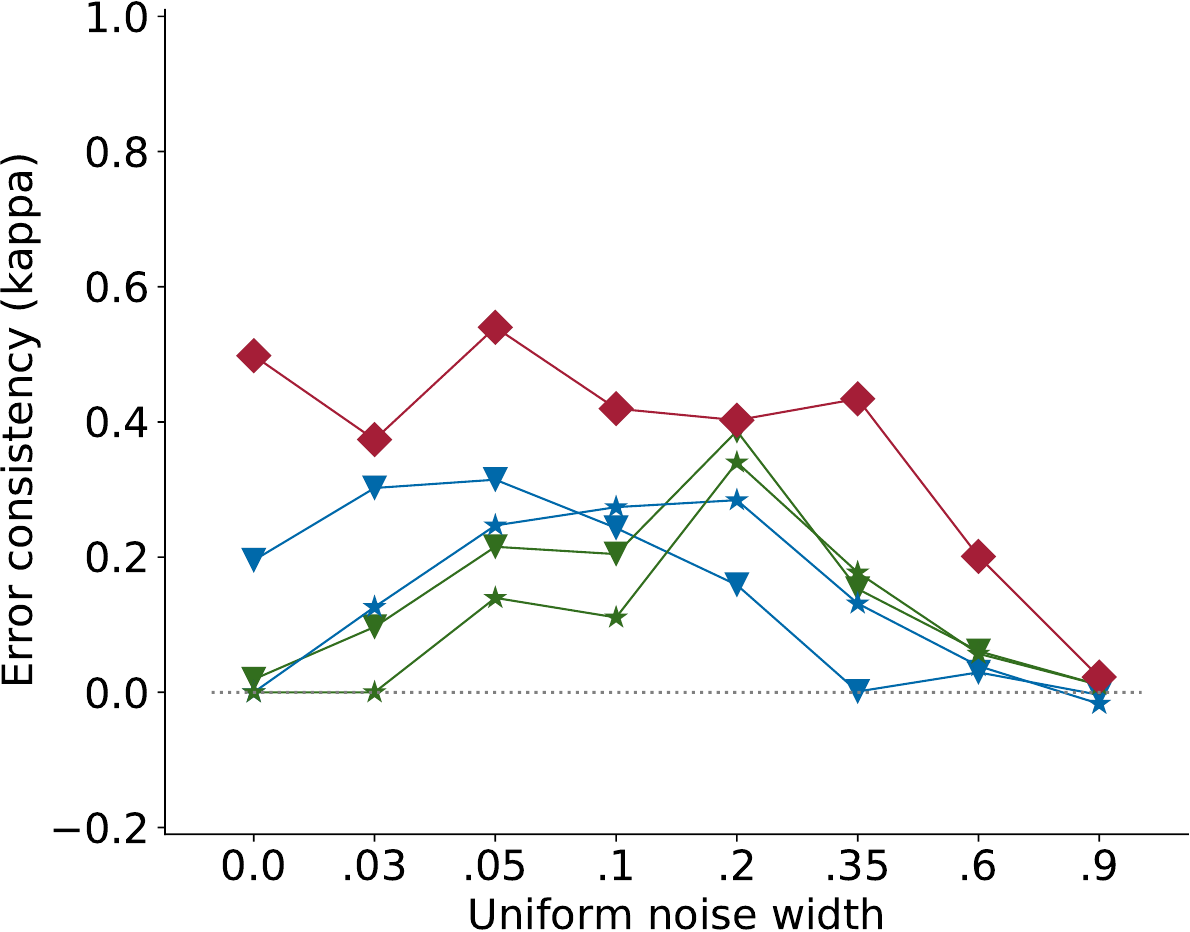}
			\vspace{\captionspace}
			\caption*{}
			\vspace{\captionspaceII}
		\end{subfigure}\hfill
		\begin{subfigure}{\figwidth}
			\centering
			\includegraphics[width=\linewidth]{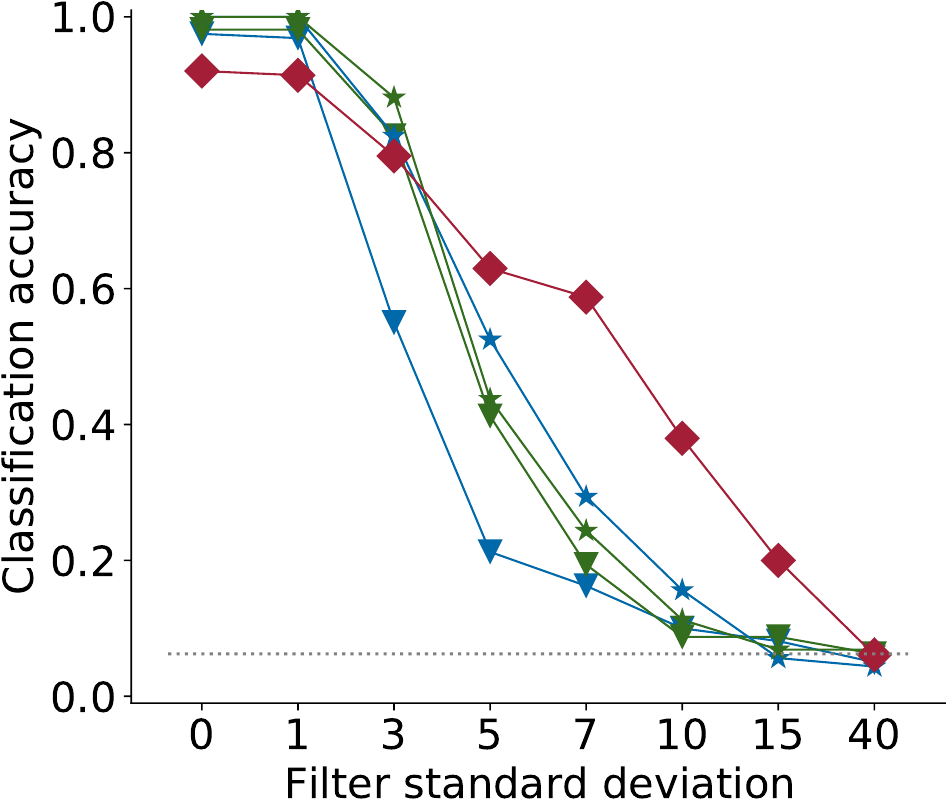}
			\vspace{\captionspace}
			\caption{Low-pass}
			\vspace{\captionspaceII}
		\end{subfigure}\hfill
		\begin{subfigure}{\figwidth}
			\centering
			\includegraphics[width=\linewidth]{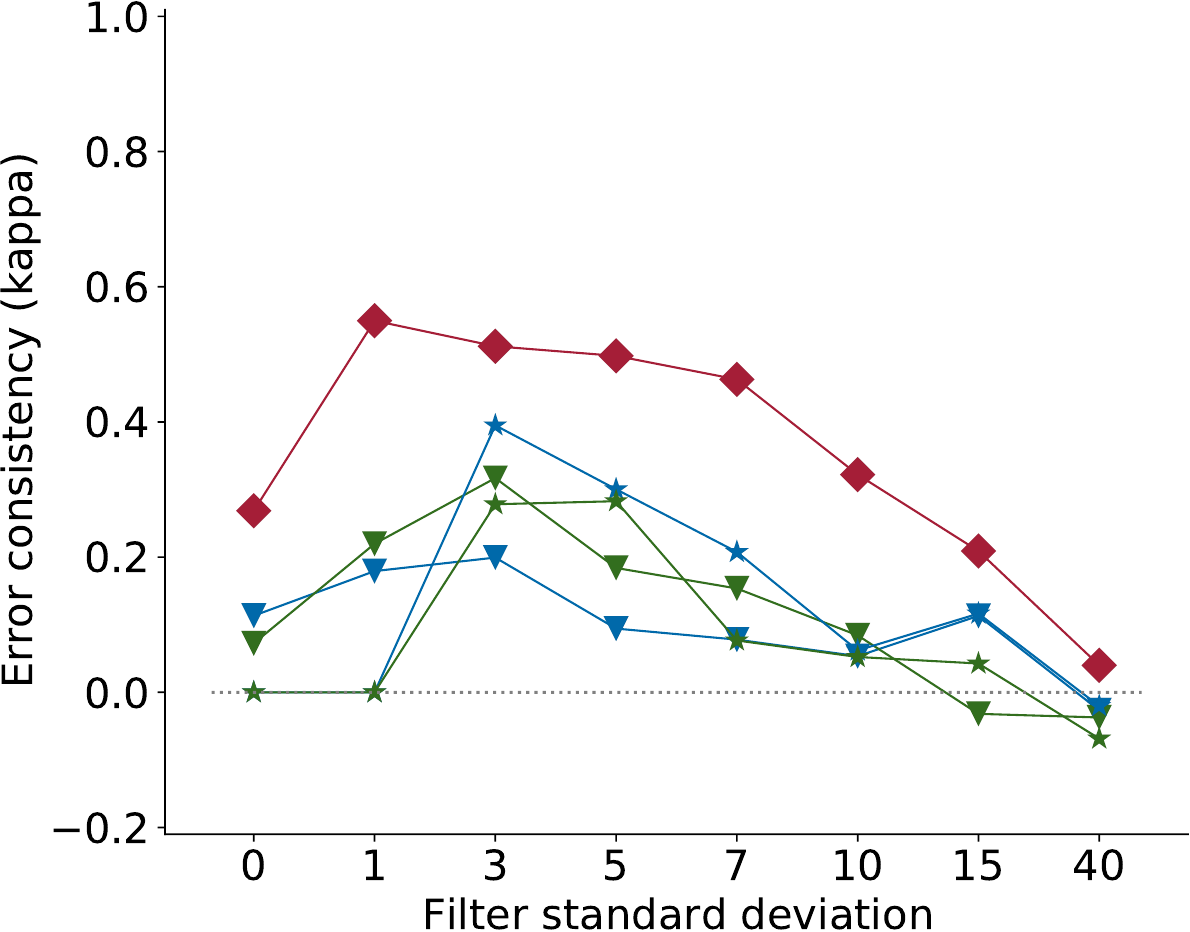}
			\vspace{\captionspace}
			\caption*{}
			\vspace{\captionspaceII}
		\end{subfigure}\hfill

		\begin{subfigure}{\figwidth}
			\centering
			\includegraphics[width=\linewidth]{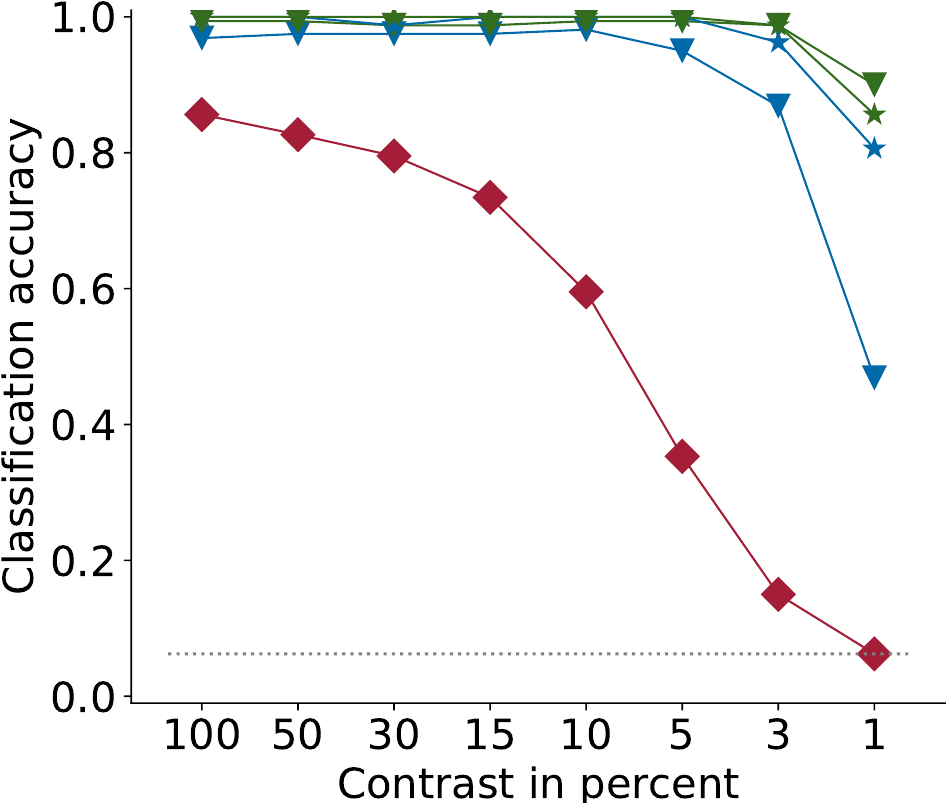}
			\vspace{\captionspace}
			\caption{Contrast}
			\vspace{\captionspaceII}
		\end{subfigure}\hfill
		\begin{subfigure}{\figwidth}
			\centering
			\includegraphics[width=\linewidth]{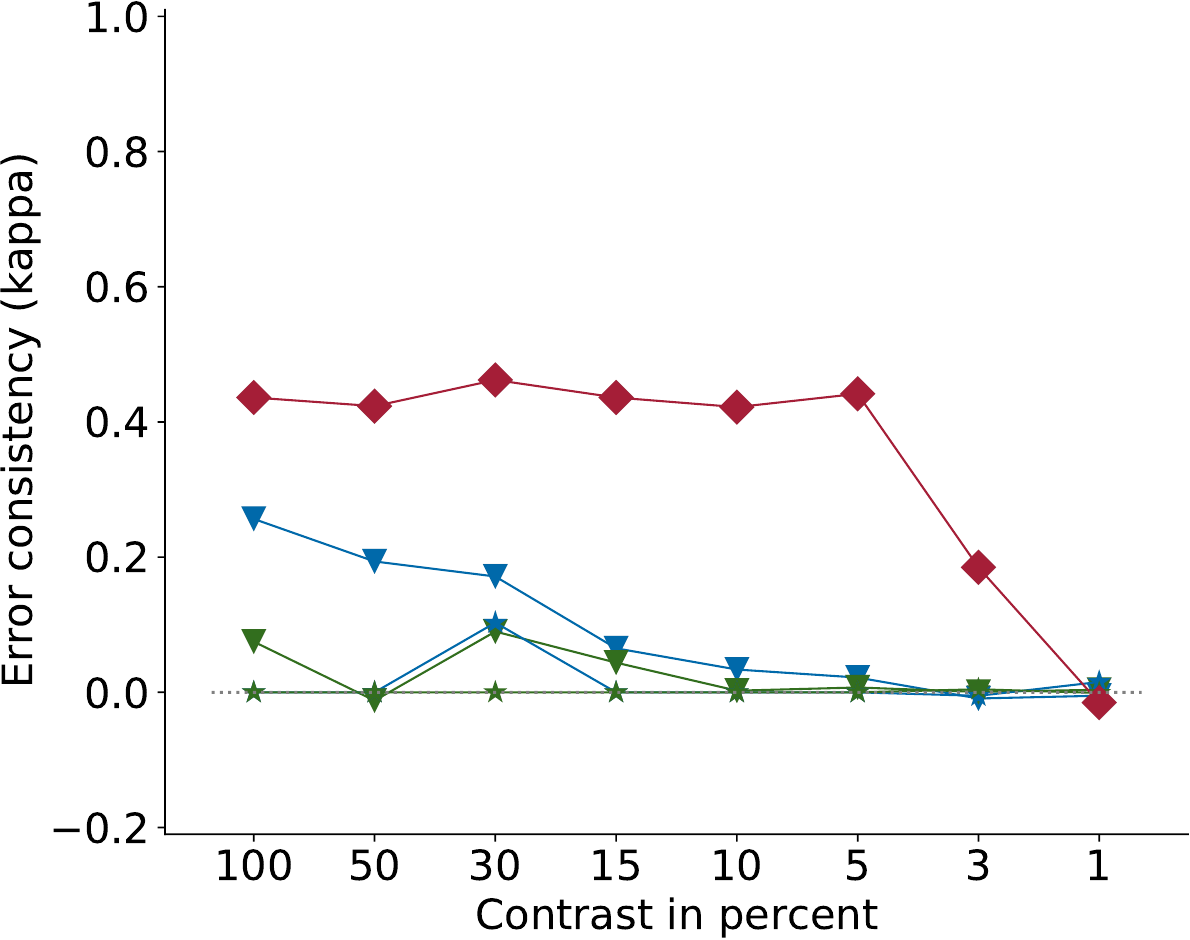}
			\vspace{\captionspace}
			\caption*{}
			\vspace{\captionspaceII}
		\end{subfigure}\hfill
		\begin{subfigure}{\figwidth}
			\centering
			\includegraphics[width=\linewidth]{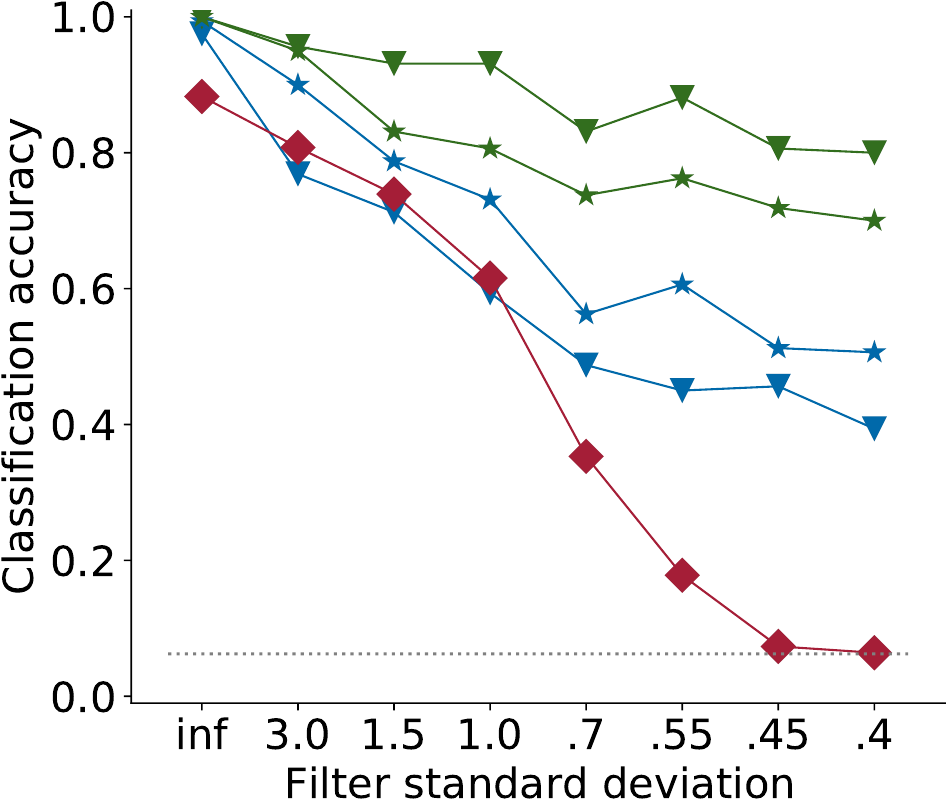}
			\vspace{\captionspace}
			\caption{High-pass}
			\vspace{\captionspaceII}
		\end{subfigure}\hfill
		\begin{subfigure}{\figwidth}
			\centering
			\includegraphics[width=\linewidth]{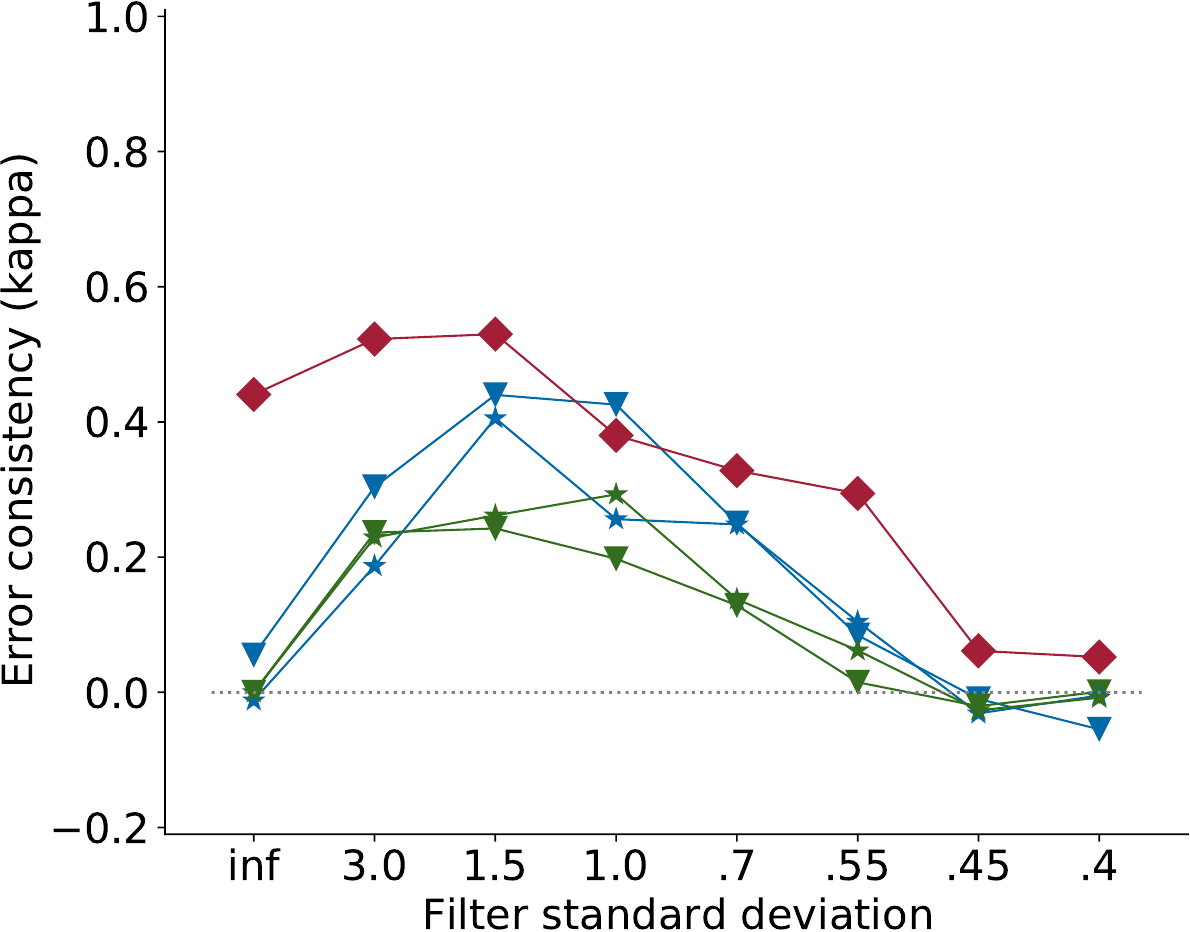}
			\vspace{\captionspace}
			\caption*{}
			\vspace{\captionspaceII}
		\end{subfigure}\hfill

		\begin{subfigure}{\figwidth}
			\centering
			\includegraphics[width=\linewidth]{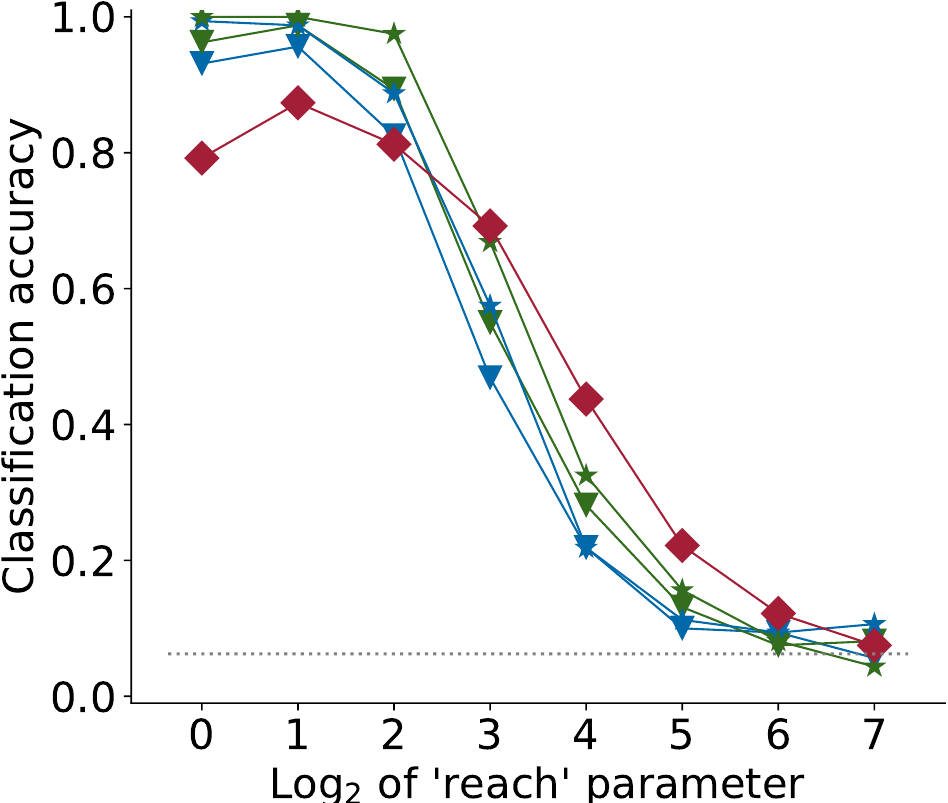}
			\vspace{\captionspace}
			\caption{Eidolon I}
			\vspace{\captionspaceII}
		\end{subfigure}\hfill
		\begin{subfigure}{\figwidth}
			\centering        \includegraphics[width=\linewidth]{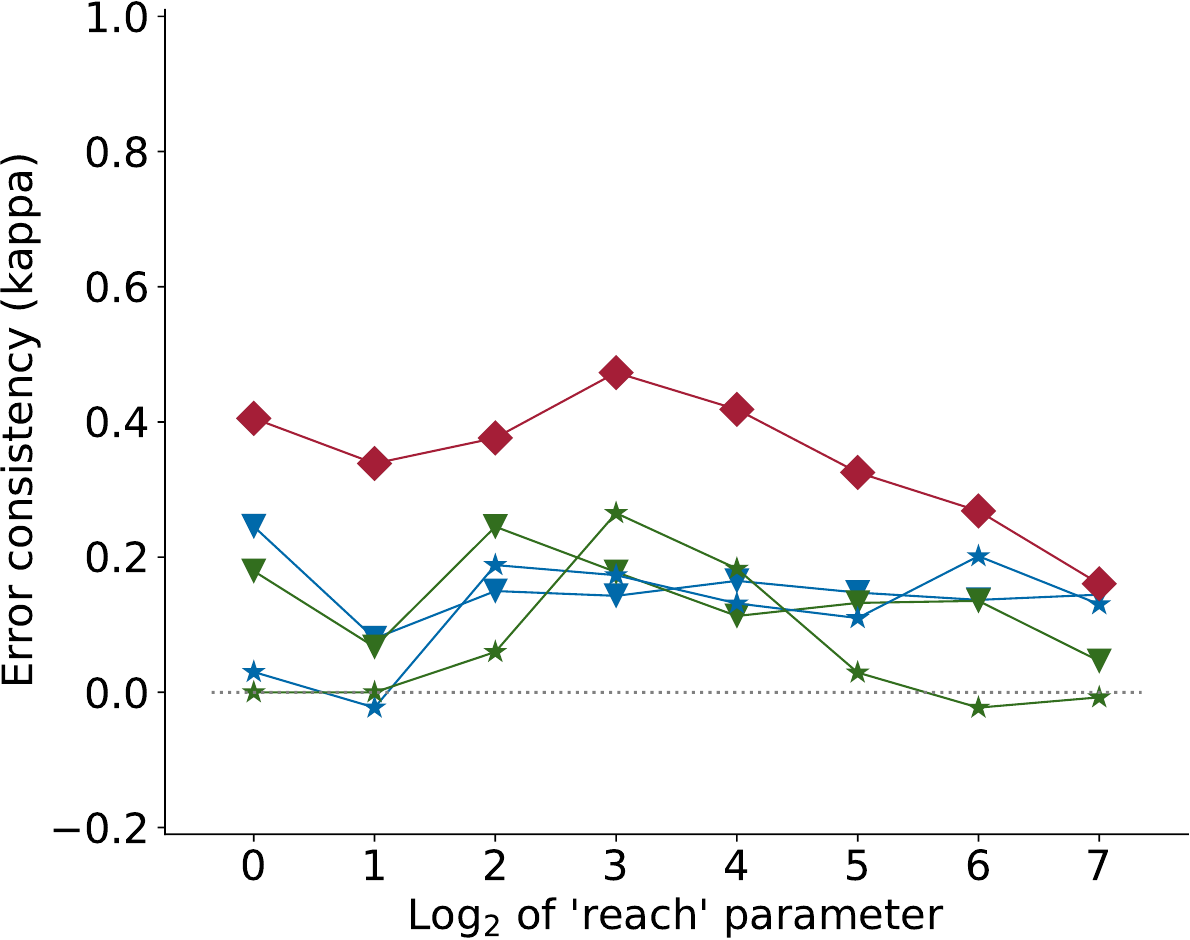}
			\vspace{\captionspace}
			\caption*{}
			\vspace{\captionspaceII}
		\end{subfigure}\hfill
		\begin{subfigure}{\figwidth}
			\centering
			\includegraphics[width=\linewidth]{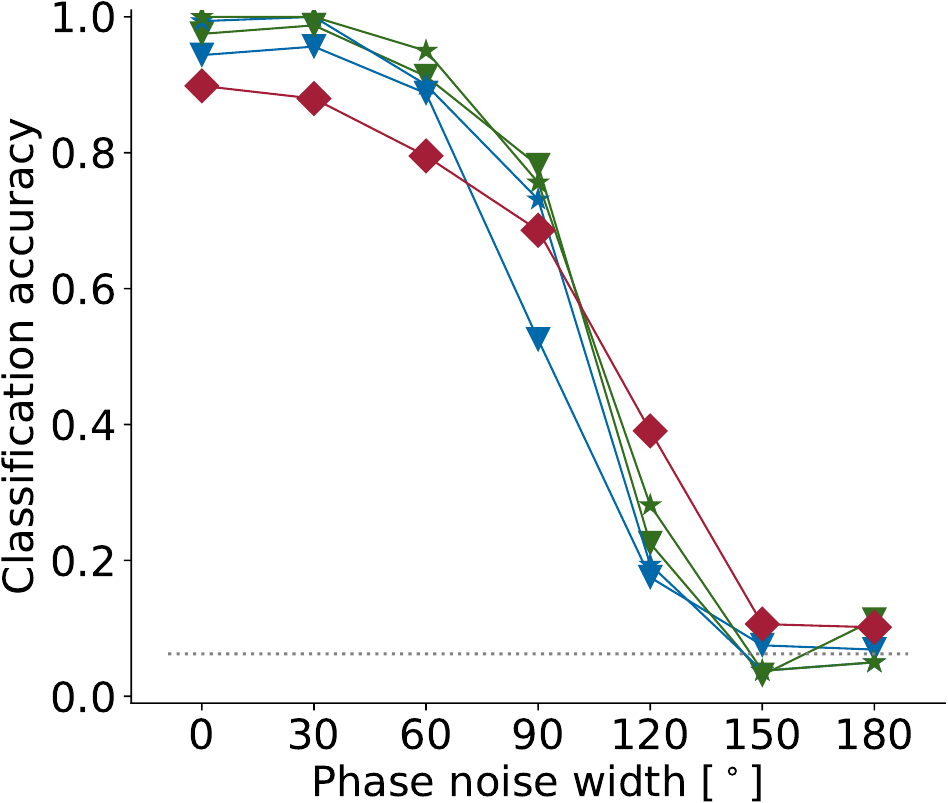}
			\vspace{\captionspace}
			\caption{Phase noise}
			\vspace{\captionspaceII}
		\end{subfigure}\hfill
		\begin{subfigure}{\figwidth}
			\centering
			\includegraphics[width=\linewidth]{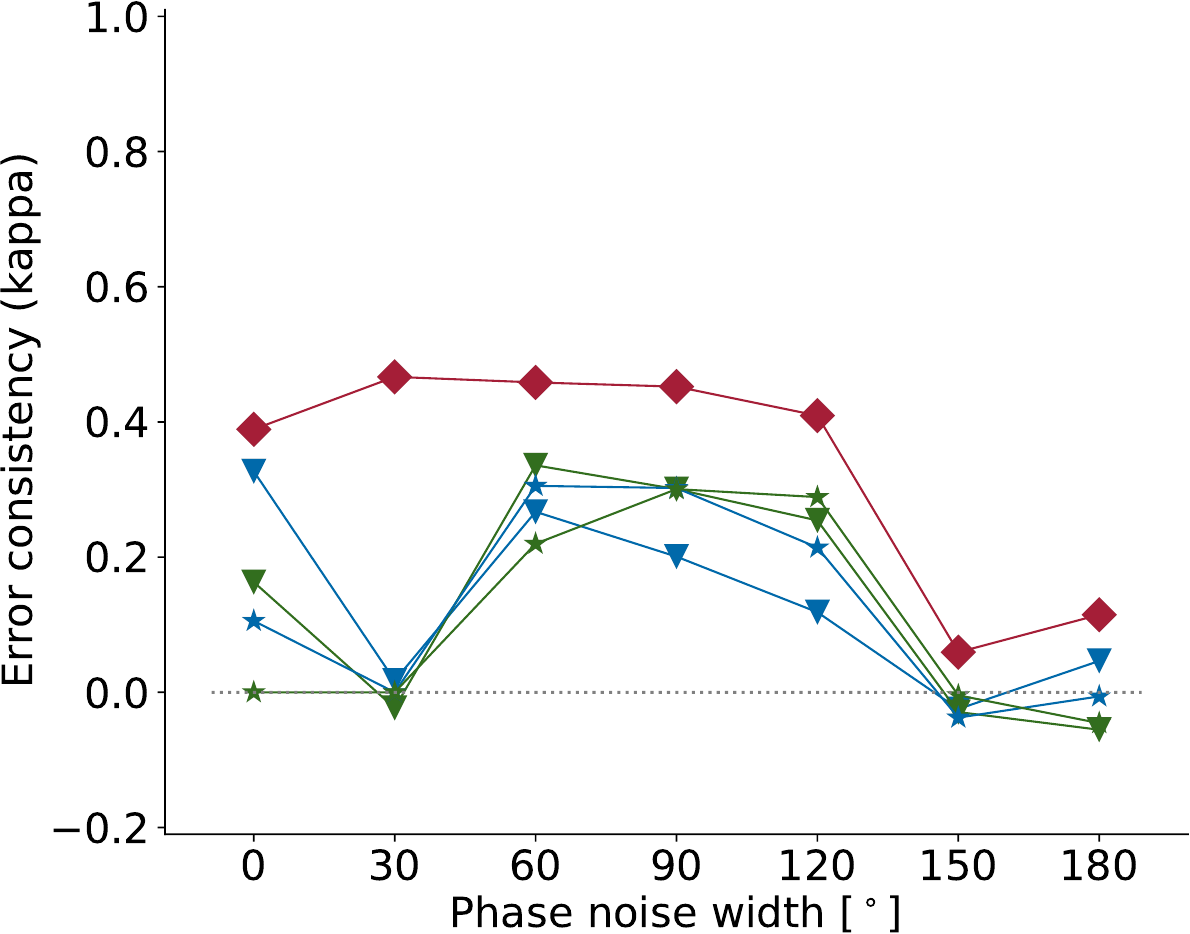}
			\vspace{\captionspace}
			\caption*{}
			\vspace{\captionspaceII}
		\end{subfigure}\hfill

	\begin{subfigure}{\figwidth}
		\centering
		\includegraphics[width=\linewidth]{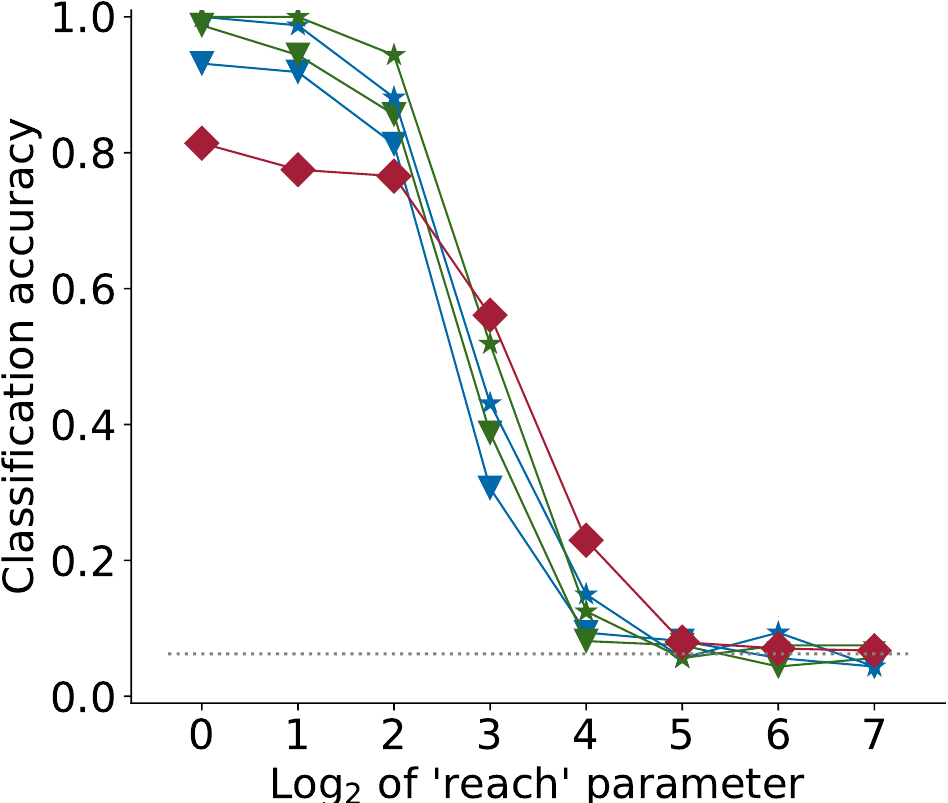}
		\vspace{\captionspace}
		\caption{Eidolon II}
		\vspace{\captionspaceII}
	\end{subfigure}\hfill
	\begin{subfigure}{\figwidth}
		\centering
		\includegraphics[width=\linewidth]{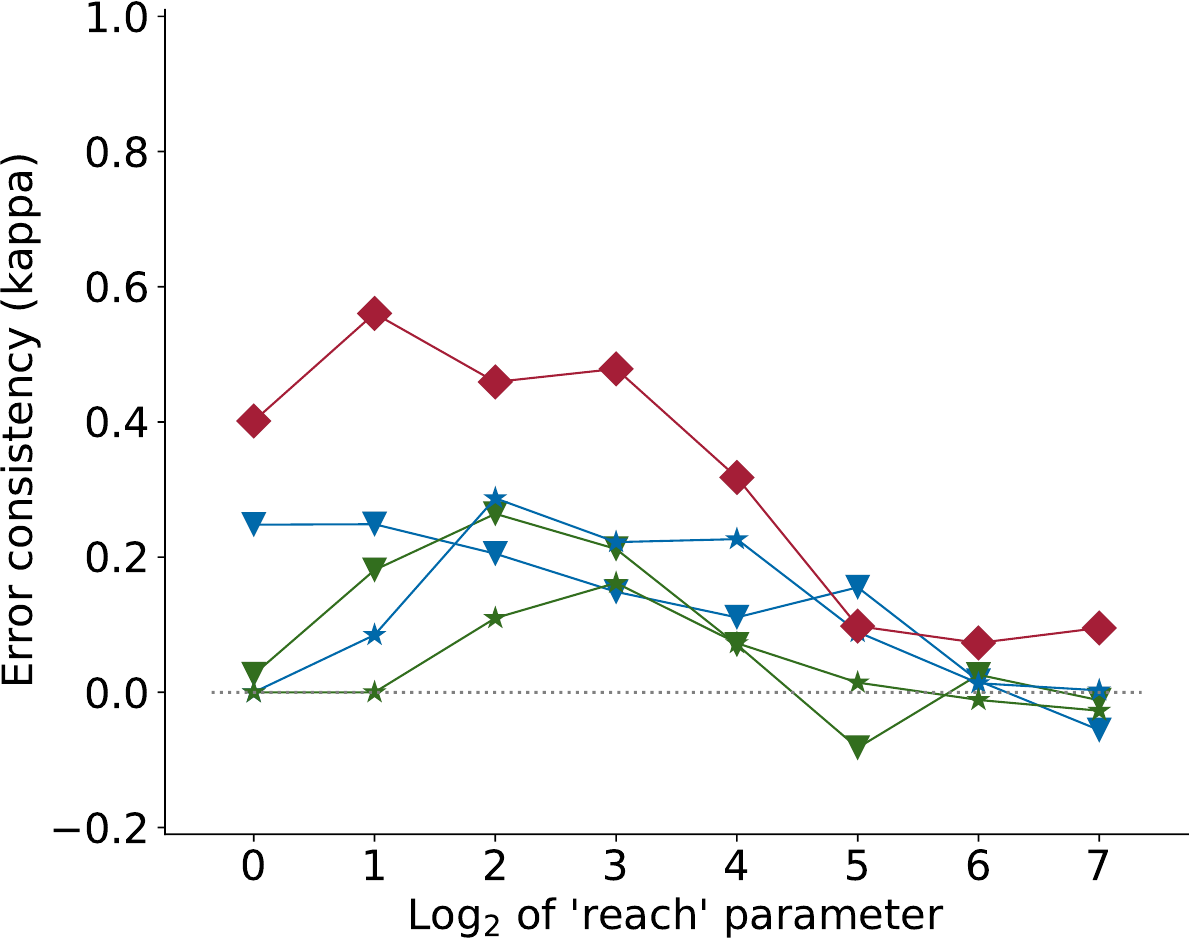}
		\vspace{\captionspace}
		\caption*{}
		\vspace{\captionspaceII}
	\end{subfigure}\hfill
	\begin{subfigure}{\figwidth}
		\centering
		\includegraphics[width=\linewidth]{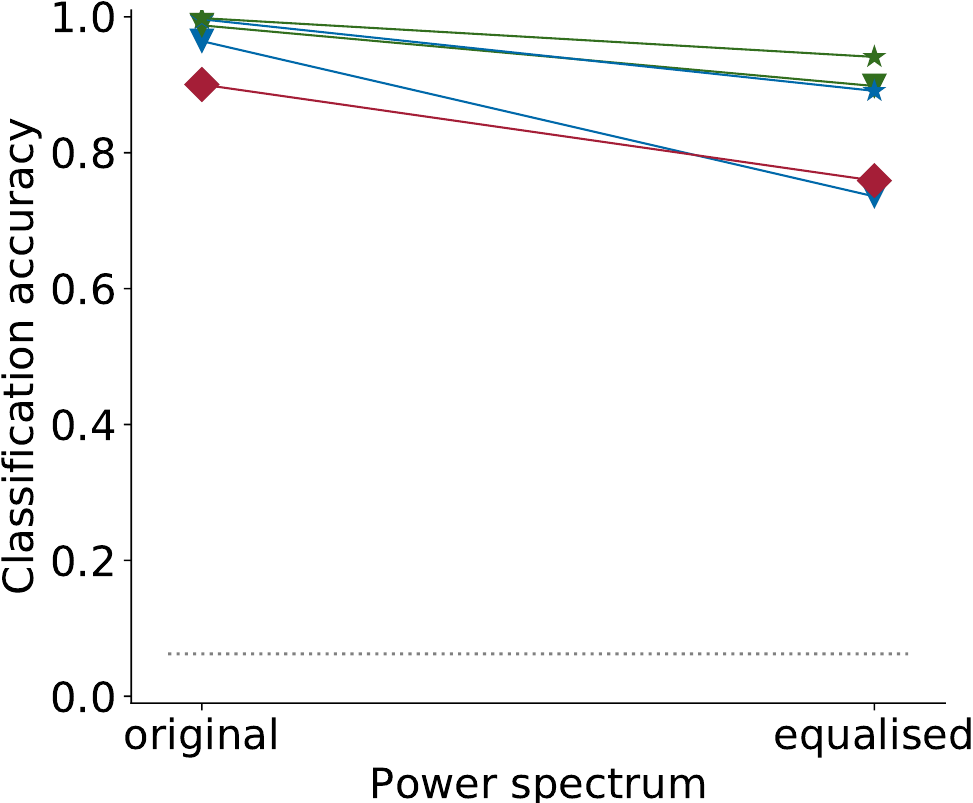}
		\vspace{\captionspace}
		\caption{Power equalisation}
		\vspace{\captionspaceII}
	\end{subfigure}\hfill
	\begin{subfigure}{\figwidth}
		\centering
		\includegraphics[width=\linewidth]{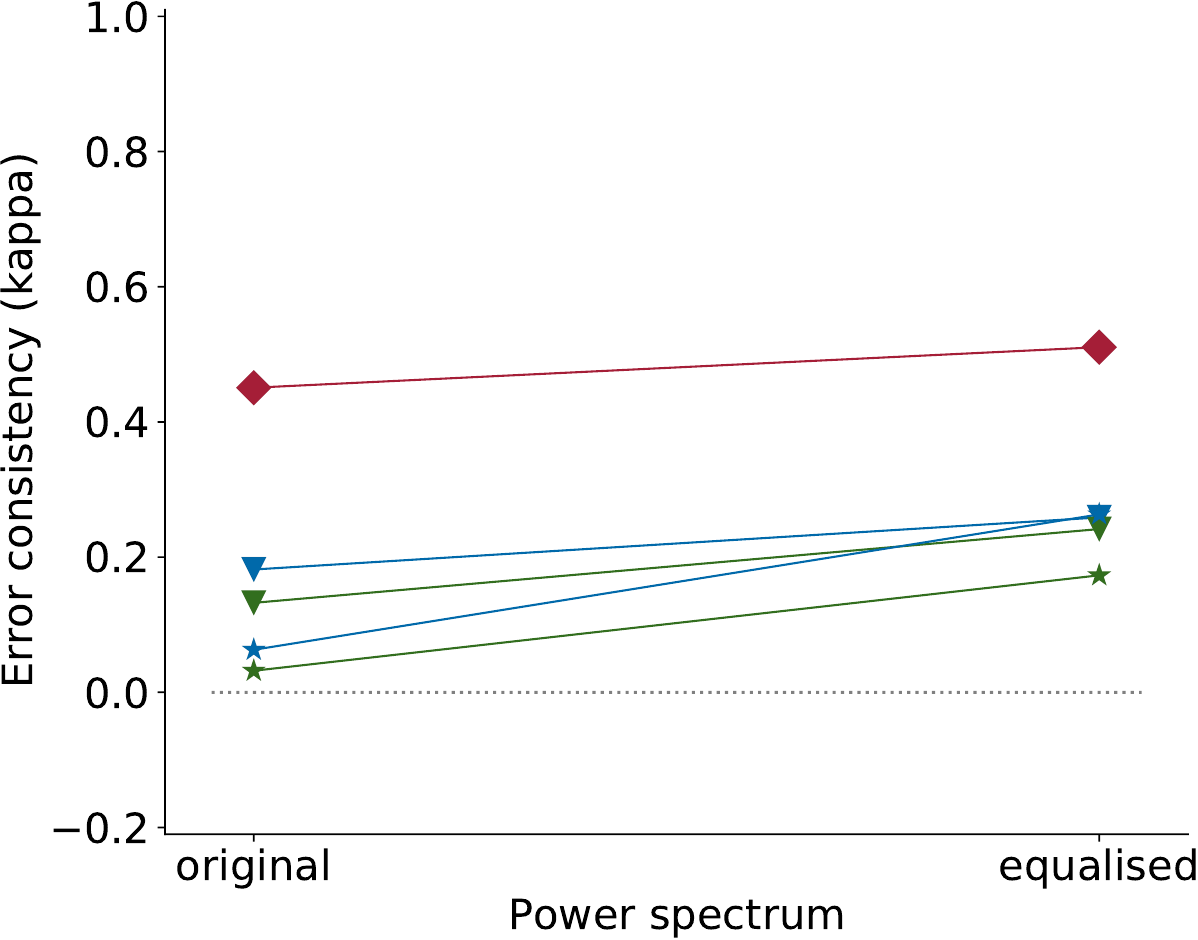}
		\vspace{\captionspace}
		\caption*{}
		\vspace{\captionspaceII}
	\end{subfigure}\hfill

	\begin{subfigure}{\figwidth}
		\centering
		\includegraphics[width=\linewidth]{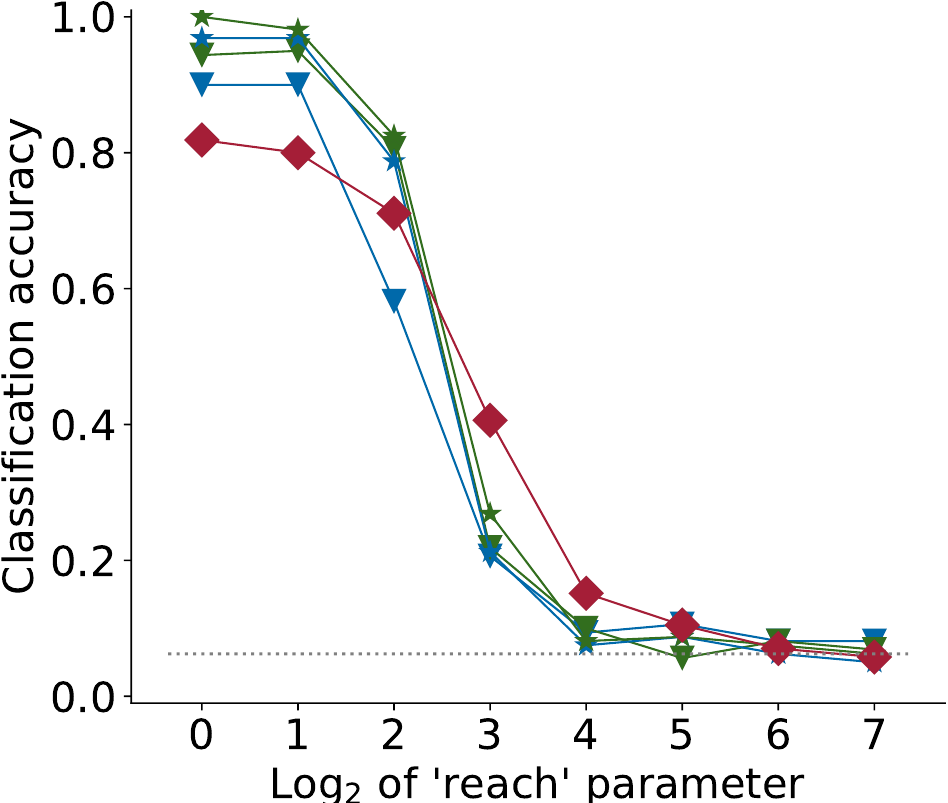}
		\vspace{\captionspace}
		\caption{Eidolon III}
	\end{subfigure}\hfill
	\begin{subfigure}{\figwidth}
		\centering        \includegraphics[width=\linewidth]{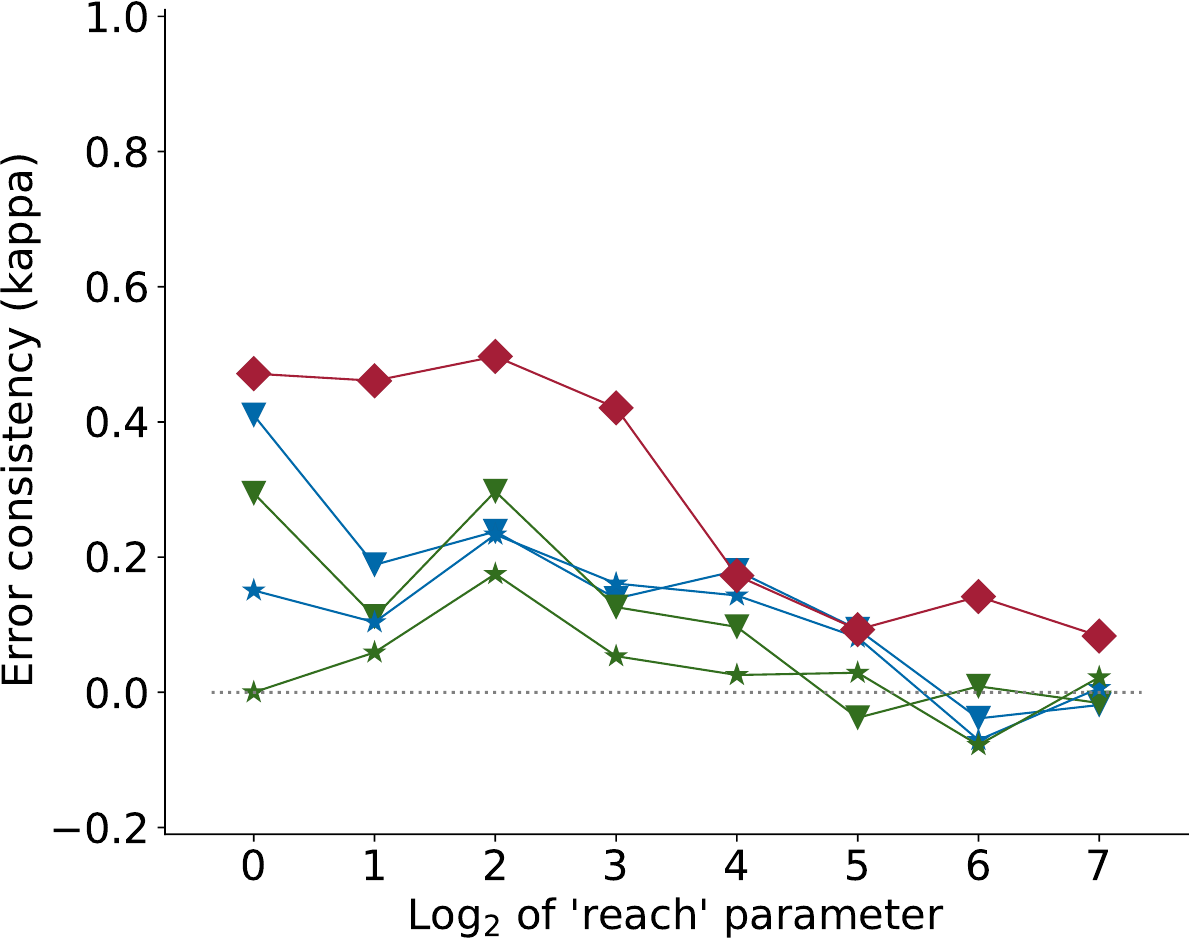}
		\vspace{\captionspace}
		\caption*{}
	\end{subfigure}\hfill
	\begin{subfigure}{\figwidth}
		\centering
		\includegraphics[width=\linewidth]{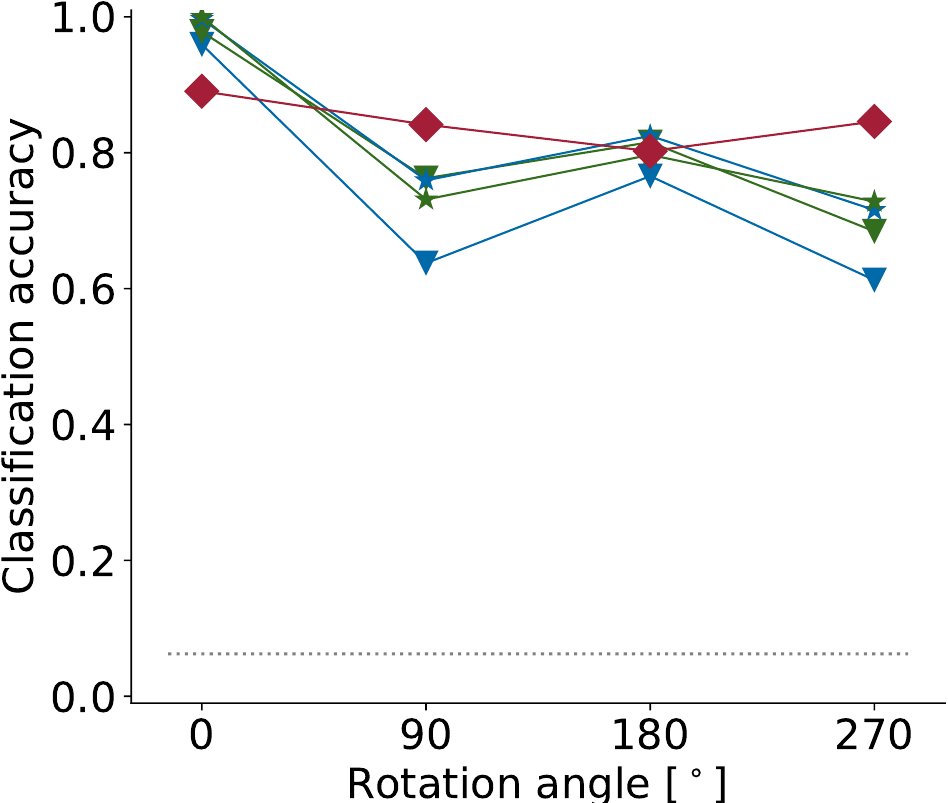}
		\vspace{\captionspace}
		\caption{Rotation}
	\end{subfigure}\hfill
	\begin{subfigure}{\figwidth}
		\centering
		\includegraphics[width=\linewidth]{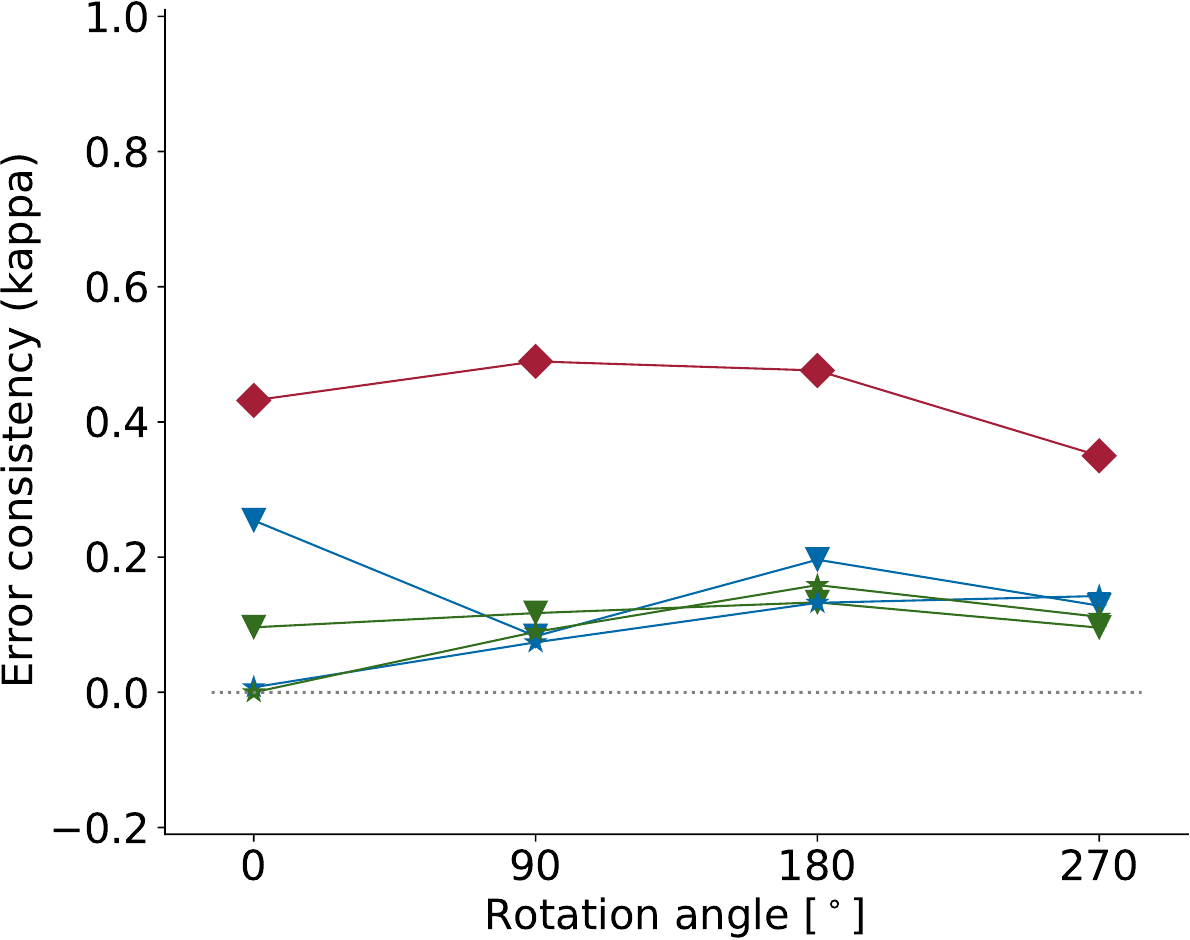}
		\vspace{\captionspace}
		\caption*{}
	\end{subfigure}\hfill
	\caption{Comparison of self-supervised SimCLR models with supervised, augmentation-matched baseline models. Note that for better visibility, the colours and symbols deviate from previous plots. Plotting symbols: triangles for self-supervised models, stars for supervised baselines. Two different model-baseline pairs are plotted; they differ in the model width: \textcolor{blue2}{blue} models have 1x ResNet width, \textcolor{green1}{green} models have 4x ResNet width \cite{chen2020simple}. For context, human observers are plotted as \textcolor{red}{red diamonds}. Baseline models kindly provided by \citet{hermann2020the}.}
	\label{fig:results_accuracy_entropy_simclr_baseline}
\end{figure}

Figure~\ref{fig:results_accuracy_entropy_simclr_baseline} compares the noise generalisation performance self-supervised SimCLR models against augmentation-matched baseline models. The results indicate that the superior performance of SimCLR in Figure~\ref{fig:results_accuracy} are largely a consequence of SimCLR's data augmentation scheme, rather than a property of the self-supervised contrastive loss.

\section{Benchmark scores}
\label{app:benchmark_scores}
Figure~\ref{fig:benchmark_barplots} in the main paper shows aggregated scores for the most robust model in terms of OOD accuracy (Figure~\ref{subfig:benchmark_a}), and for the most human-like models in terms of accuracy, observed and error consistency (Figures~\ref{subfig:benchmark_b}, \ref{subfig:benchmark_c}, \ref{subfig:benchmark_d}). Numerically, these metrics are represented in two tables, ranking the models according to out-of-distribution robustness (Table~\ref{tab:benchmark_table_accurate}) and human-like behaviour (Table~\ref{tab:benchmark_table_humanlike}). Since the latter is represented by three different metrics (each characterising a distinct aspect), the mean rank across those three metrics is used to obtain a final ordering. The following conditions and datasets influence benchmark scores: For the five nonparametric datasets, all datasets are taken into account. For the twelve parametric datasets, we also take all datasets into account (overall, all 17 datasets are weighted equally); however, we exclude certain conditions for principled reasons. First of all, the easiest condition is always excluded since it does not test out-of-distribution behaviour (e.g., for the contrast experiment, 100\% contrast is more of a baseline condition rather than a condition of interest). Furthermore, we exclude all conditions for which human average performance is strictly smaller than 0.2, since e.g.\ comparisons against human error patterns are futile if humans are randomly guessing since they cannot identify the stimuli anymore. For these reasons, the following conditions are not taken into account when computing the benchmark scores. Colour vs.\ greyscale experiment: condition ``colour''. True vs.\ false colour experiment: condition ``true colour''. Uniform noise experiment: conditions 0.0, 0.6, 0.9. Low-pass experiment: conditions 0, 15, 40. Contrast experiment: conditions 100, 3, 1. High-pass experiment: conditions inf, 0.55, 0.45, 0.4. Eidolon~I experiment: conditions 0, 6, 7. Phase noise experiment: conditions 0, 150, 180. Eidolon~II experiment: conditions 0, 5, 6, 7. Power-equalisation experiment: condition ``original power spectrum''. Eidolon~III experiment: conditions 0, 4, 5, 6, 7. Rotation experiment: condition 0.

\section{Regression model}
\label{app:regression_model}
In order to quantify the influence of known independent variables (architecture: transformers vs. ConvNets; data: small (ImageNet) vs. large (``more'' than standard ImageNet); objective: supervised vs. self-supervised) on known dependent variables (OOD accuracy and error consistency with humans), we performed a regression analysis using R version 3.6.3 (functions \texttt{lm} for fitting and \texttt{anova} for regression model comparison). We modelled the influence of those predictors on OOD accuracy, and on error consistency with human observers in two separate linear regression models (one per dependent variable). To this end, we used incremental model building, i.e.\ starting with one significant predictor and subsequently adding predictors if the reduction of degrees of freedom is justified by a significantly higher degree of explained variance (alpha level: .05). Both error consistency and accuracy, for our 52 models, followed (approximately) a normal distribution, as confirmed by density and Q-Q-plots. That being said, the fit was better for error consistency than for accuracy.

The final regression model for error consistency showed:

\begin{itemize}
    \item a significant main effect for transformers over CNNs (p = 0.01936 *),
    \item a significant main effect for large datasets over small datasets (p = 3.39e-05 ***),
    \item a significant interaction between dataset size and objective function (p = 0.00625 **),
    \item no significant main effect of objective function (p = 0.10062, n.s.)
\end{itemize}
    
Residual standard error: 0.02156 on 47 degrees of freedom\\
Multiple R-squared: 0.5045, Adjusted R-squared:  0.4623\\ 
F-statistic: 11.96 on 4 and 47 DF, p-value: 8.765e-07\\
Significance codes:  0 ‘***’ 0.001 ‘**’ 0.01 ‘*’ 0.05\\

The final regression model for OOD accuracy showed:
\begin{itemize}
    \item a significant main effect for large datasets over small datasets (p = 4.65e-09 ***),
    \item a significant interaction between dataset size and architecture type (transformer vs. CNNs; p = 0.0174 *),
    \item no significant main effect for transformers vs. CNNs (p = 0.8553, n.s.)
\end{itemize}

Residual standard error: 0.0593 on 48 degrees of freedom\\
Multiple R-squared: 0.5848, Adjusted R-squared:  0.5588\\ 
F-statistic: 22.53 on 3 and 48 DF,  p-value: 3.007e-09\\
Significance codes:  0 ‘***’ 0.001 ‘**’ 0.01 ‘*’ 0.05\\

Limitations: a linear regression model can only capture linear effects; furthermore, diagnostic plots showed a better fit for the error consistency model (where residuals roughly followed the expected distribution as confirmed by a Q-Q-plot) than for the OOD accuracy model (where residuals were not perfectly normal distributed).

\section{Mapping behavioural decisions}
\label{app:mapping_decisions}
Comparing model and human classification decisions comes with a challenge: we simply cannot ask human observers to classify objects into 1,000 classes (as for standard ImageNet models). Even if this were feasible in terms of experimental time constraints, most humans don't routinely know the names of a hundred different dog breeds. What they do know, however, is how to tell dogs apart from cats and from airplanes, chairs and boats. Those are so-called ``basic'' or ``entry-level'' categories \citep{Rosch1999}. In line with previous work \citep{geirhos2018generalisation, geirhos2019imagenettrained, geirhos2020beyond}, we therefore used a set of 16 basic categories in our experiments. For ImageNet-trained models, to obtain a choice from the same 16 categories, the 1,000 class decision vector was mapped to those 16 classes using the WordNet hierarchy \citep{miller1995wordnet}. Those 16 categories were chosen to reflect a large chunk of ImageNet (227 classes, i.e.\ roughly a quarter of all ImageNet categories is represented by those 16 basic categories). In order to obtain classification decisions from ImageNet-trained models for those 16 categories, at least two choices are conceivable: re-training the final classification layer or using a principled mapping. Since any training involves making a number of choices (hyperparameters, optimizer, dataset, ...) that may potentially influence and in the worst case even bias the results (e.g. for ShuffleNet, more than half of the model’s parameters are contained in the final classification layer!), we decided against training and for a principled mapping by calculating the probability of a coarse class as the average of the probabilities of the corresponding fine-grained classes. Why is this mapping principled? As derived by \citep{geirhos2018generalisation} (pages 22 and 23 in the Appendix of the arXiv version, \url{https://arxiv.org/pdf/1808.08750.pdf}), this is the optimal way to map (i.e. aggregate) probabilities from many fine-grained classes to a few coarse classes. Essentially, the aggregation can be derived by calculating the posterior distribution of a discriminatively trained CNN under a new prior chosen at test time (here: 1/16
over coarse classes).

\begin{figure}
	\begin{subfigure}{\figwidth}
			\centering
			\textbf{Accuracy}\\
			\includegraphics[width=\linewidth]{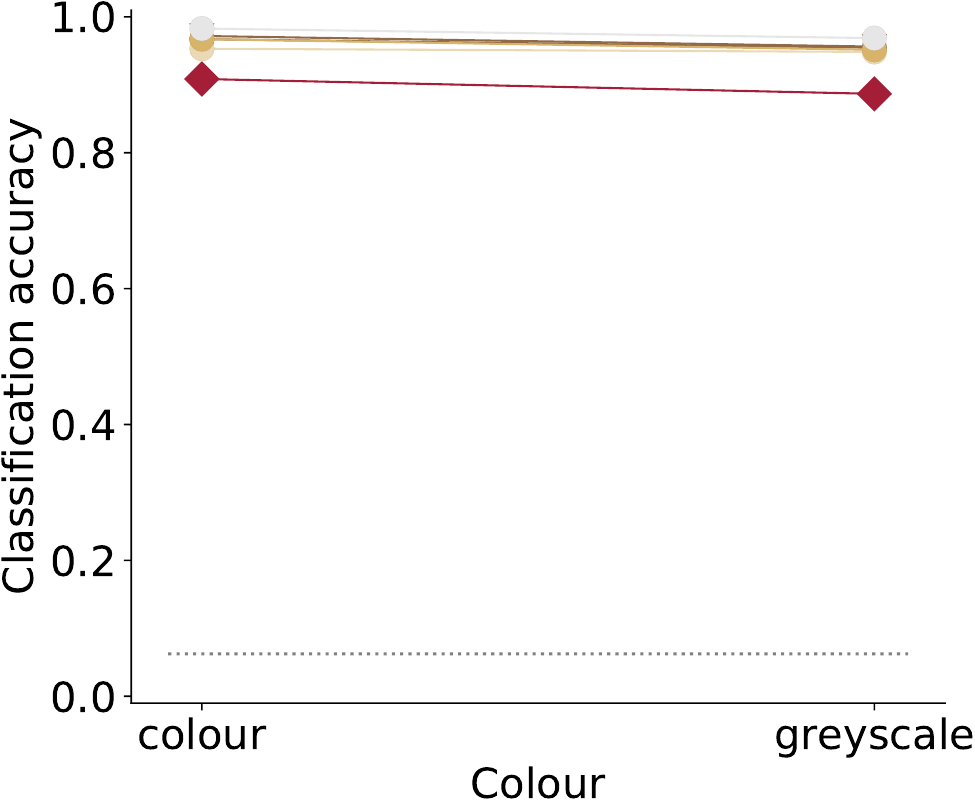}
			\vspace{\captionspace}
			\caption{Colour vs. greyscale}
			\vspace{\captionspaceII}
		\end{subfigure}\hfill
		\begin{subfigure}{\figwidth}
			\centering
			\textbf{Error consistency}\\
			\includegraphics[width=\linewidth]{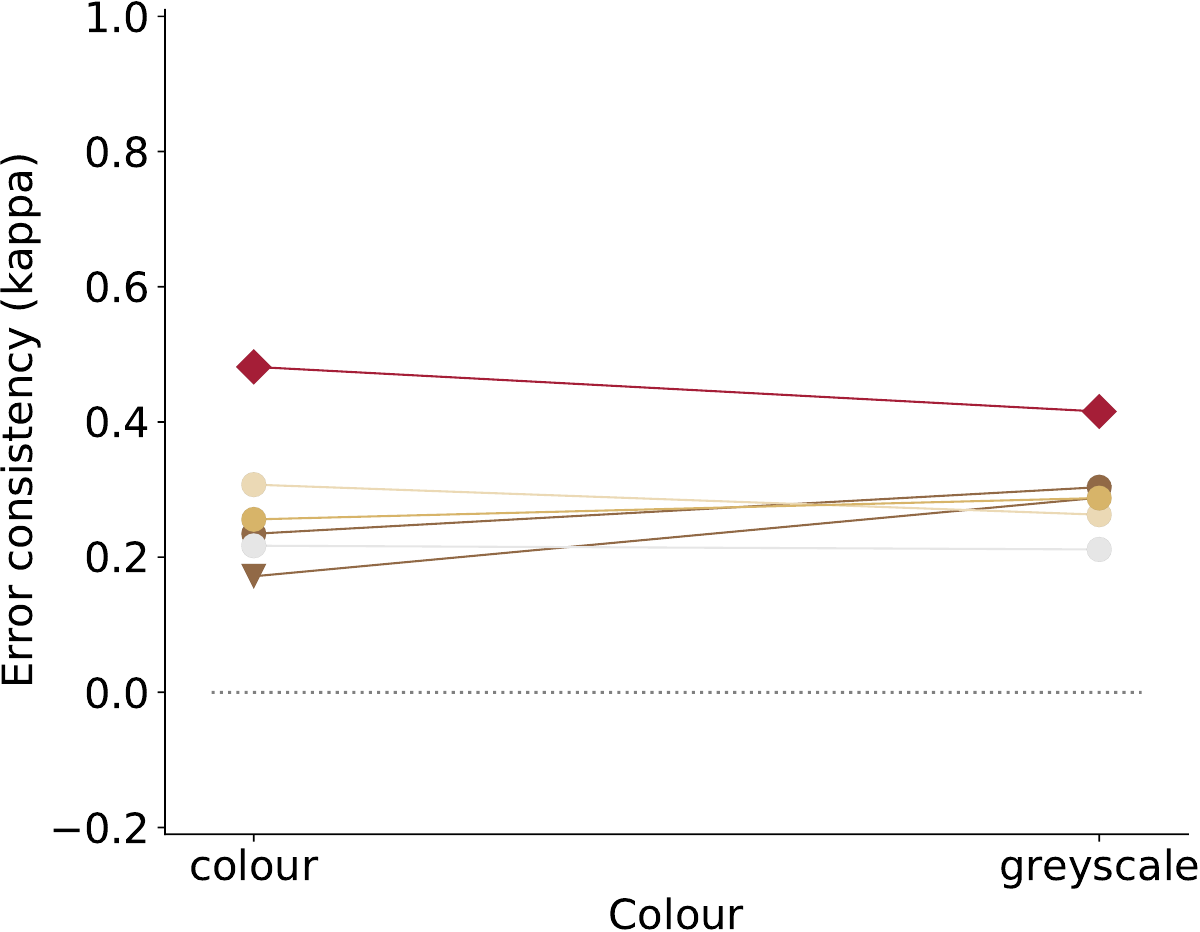}
			\vspace{\captionspace}
			\caption*{}
			\vspace{\captionspaceII}
		\end{subfigure}\hfill
		\begin{subfigure}{\figwidth}
			\centering
			\textbf{Accuracy}\\
			\includegraphics[width=\linewidth]{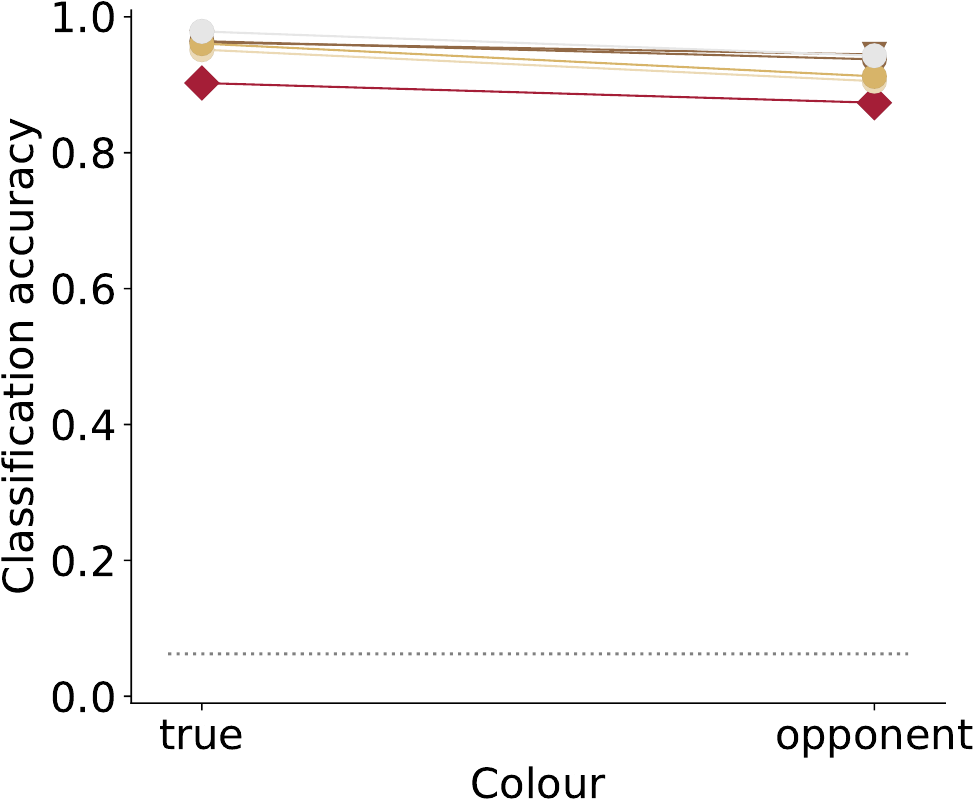}
			\vspace{\captionspace}
			\caption{True vs. false colour}
			\vspace{\captionspaceII}
		\end{subfigure}\hfill
		\begin{subfigure}{\figwidth}
			\centering
			\textbf{Error consistency}\\
			\includegraphics[width=\linewidth]{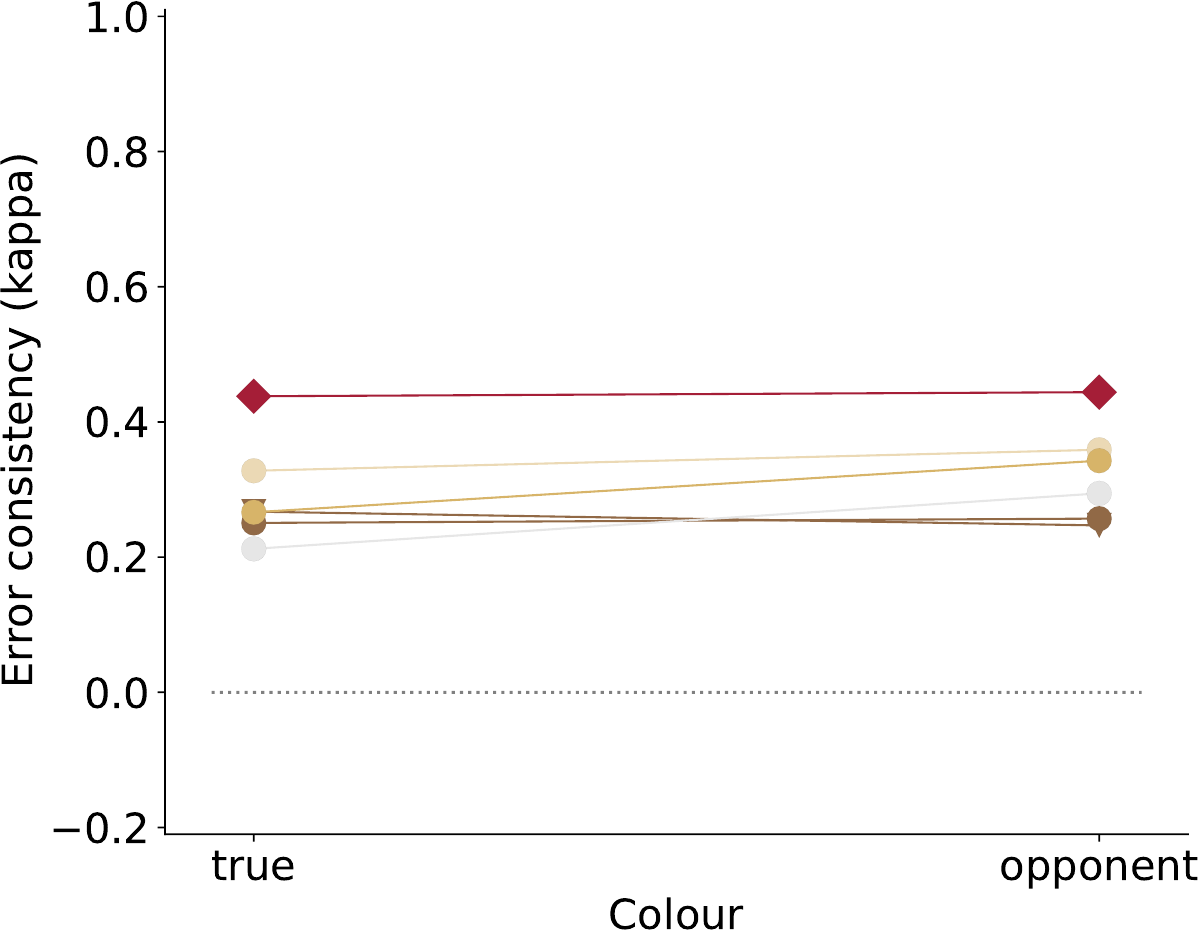}
			\vspace{\captionspace}
			\caption*{}
			\vspace{\captionspaceII}
		\end{subfigure}\hfill

		\begin{subfigure}{\figwidth}
			\centering
			\includegraphics[width=\linewidth]{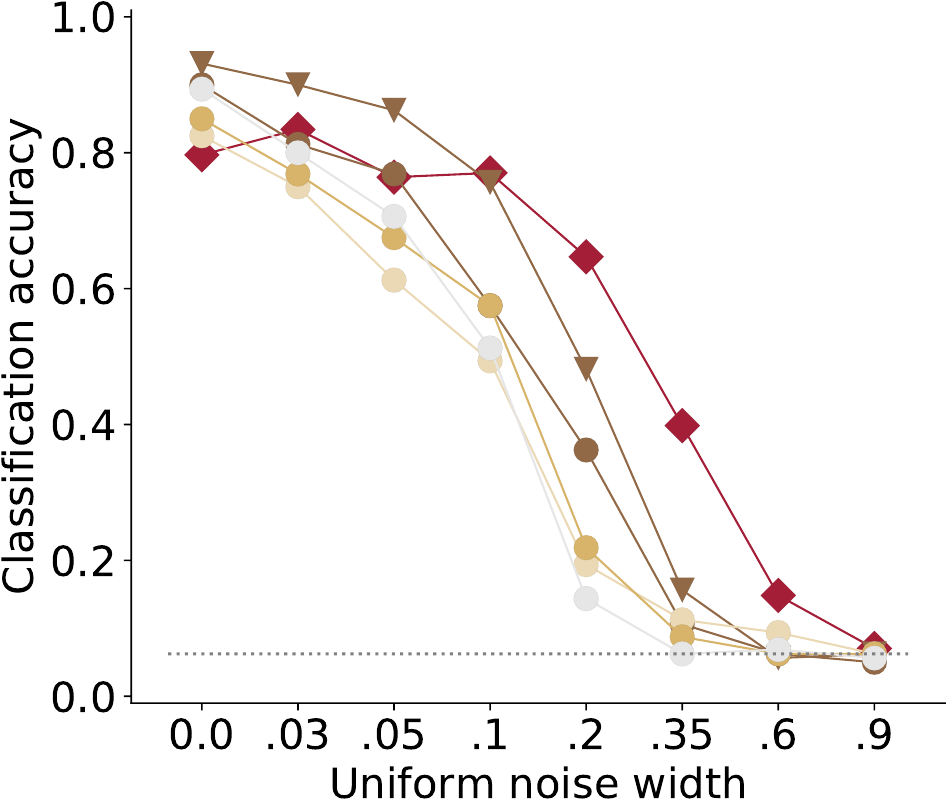}
			\vspace{\captionspace}
			\caption{Uniform noise}
			\vspace{\captionspaceII}
		\end{subfigure}\hfill
		\begin{subfigure}{\figwidth}
			\centering
			\includegraphics[width=\linewidth]{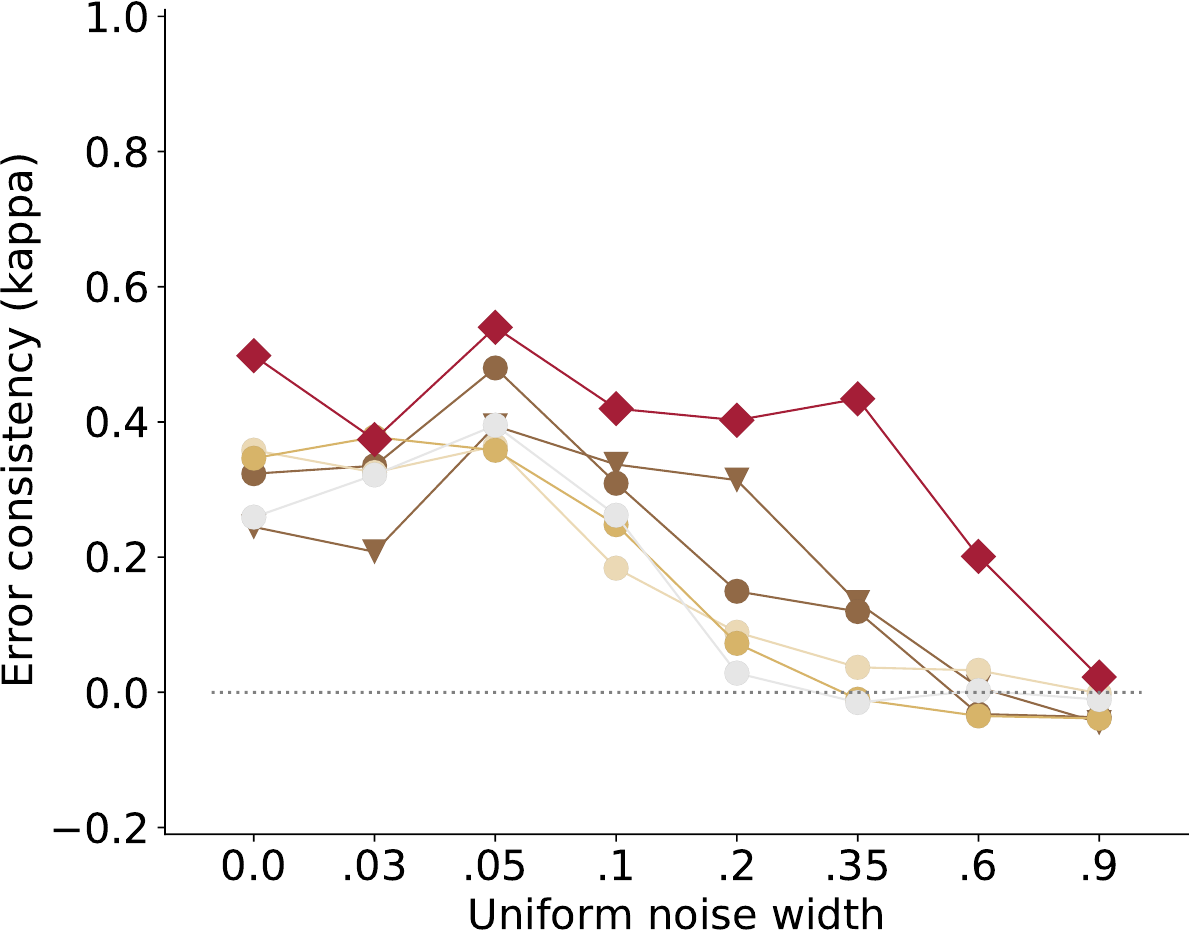}
			\vspace{\captionspace}
			\caption*{}
			\vspace{\captionspaceII}
		\end{subfigure}\hfill
		\begin{subfigure}{\figwidth}
			\centering
			\includegraphics[width=\linewidth]{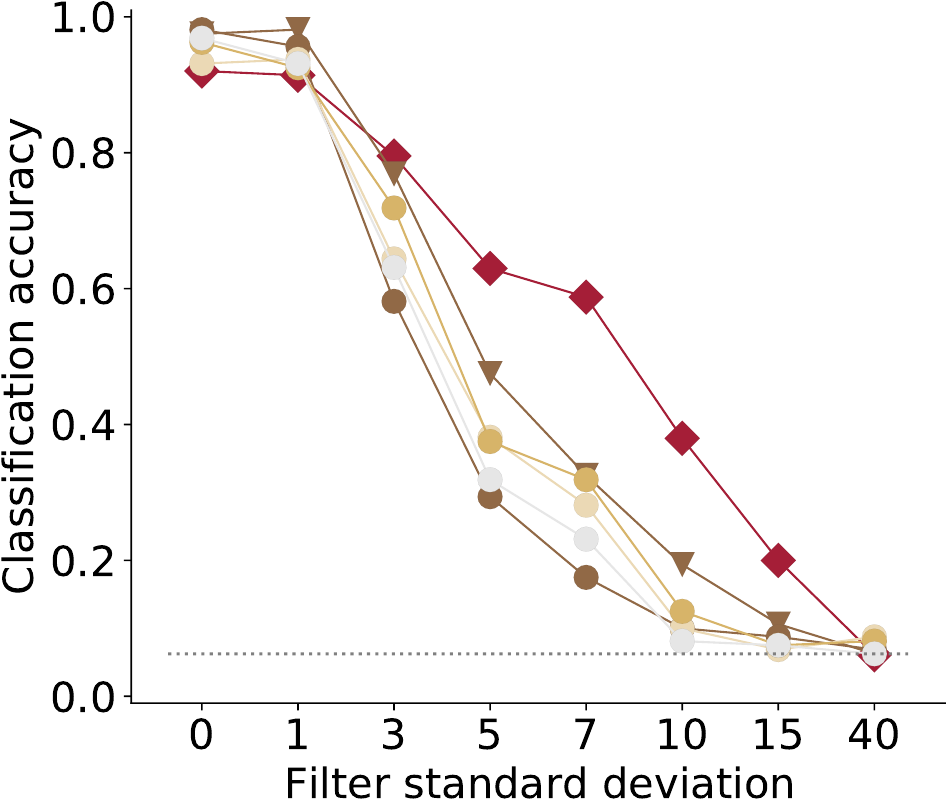}
			\vspace{\captionspace}
			\caption{Low-pass}
			\vspace{\captionspaceII}
		\end{subfigure}\hfill
		\begin{subfigure}{\figwidth}
			\centering
			\includegraphics[width=\linewidth]{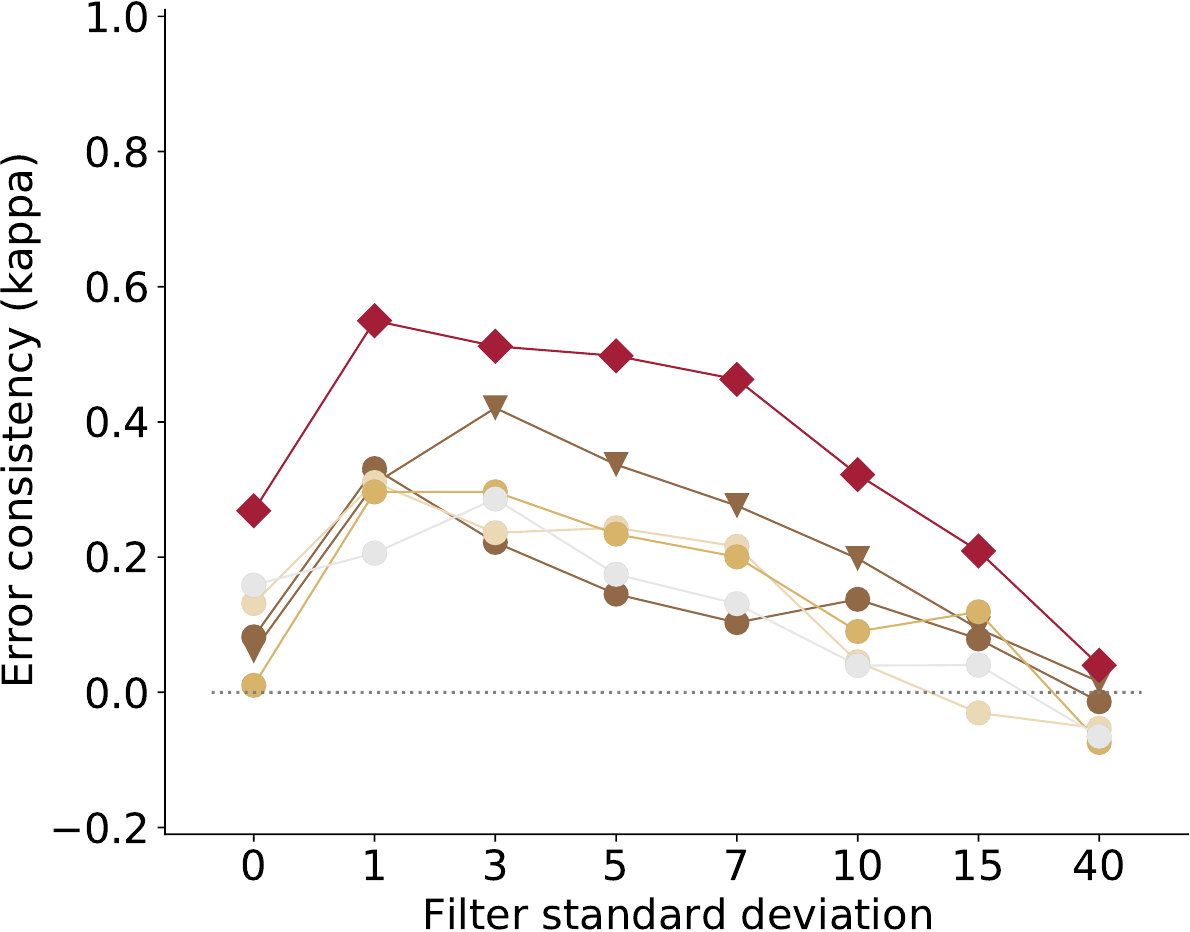}
			\vspace{\captionspace}
			\caption*{}
			\vspace{\captionspaceII}
		\end{subfigure}\hfill

		\begin{subfigure}{\figwidth}
			\centering
			\includegraphics[width=\linewidth]{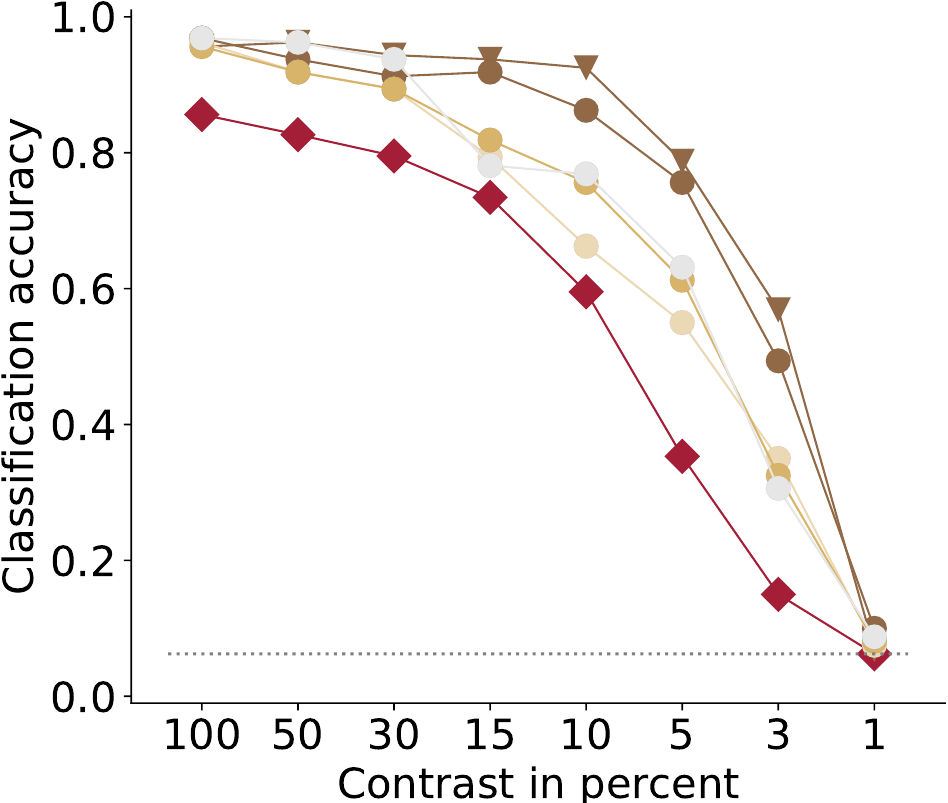}
			\vspace{\captionspace}
			\caption{Contrast}
			\vspace{\captionspaceII}
		\end{subfigure}\hfill
		\begin{subfigure}{\figwidth}
			\centering
			\includegraphics[width=\linewidth]{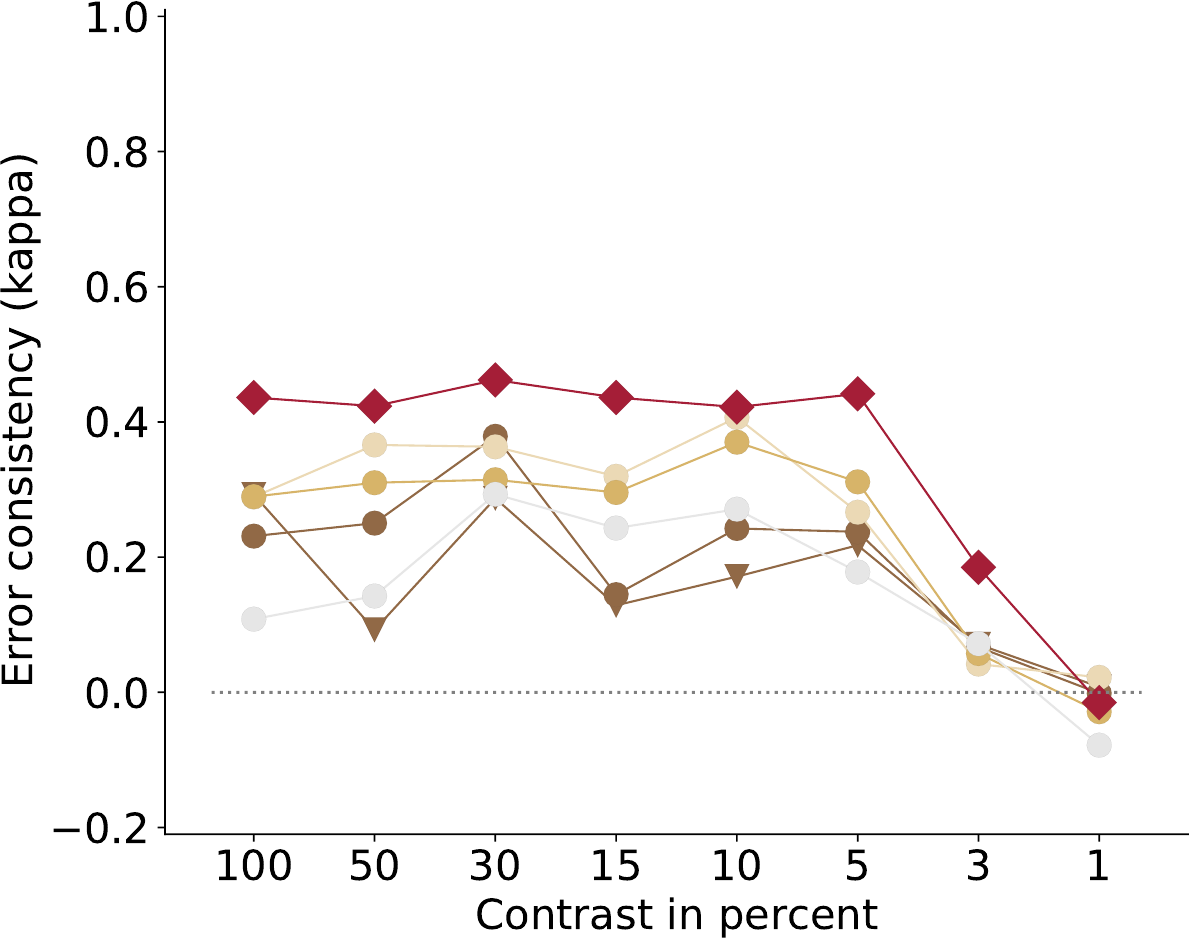}
			\vspace{\captionspace}
			\caption*{}
			\vspace{\captionspaceII}
		\end{subfigure}\hfill
		\begin{subfigure}{\figwidth}
			\centering
			\includegraphics[width=\linewidth]{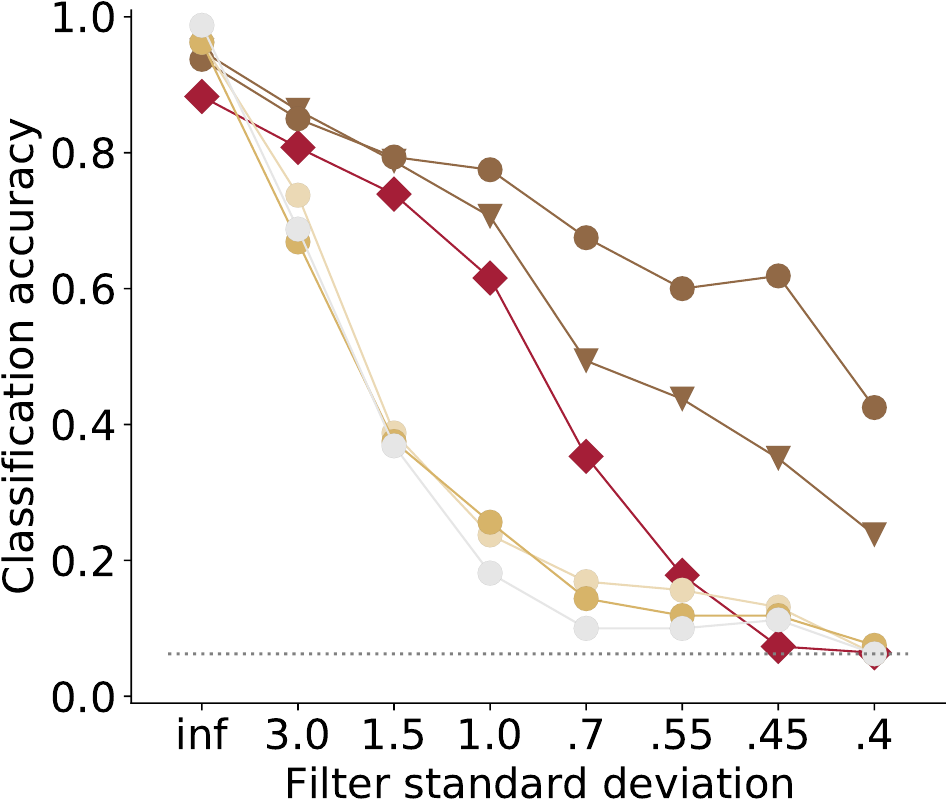}
			\vspace{\captionspace}
			\caption{High-pass}
			\vspace{\captionspaceII}
		\end{subfigure}\hfill
		\begin{subfigure}{\figwidth}
			\centering
			\includegraphics[width=\linewidth]{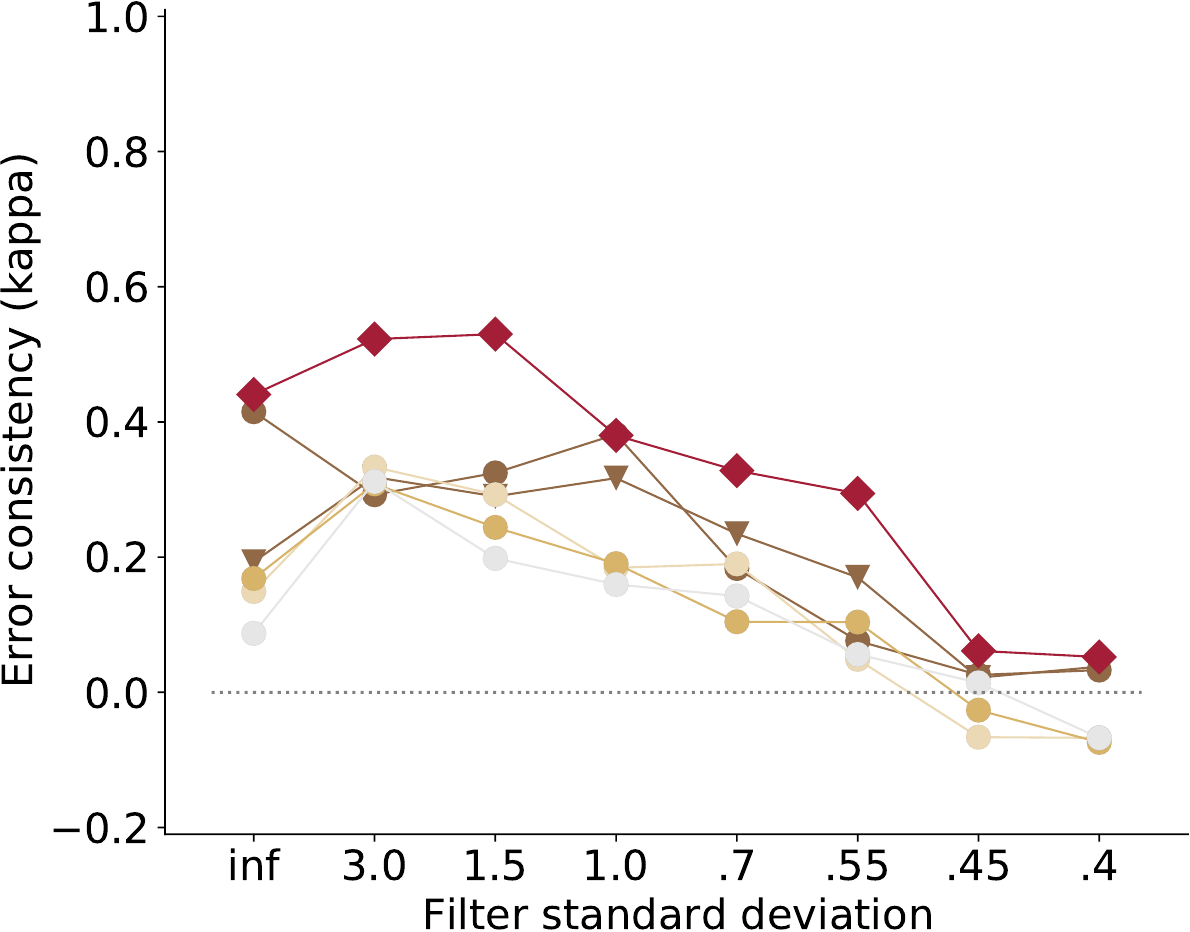}
			\vspace{\captionspace}
			\caption*{}
			\vspace{\captionspaceII}
		\end{subfigure}\hfill

		\begin{subfigure}{\figwidth}
			\centering
			\includegraphics[width=\linewidth]{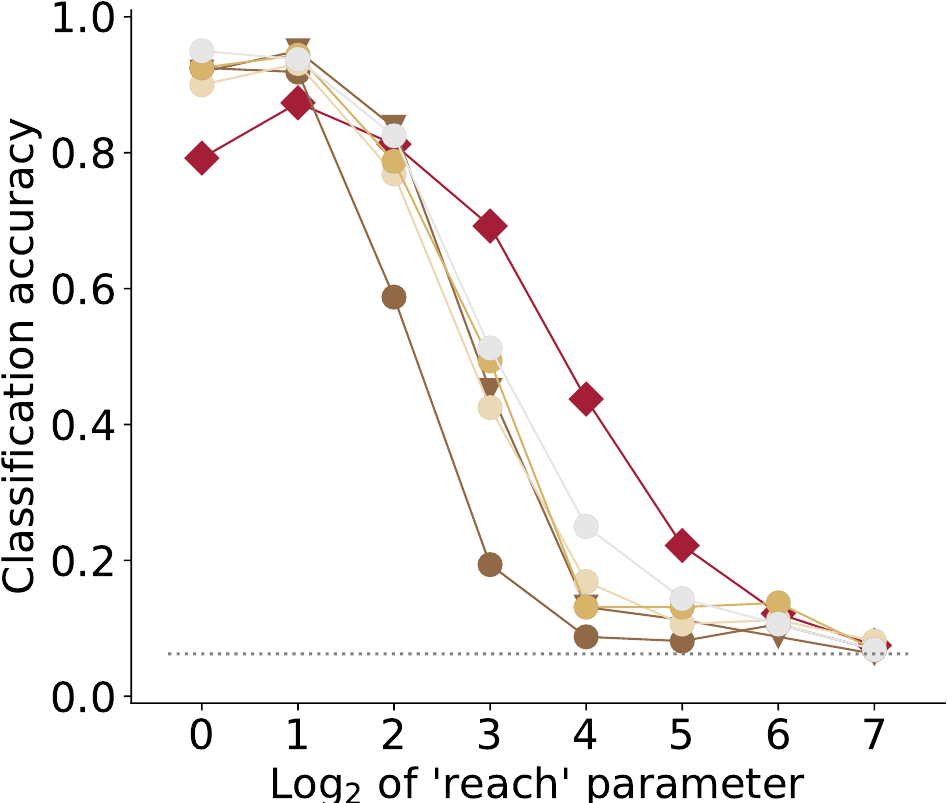}
			\vspace{\captionspace}
			\caption{Eidolon I}
			\vspace{\captionspaceII}
		\end{subfigure}\hfill
		\begin{subfigure}{\figwidth}
			\centering        \includegraphics[width=\linewidth]{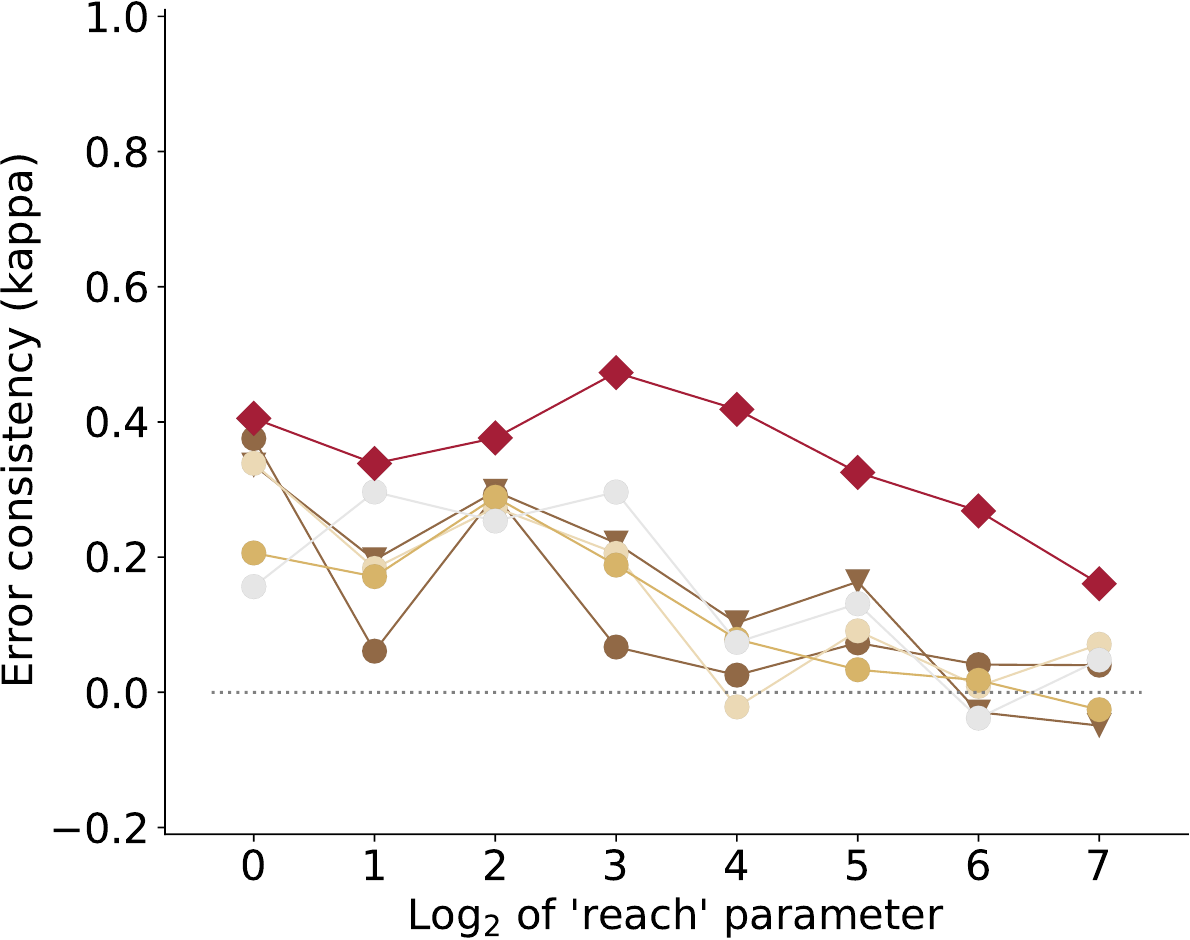}
			\vspace{\captionspace}
			\caption*{}
			\vspace{\captionspaceII}
		\end{subfigure}\hfill
		\begin{subfigure}{\figwidth}
			\centering
			\includegraphics[width=\linewidth]{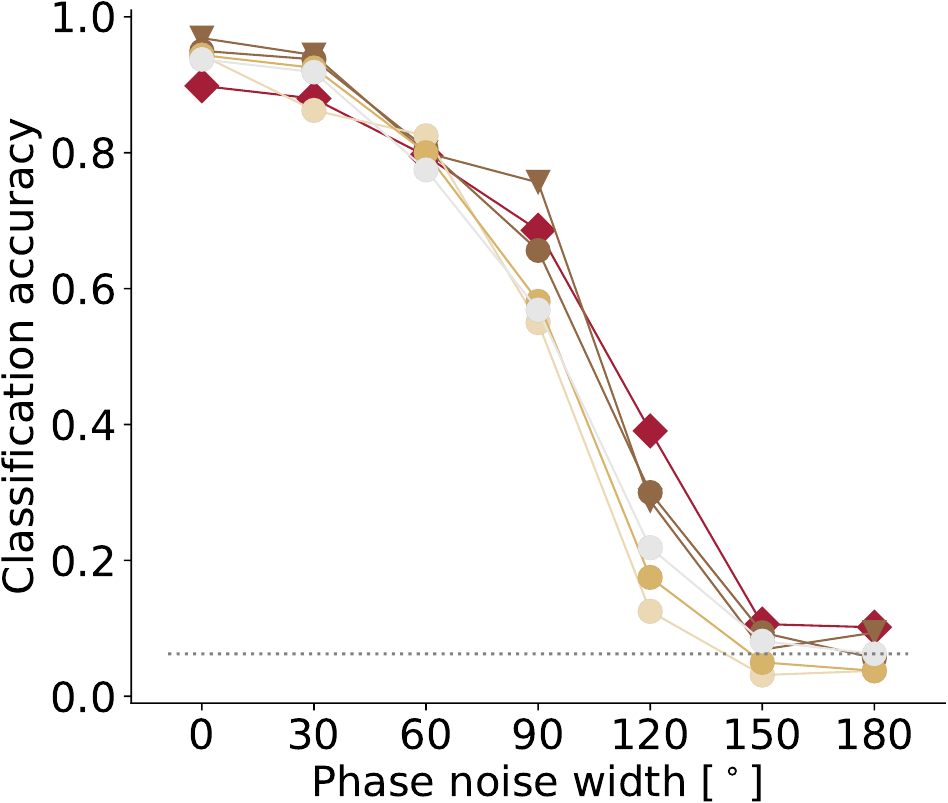}
			\vspace{\captionspace}
			\caption{Phase noise}
			\vspace{\captionspaceII}
		\end{subfigure}\hfill
		\begin{subfigure}{\figwidth}
			\centering
			\includegraphics[width=\linewidth]{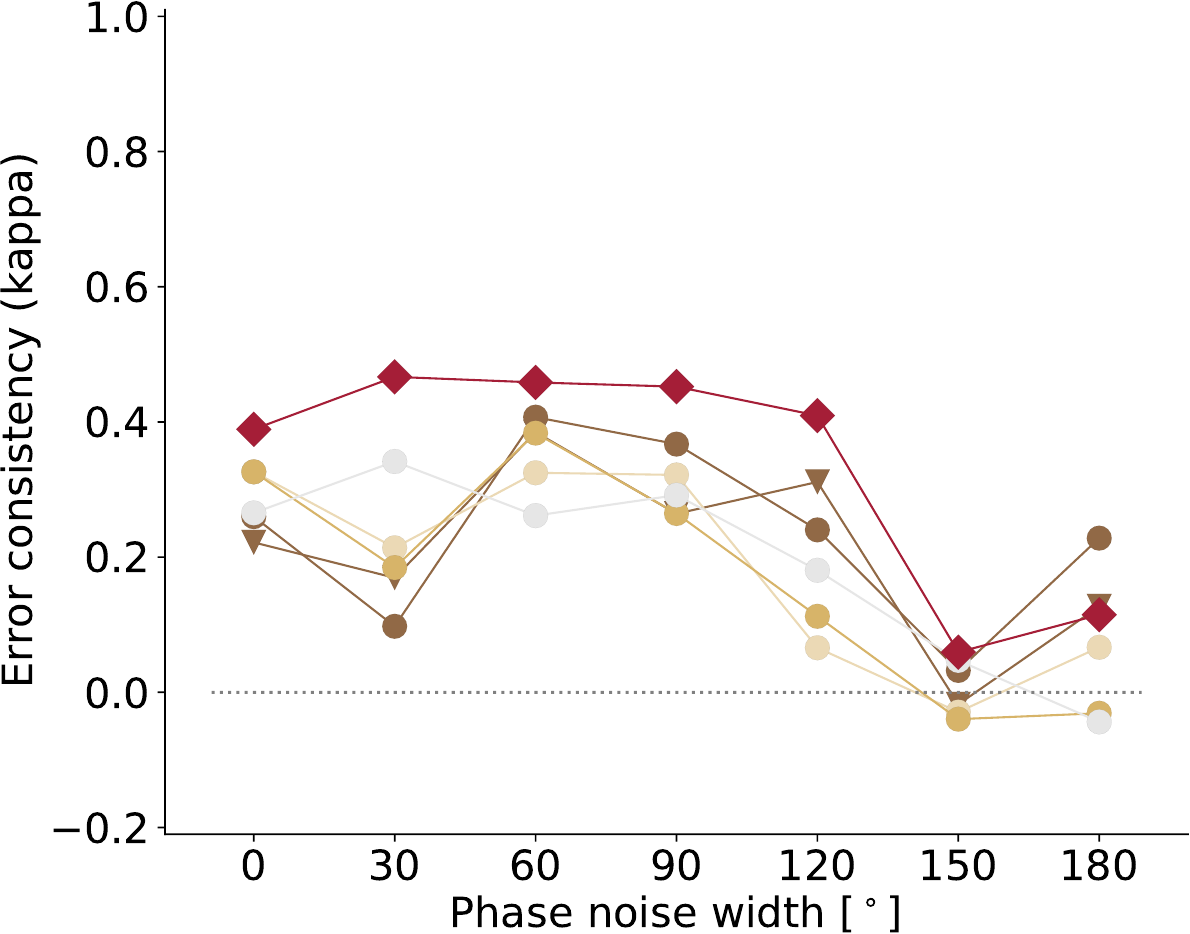}
			\vspace{\captionspace}
			\caption*{}
			\vspace{\captionspaceII}
		\end{subfigure}\hfill

	\begin{subfigure}{\figwidth}
		\centering
		\includegraphics[width=\linewidth]{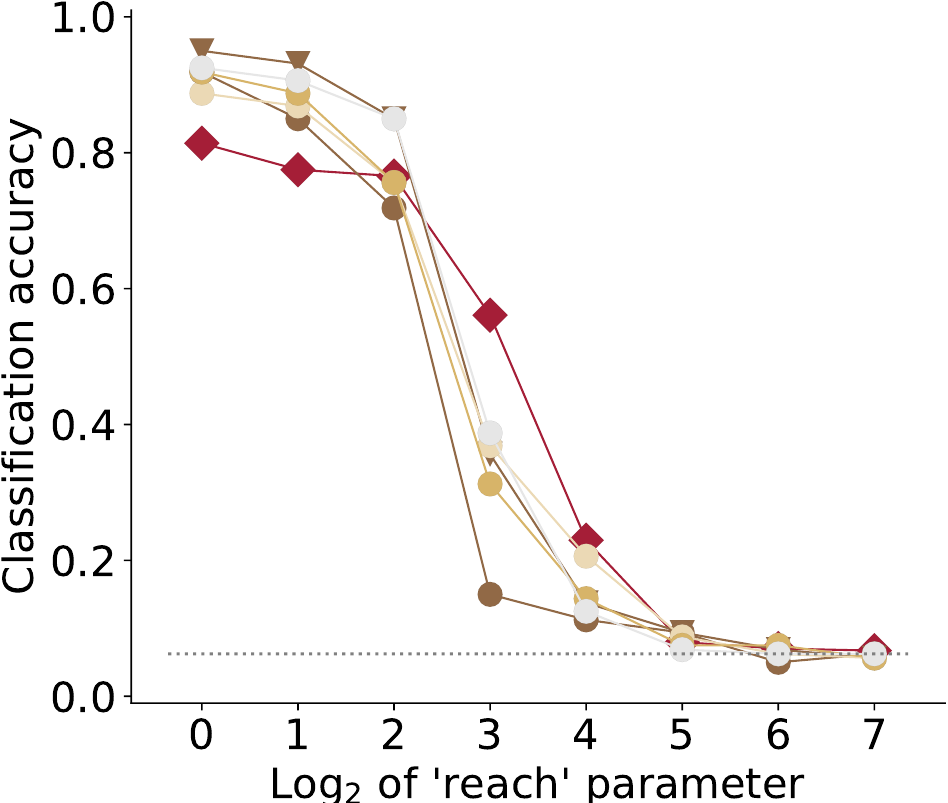}
		\vspace{\captionspace}
		\caption{Eidolon II}
		\vspace{\captionspaceII}
	\end{subfigure}\hfill
	\begin{subfigure}{\figwidth}
		\centering
		\includegraphics[width=\linewidth]{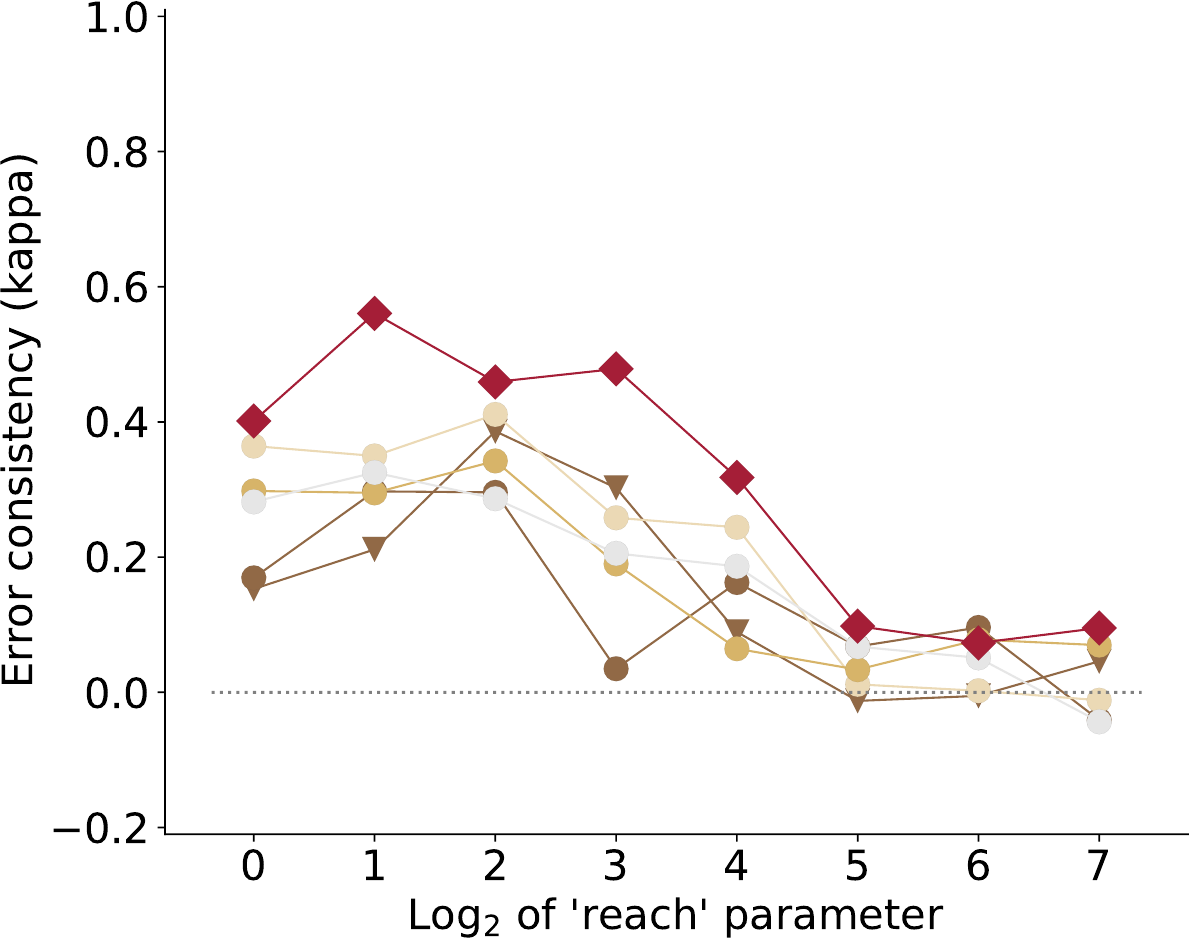}
		\vspace{\captionspace}
		\caption*{}
		\vspace{\captionspaceII}
	\end{subfigure}\hfill
	\begin{subfigure}{\figwidth}
		\centering
		\includegraphics[width=\linewidth]{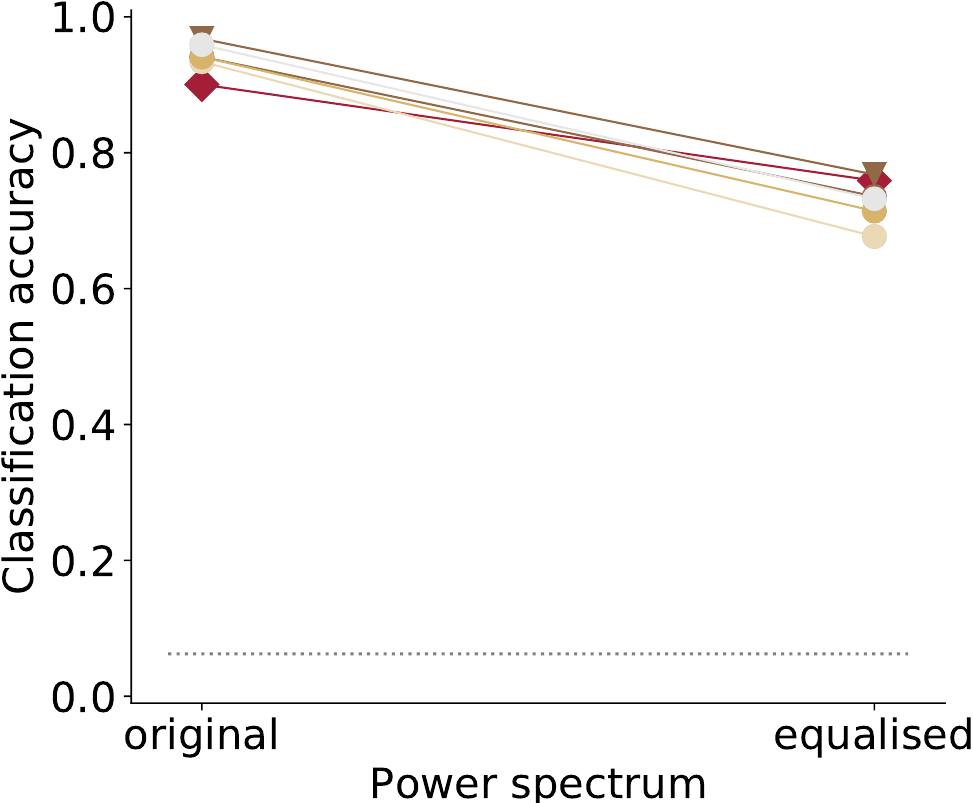}
		\vspace{\captionspace}
		\caption{Power equalisation}
		\vspace{\captionspaceII}
	\end{subfigure}\hfill
	\begin{subfigure}{\figwidth}
		\centering
		\includegraphics[width=\linewidth]{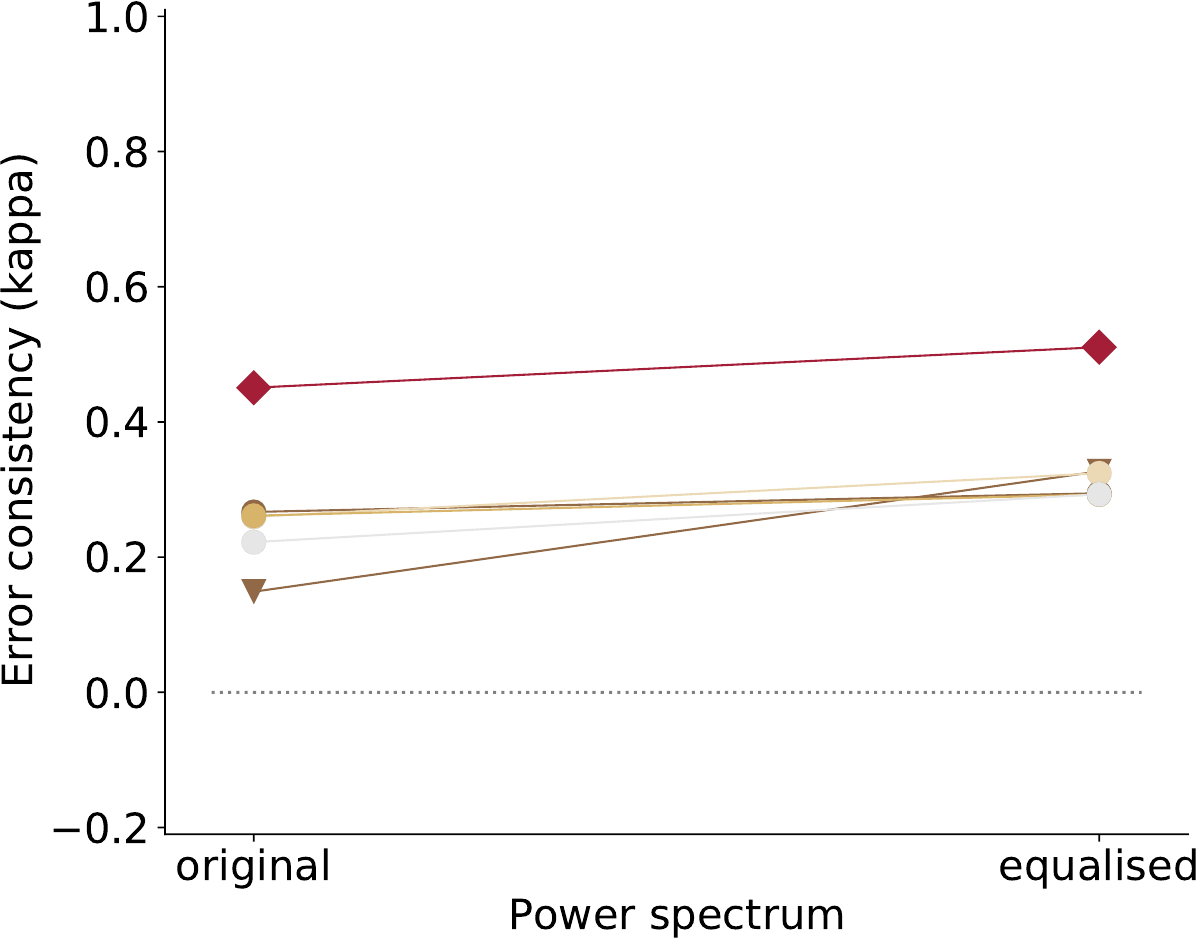}
		\vspace{\captionspace}
		\caption*{}
		\vspace{\captionspaceII}
	\end{subfigure}\hfill

	\begin{subfigure}{\figwidth}
		\centering
		\includegraphics[width=\linewidth]{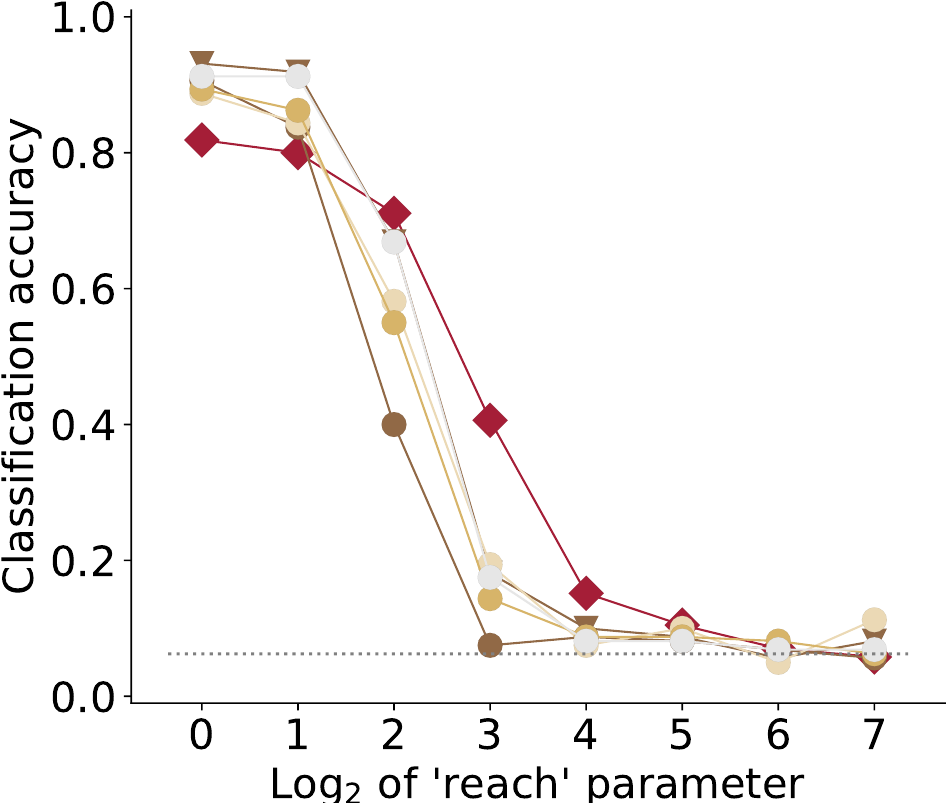}
		\vspace{\captionspace}
		\caption{Eidolon III}
	\end{subfigure}\hfill
	\begin{subfigure}{\figwidth}
		\centering        \includegraphics[width=\linewidth]{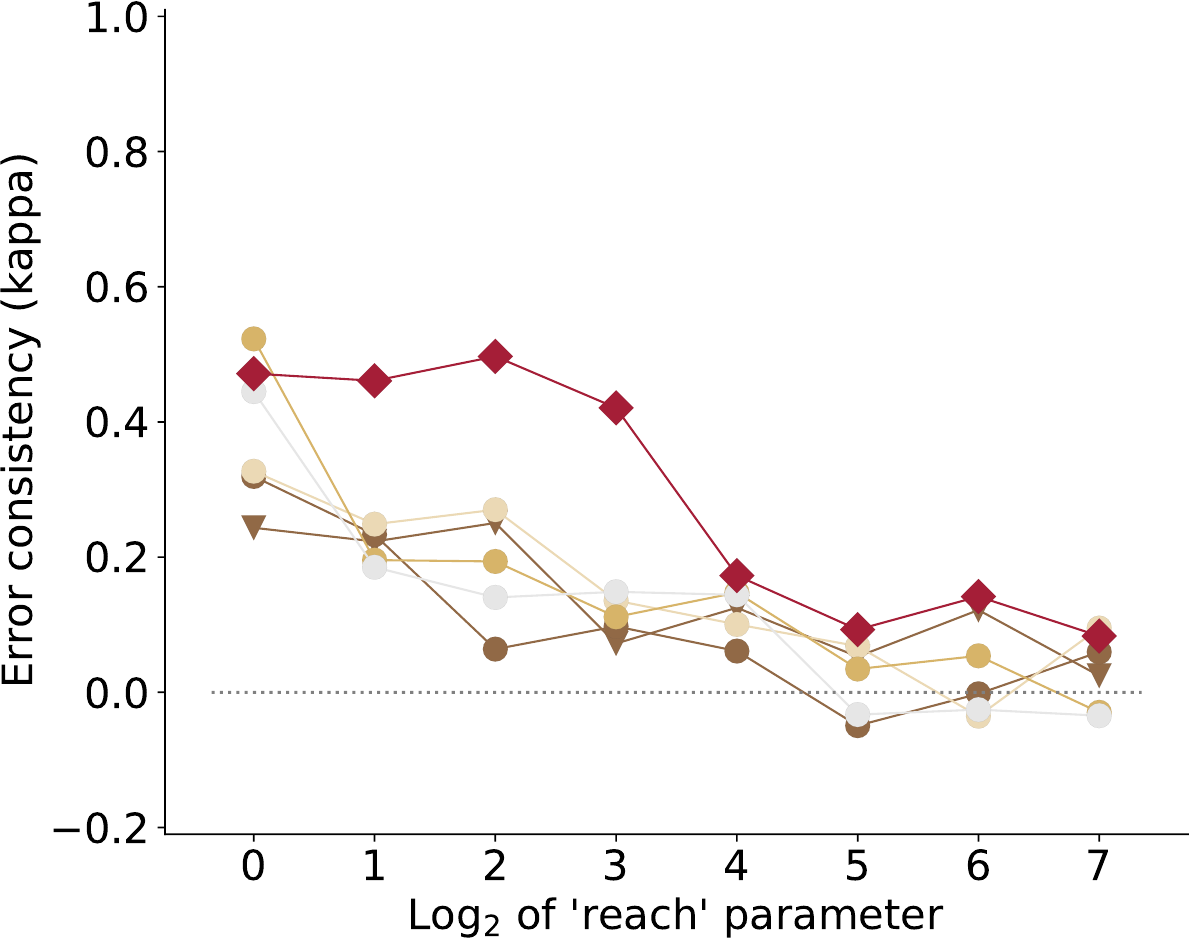}
		\vspace{\captionspace}
		\caption*{}
	\end{subfigure}\hfill
	\begin{subfigure}{\figwidth}
		\centering
		\includegraphics[width=\linewidth]{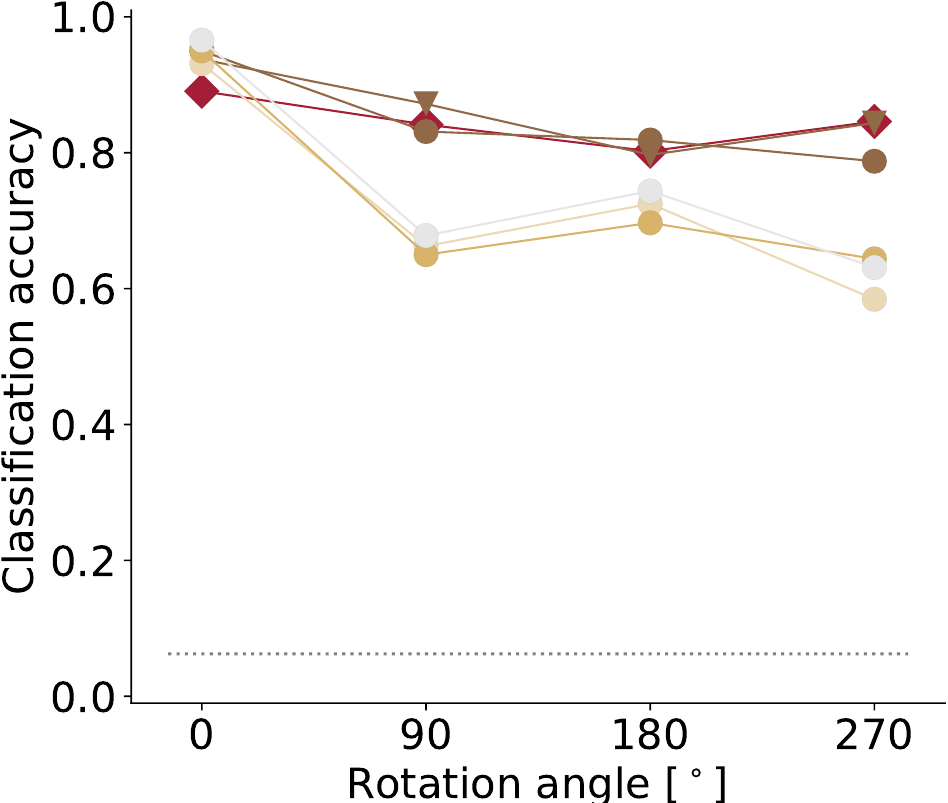}
		\vspace{\captionspace}
		\caption{Rotation}
	\end{subfigure}\hfill
	\begin{subfigure}{\figwidth}
		\centering
		\includegraphics[width=\linewidth]{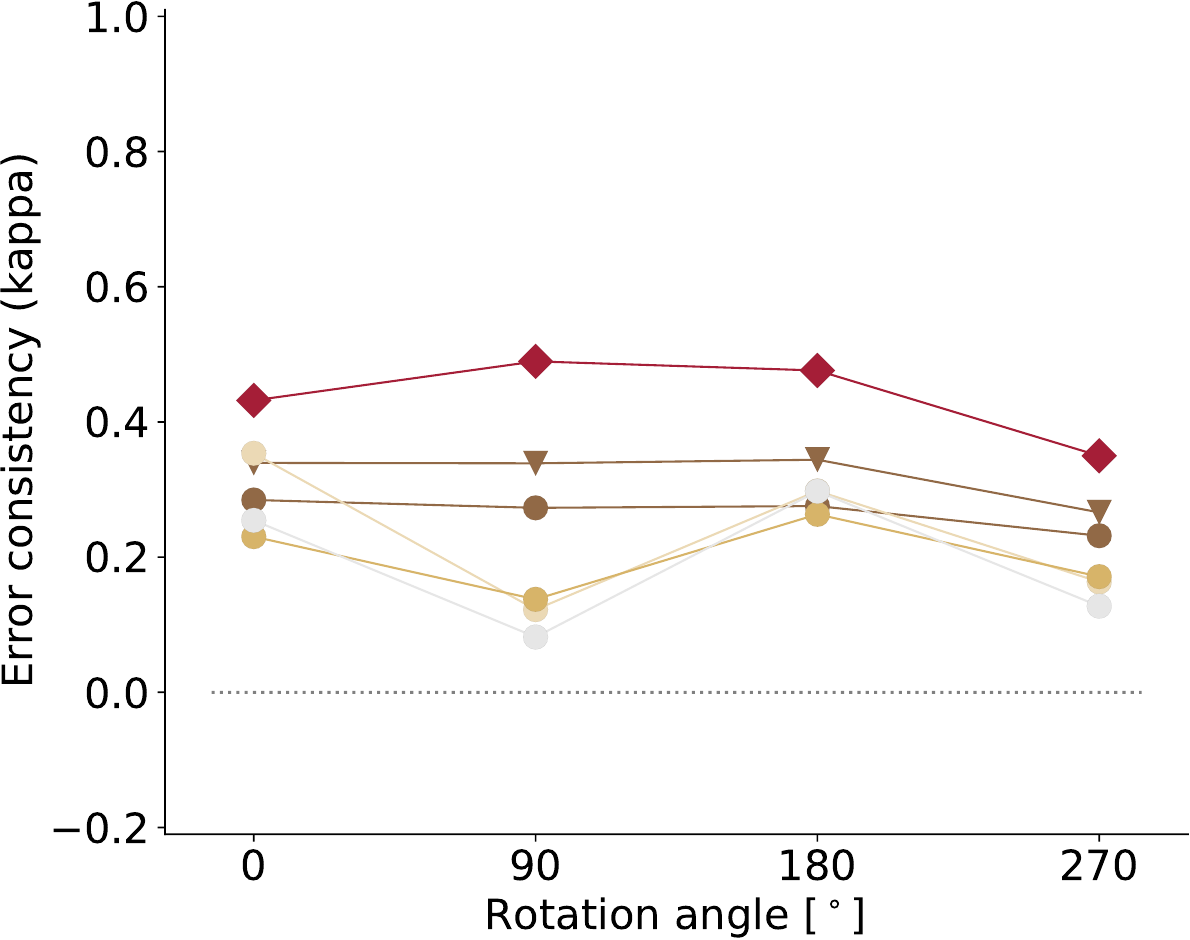}
		\vspace{\captionspace}
		\caption*{}
	\end{subfigure}\hfill
	\caption{Do CLIP-provided labels lead to better performance? Comparison of standard ResNet-50 (\textcolor{supervised.grey}{light grey circles}), CLIP with vision transformer backend (\textcolor{brown1}{brown triangles}), CLIP with ResNet-50 backend (\textcolor{brown1}{brown circles}), and standard ResNet-50 with hard labels (\textcolor{clip.hard.labels}{bright yellow circles}) vs.\ soft labels (\textcolor{clip.soft.labels}{dark yellow circles}) provided by evaluating standard CLIP on ImageNet; as well as humans \textcolor{red}{(red diamonds)} for comparison. Symbols indicate architecture type ($\ocircle$ convolutional, $\triangledown$ vision transformer, $\lozenge$ human); best viewed on screen. With the exception of high-pass filtered images, standard CLIP training with a ResNet-50 backbone performs fairly poorly.}
	\label{fig:CLIP_results_accuracy_error_consistency}
\end{figure}

\end{document}